%% file: example.tex
\documentclass{article}

\usepackage[final]{corl_2025} % Uncomment for the camera-ready ``final'' version.
\usepackage{amssymb}
\usepackage{graphicx}
\usepackage[table]{xcolor}
\usepackage{booktabs}
\usepackage{array}
\usepackage{tabularx} 
\usepackage{adjustbox} 
\usepackage{listings}
\usepackage{makecell}
\usepackage{placeins}
\usepackage{pifont} % For \ding
\usepackage{tikz}
\usepackage{float} 
\usepackage{subcaption}
\usepackage{caption}
\usepackage{multirow}

\definecolor{darkgreen}{RGB}{0, 125, 0} % Darker green color
\definecolor{codegreen}{rgb}{0,0.6,0}
\definecolor{codegray}{rgb}{0.5,0.5,0.5}
\definecolor{codepurple}{rgb}{0.58,0,0.82}
\definecolor{backcolour}{rgb}{0.95,0.95,0.92}
\definecolor{codeblue}{rgb}{0.21,0.47,0.56}

\lstdefinestyle{mystyle}{
    backgroundcolor=\color{backcolour},   % Background color
    commentstyle=\color{codegreen},      % Comments
    keywordstyle=\color{codeblue},           % Keywords
    numberstyle=\tiny\color{codegray},   % Line numbers
    stringstyle=\color{codepurple},      % Strings
    basicstyle=\scriptsize\ttfamily,                % Basic font style
    breakatwhitespace=false,             % Automatic breaks at whitespaces
    breaklines=true,                     % Automatic line breaking
    captionpos=b,                        % Caption position
    keepspaces=true,                     % Keep spaces
    numbers=none,                        % Line numbers on the left
    showspaces=false,                    % Do not show spaces
    showstringspaces=false,              % Do not show string spaces
    showtabs=false,                      % Do not show tabs
    tabsize=2,                            % Tab size
    escapeinside={(*@}{@*)},  % define markers for escaping
}

\title{IRIS: An Immersive Robot Interaction System}

\author{\small\textbf{Xinkai Jiang}$^1$, \textbf{Qihao Yuan}$^2$, \textbf{Enes Ulas Dincer}$^1$, \textbf{Hongyi Zhou}$^1$, \textbf{Ge Li}$^1$, \textbf{Xueyin Li}$^1$, \\
\textbf{Xiaogang Jia}$^1$, \textbf{Timo Schnizer}$^1$, \textbf{Nicolas Schreiber}$^1$, \textbf{Weiran Liao}$^1$, \textbf{Julius Haag}$^1$, \\
\textbf{Kailai Li}$^2$, \textbf{Gerhard Neumann}$^1$, \textbf{Rudolf Lioutikov}$^1$
\\[0.2em]
$^{1}$Karlsruhe Institute of Technology, $^{2}$University of Groningen
\\
Project Page: \url{https://intuitive-robots.github.io/iris-project-page/}
}

\begin{document}
\maketitle

% \nolinenumbers
\begin{center}
\centering
\captionsetup{type=figure}
\includegraphics[width=\linewidth]{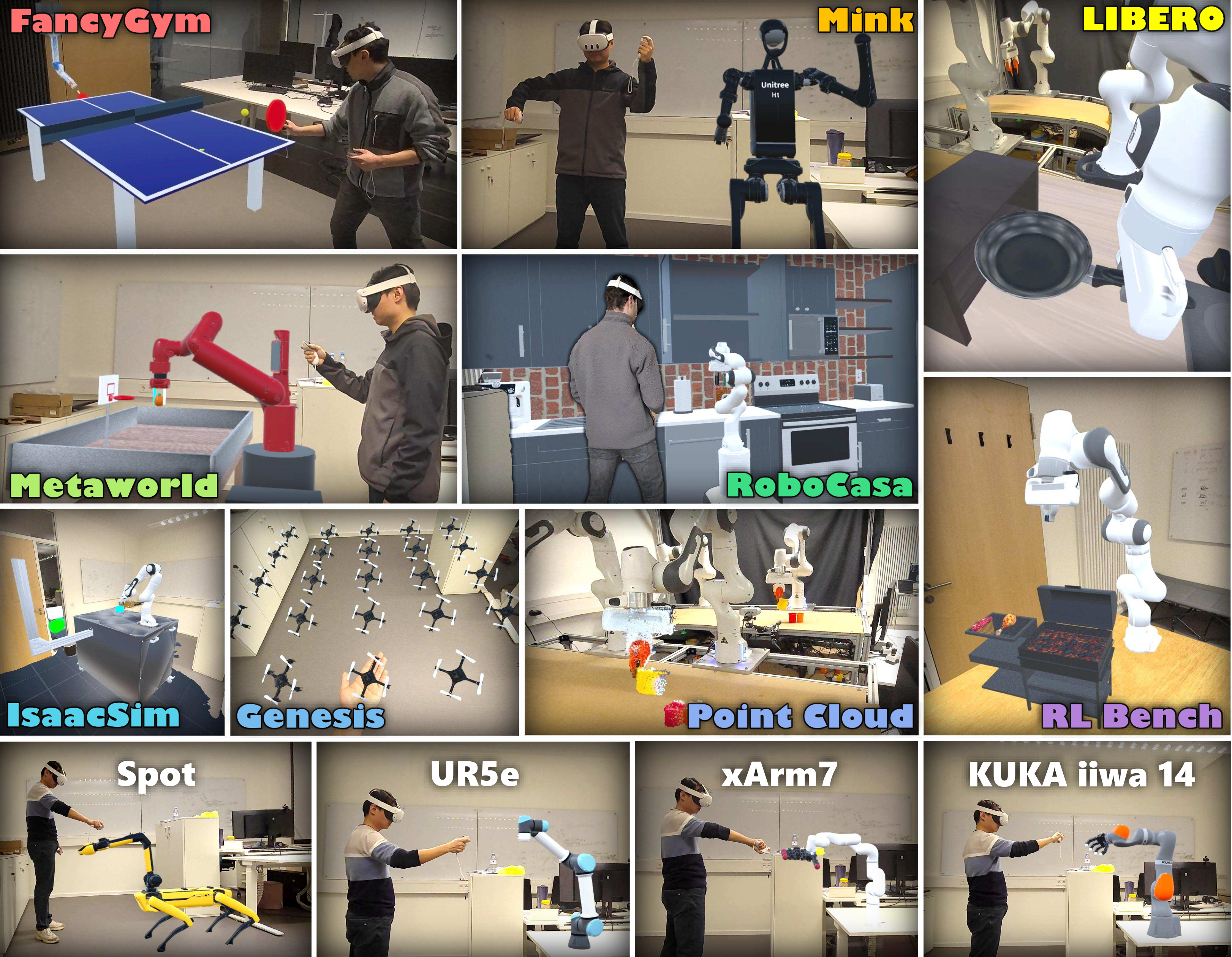} % old layout
\captionof{figure}{
We present \textbf{IRIS},
an \textbf{I}mmersive \textbf{R}obot \textbf{I}nteraction \textbf{S}ystem designed to support various simulators and real-world scenarios.
% The images shown were captured using another Meta Quest 3 headset, which leverages the Depth API \citep{metaMetaDevelopers} to enable virtual objects to blend naturally with real-world elements rather than always rendering in the foreground. To differentiate between benchmarks, we use distinct labels for each.
}
\label{fig:front_page}
\end{center}
% \linenumbers

%===============================================================================

% \begin{abstract}
%     The purpose of this document is to provide both the basic paper template and submission guidelines. Abstracts should be a single paragraph, between 4--6 sentences long, ideally. Gross violations will trigger corrections at the camera-ready phase.
% \end{abstract}
\input{sections/00_abstract}


%===============================================================================

\input{sections/01_introduction}

\input{sections/02_related_work}
\input{sections/03_system_overview}

\input{sections/04_system_application}

\input{sections/05_experiments}

\input{sections/06_limitation_and_conclusion}

\clearpage
\input{sections/07_limitations}
% The acknowledgments are automatically included only in the final and preprint versions of the paper.
\acknowledgments{
The presented work was funded by the Deutsche Forschungsgemeinschaft (DFG, German Research Foundation) – 448648559.
Xinkai Jiang and Xiaogang Jia acknowledge the support from the China Scholarship Council (CSC).
}

%===============================================================================

\bibliography{example}  % .bib

\clearpage
\input{appendix/all_the_appendix}

\end{document}

%% file: sections/00_abstract.tex
\begin{abstract}

This paper introduces IRIS, an Immersive Robot Interaction System leveraging Extended Reality (XR). Existing XR-based systems enable efficient data collection but are often challenging to reproduce and reuse due to their specificity to particular robots, objects, simulators, and environments. IRIS addresses these issues by supporting immersive interaction and data collection across diverse simulators and real-world scenarios. It visualizes arbitrary rigid and deformable objects, robots from simulation, and integrates real-time sensor-generated point clouds for real-world applications. Additionally, IRIS enhances collaborative capabilities by enabling multiple users to simultaneously interact within the same virtual scene. Extensive experiments demonstrate that IRIS offers efficient and intuitive data collection in both simulated and real-world settings.

\end{abstract}

%% file: sections/01_introduction.tex
\section{Introduction}

Robot learning relies on diverse and high-quality data to acquire complex behaviors \cite{aldaco2024aloha, wang2024dexcap}. Recent studies indicate that models trained on more varied and complex datasets generalize more effectively across diverse scenarios \cite{mann2020language, radford2021learning, gao2024efficient}. By providing immersive perspectives and interactions, Extended Reality
\footnote{Extended Reality (XR) is an umbrella term encompassing Augmented Reality, Mixed Reality, and Virtual Reality \cite{wikipediaExtendedReality}.}
(XR) has emerged as a promising tool for efficient and intuitive large-scale data collection in both simulation \cite{jiang2024comprehensive, arcade, dexhub-park} and real-world environments \cite{openteach, opentelevision}. However, existing XR approaches face significant challenges when reused or reproduced in new scenarios, primarily due to three limitations: \textit{asset diversity}, \textit{platform dependency}, and \textit{XR device compatibility}.

% Approaches \cite{arclfd, jiang2024comprehensive, arcade, george2025openvr, vicarios}
% suffering from
% \textit{asset limitation} 
% can only use several predefined object and robot models.
% Then,
% most existing works \cite{mosbach2022accelerating, lipton2017baxter, dexhub-park, arcade} are developed for specific simulators or real-world scenarios,
% resulting in \textit{simulator limitation}.
% This constraint significantly reduces reusability and makes adaptation to new simulation platforms challenging.
% Additionally,
% current XR frameworks
% \cite{lipton2017baxter, armada, openteach, meng2023virtual}.
% are designed for a specific version of a single XR headset,
% leading to a \textit{device limitation} 
% These limitations hamper reproducibility and broader contributions of XR based data collection and interaction to the research community.

Current approaches \cite{arclfd, jiang2024comprehensive, arcade, george2025openvr, vicarios} rely heavily on predefined sets of objects and robot models, thereby exhibiting limited \textit{asset diversity}. Furthermore, most methods \cite{mosbach2022accelerating, lipton2017baxter, dexhub-park, arcade} are specifically tailored to particular simulators or real-world conditions, resulting in substantial \textit{platform dependency}. This limitation significantly reduces reusability and complicates adaptation to different simulation platforms. Additionally, existing XR frameworks \cite{lipton2017baxter, armada, openteach, meng2023virtual} are typically optimized for specific XR headset versions, leading to poor \textit{device compatibility}. Together, these limitations severely constrain reproducibility and broader adoption of XR-based data collection and robot interaction methodologies within the research community.

To address these challenges, we propose \textbf{IRIS}—an \textbf{I}mmersive \textbf{R}obot \textbf{I}nteraction \textbf{S}ystem, demonstrated in Figure\ref{fig:front_page}. IRIS is a general and extensible framework that supports various simulators and real-world environments, with compatibility across different XR headsets. It is designed to generalize robot applications with XR across six key features: \textit{Cross-Scene}, \textit{Cross-Embodiment}, \textit{Cross-Simulator}, \textit{Cross-Reality}, \textit{Cross-Platform}, and \textit{Cross-User}.

\textbf{Cross-Scene} enables XR systems to handle arbitrary simulated objects, removing constraints from predefined models. IRIS introduces a unified scene specification representing all objects as data structures with meshes, materials, and textures. This specification is transmitted to XR headsets for consistent scene rendering, with dynamic updates during simulation. Through its flexible and dynamic architecture, IRIS is also the first XR-based system that supports deformable objects manipulation.
\textbf{Cross-Embodiment} is achieved by modeling robots as compositions of standard objects, enabling seamless compatibility with diverse robot embodiments without requiring specialized configurations.
\textbf{Cross-Simulator} ensures compatibility with a range of simulation engines. Since the unified scene specification is simulator-agnostic, new simulators can be supported by implementing a parser to translate their scenes into this format. This flexibility is demonstrated by IRIS's support for MuJoCo \cite{todorov2012mujoco}, IsaacSim \cite{mittal2023orbit}, CoppeliaSim \cite{coppeliaSim}, and Genesis \cite{Genesis}.
\textbf{Cross-Reality} allows IRIS to operate across both simulated and real-world environments. For real-world applications, IRIS incorporates point cloud visualization using camera data, facilitating immersive data collection.
\textbf{Cross-Platform} ensures compatibility across XR devices. IRIS implements its XR application using the Unity framework \cite{unity3dUnityManual}, with modular design separating visualization and interaction logic. This allows developers to deploy the system on new XR headsets by reusing visualization modules and implementing device-specific input handling. IRIS has been successfully deployed on the Meta Quest 3 and HoloLens 2.
\textbf{Cross-User} supports collaborative multi-user interaction within a shared scene via a communication protocol that synchronizes XR headsets. This enables coordinated tasks and collective data collection in both virtual and real environments.
Table~\ref{tab:xr_features} highlights the advantages of IRIS over existing XR-based systems across these features.

The contributions of IRIS are summarized as follows:
(1) A unified scene specification that integrates seamlessly with multiple robot simulators, enabling consistent visualization and interaction across diverse XR headsets, while promoting reproducibility and reusability.
% (2) The first XR-based system to support deformable object manipulation, allowing realistic interaction and data collection for soft-body tasks.
(2) The first XR-based system to support deformable object manipulation in robot simulators, allowing realistic interaction and data collection for soft-body tasks.
(3) A collaborative, multi-user framework for XR applications that enhances robot data collection through synchronized interactions in shared virtual or physical environments.

% \textbf{(3) A user study} demonstrating that IRIS significantly improves data collection efficiency and intuitiveness compared to the LIBERO baseline.

%% file: sections/02_related_work.tex
\section{Related Work}

\input{table/vr_work_comparison_table}

\textbf{Teleoperation-Based Data Collection on Real Robots.}
Collecting data using tele-operation on real robots has been explored by many previous works. Aloha \cite{zhao2023learning} introduced a low-cost teleoperation system that collects real-world demonstrations for imitation learning. A bimanual workspace is set up, where leader robots are used to control the follower robots. Followup work \cite{aldaco2024aloha} improved the performance, ergonomics, and robustness compared to the original design. In addition, a mobile version of Aloha \cite{fu2024mobile} improved data collection outside of lab settings. GELLO \cite{wu2023gello} supports a variety of robot arms through a 3D-printed low-cost leader robots with off-the-shelf motors. In order to tele-operate dexterous end effectors, prior work has retrieved hand motion data through visual hand tracking \cite{Qin2023AnyTeleopAG} or customized gloves \cite{wang2024dexcap}. In contrast to IRIS, none of these approaches leverages the immersive advantages of XR.

% A major disadvantage of tele-operation using controllers or leader robots is that, for each different kinematics of the target robot, a specialized physical control device has to be built.
% To tackle this issue, some XR-based teleoperation methods have been proposed, exploiting immersive approaches for real-world interactions with robots.
\textbf{XR-Based Data Collection in Real World.}
Common XR systems show virtual robots to help users understand how their movements control real robots
\cite{Qin2023AnyTeleopAG}.
For instance, recent works developed mobile apps to allow data collection in augmented reality without the need for XR headsets \cite{ar2-d2-pmlr-v229-duan23a, eve},
which allow more intuitive robot manipulation \cite{arcade,armada,jiang2024comprehensive}.
% , e.g., by aligning a virtual robot arm with the user's arm.
Instead of displaying the virtual robot in a third-person view,
\citet{opentelevision, openteach} directly provide the first-person camera feed of the real robot to the user.
% \citet{opentelevision} and \citet{openteach} provide the camera view from real robots directly instead of displaying a virtual robot.
Other systems \cite{vicarios,augmentedvisualcues,wang2024robotic,immertwin,sharedctlframework,digitaltwinmr} visualize the real-world scene in the headset and control robot arms with controllers \cite{sharedctlframework} or hand tracking \cite{wang2024robotic}.
XR-based data collection for dexterous hands has also been explored.
For example, \citet{arunachalam2023holo} tracks hand motion using camera and retargets it on the real robot hand. \citet{chen2024arcap} controls robot hand and robot arm at the same time. While these approaches do use XR, the robot data collection and interaction is limited to the real world, with no simulators used in the process.

% \textcolor{red}{More work about MR/VR/AR}

% \subsection{Robot Learning by Interaction}

% \textcolor{red}{what is the combination of demonstration and correction??? Interaction is not a good word}

% \textcolor{red}{a big table of comparison to other AR robot data collection}

\textbf{XR-Based Data Collection in Simulation.}
Real robot data collection is limited by available environments and objects.
Virtual data collection offers a more efficient way to gather demonstrations while providing access to extensive 3D asset libraries.
For instance, DART \cite{dexhub-park} runs a cloud-based simulation, and users collect demonstrations in any virtualized environment from any location. \citet{mosbach2022accelerating} collects dexterous hand manipulation data with a special glove device in physics simulations.
Although \citet{meng2023virtual} also leverages simulators, their virtual scene is a replica of the real scene, thus the flexibility of simulation is not fully exploited.

% This discussion reveals that state-of-the-art, XR-based data collection in simulation has barely been explored. Hence, IRIS presents a significant contribution towards immersive robot data collection and interaction. IRIS not only supports almost all popular simulation environments, but also connects seamlessly with robots in the real world, opening up enormous possibilities in human-involved robot learning in both virtual and real-world settings.

%% file: table/vr_work_comparison_table.tex
\definecolor{goodgreen}{HTML}{228833}
\definecolor{goodred}{HTML}{EE6677}
\definecolor{goodgray}{HTML}{BBBBBB}

\begin{table*}[t]
    \centering
    \begin{adjustbox}{max width=\textwidth}
    \renewcommand{\arraystretch}{1.2}    
    \begin{tabular}{>{\bfseries}l>{\bfseries}c>{\bfseries}c>{\bfseries}c>{\bfseries}c>{\bfseries}c>{\bfseries}c>{\bfseries}c}
        \toprule
        & \makecell{Cross-Scene}
        & \makecell{Cross-Embodiment}
        & \makecell{Cross-Simulator}
        & \makecell{Cross-Reality}
        & \makecell{Cross-Platform}
        & \makecell{Cross-User}
        & \makecell{Control Space} \\
        \midrule
        % Vicarios \cite{vicarios}                           & \xmark & \xmark & \xmark & \xmark & \xmark & \xmark \\     
        % Augmented Visual Cues \cite{augmentedvisualcues}   & \xmark & \xmark & \xmark & \xmark & \xmark & \xmark \\ 
        % OpenVR \cite{george2025openvr}                     & \xmark & \xmark & \xmark & \xmark & \xmark & \xmark \\
        \citet{digitaltwinmr}                              & \textcolor{goodred}{Limited}     & \textcolor{goodred}{Single Robot} & \textcolor{goodred}{Unity}    & \textcolor{goodred}{Real}          & \textcolor{goodred}{Meta Quest 2} & \textcolor{goodgray}{N/A} & \textcolor{goodred}{Cartesian} \\
        ARC-LfD \cite{arclfd}                              & \textcolor{goodgray}{N/A}        & \textcolor{goodred}{Single Robot} & \textcolor{goodgray}{N/A}     & \textcolor{goodred}{Real}          & \textcolor{goodred}{HoloLens}     & \textcolor{goodgray}{N/A} & \textcolor{goodred}{Cartesian} \\
        \citet{sharedctlframework}                         & \textcolor{goodred}{Limited}     & \textcolor{goodred}{Single Robot} & \textcolor{goodgray}{N/A}     & \textcolor{goodred}{Real}          & \textcolor{goodred}{HTC Vive Pro} & \textcolor{goodgray}{N/A} & \textcolor{goodred}{Cartesian} \\
        \citet{jiang2024comprehensive}                     & \textcolor{goodred}{Limited}     & \textcolor{goodred}{Single Robot} & \textcolor{goodgray}{N/A}     & \textcolor{goodred}{Real}          & \textcolor{goodred}{HoloLens 2}   & \textcolor{goodgray}{N/A} & \textcolor{goodgreen}{Joint \& Cartesian} \\
        \citet{mosbach2022accelerating}                    & \textcolor{goodgreen}{Available} & \textcolor{goodred}{Single Robot} & \textcolor{goodred}{IsaacGym} & \textcolor{goodred}{Sim}           & \textcolor{goodred}{Vive}         & \textcolor{goodgray}{N/A} & \textcolor{goodgreen}{Joint \& Cartesian} \\
        Holo-Dex \cite{holodex}                            & \textcolor{goodgray}{N/A}        & \textcolor{goodred}{Single Robot} & \textcolor{goodgray}{N/A}     & \textcolor{goodred}{Real}          & \textcolor{goodred}{Meta Quest 2} & \textcolor{goodgray}{N/A} & \textcolor{goodred}{Joint} \\
        ARCADE \cite{arcade}                               & \textcolor{goodgray}{N/A}        & \textcolor{goodred}{Single Robot} & \textcolor{goodgray}{N/A}     & \textcolor{goodred}{Real}          & \textcolor{goodred}{HoloLens 2}   & \textcolor{goodgray}{N/A} & \textcolor{goodred}{Cartesian} \\
        DART \cite{dexhub-park}                            & \textcolor{goodred}{Limited}     & \textcolor{goodred}{Limited}      & \textcolor{goodred}{Mujoco}   & \textcolor{goodred}{Sim}           & \textcolor{goodred}{Vision Pro}   & \textcolor{goodgray}{N/A} & \textcolor{goodred}{Cartesian} \\
        ARMADA \cite{armada}                               & \textcolor{goodgray}{N/A}        & \textcolor{goodred}{Limited}      & \textcolor{goodgray}{N/A}     & \textcolor{goodred}{Real}          & \textcolor{goodred}{Vision Pro}   & \textcolor{goodgray}{N/A} & \textcolor{goodred}{Cartesian} \\
        \citet{meng2023virtual}                            & \textcolor{goodred}{Limited}     & \textcolor{goodred}{Single Robot} & \textcolor{goodred}{PhysX}   & \textcolor{goodgreen}{Sim \& Real} & \textcolor{goodred}{HoloLens 2}   & \textcolor{goodgray}{N/A} & \textcolor{goodred}{Cartesian} \\
        % GELLO \cite{wu2023gello}                           & \cmark & \xmark & \xmark & \xmark & \xmark & \xmark \\
        % DexCap \cite{wang2024dexcap}                       & \xmark & \xmark & \xmark & \xmark & \xmark & \xmark \\
        % AnyTeleop \cite{Qin2023AnyTeleopAG}                & \cmark & \cmark & \cmark & \cmark & \xmark & \cmark \\
        % \citet{wang2024robotic}                            & \xmark & \xmark & \xmark & \xmark & \xmark & \xmark \\
        Bunny-VisionPro \cite{bunnyvisionpro}              & \textcolor{goodgray}{N/A}        & \textcolor{goodred}{Single Robot} & \textcolor{goodgray}{N/A}     & \textcolor{goodred}{Real}          & \textcolor{goodred}{Vision Pro}   & \textcolor{goodgray}{N/A} & \textcolor{goodred}{Cartesian} \\
        IMMERTWIN \cite{immertwin}                         & \textcolor{goodgray}{N/A}        & \textcolor{goodred}{Limited}      & \textcolor{goodgray}{N/A}     & \textcolor{goodred}{Real}          & \textcolor{goodred}{HTC Vive}     & \textcolor{goodgray}{N/A} & \textcolor{goodred}{Cartesian} \\
        Open-TeleVision \cite{opentelevision}              & \textcolor{goodgray}{N/A}        & \textcolor{goodred}{Limited}      & \textcolor{goodgray}{N/A}     & \textcolor{goodred}{Real}          & \textcolor{goodgreen}{Meta Quest, Vision Pro} & \textcolor{goodgray}{N/A} & \textcolor{goodred}{Cartesian} \\
        \citet{szczurek2023multimodal}                     & \textcolor{goodgray}{N/A}        & \textcolor{goodred}{Limited}      & \textcolor{goodgray}{N/A}     & \textcolor{goodred}{Real}          & \textcolor{goodred}{HoloLens 2}   & \textcolor{goodgreen}{Available} & \textcolor{goodred}{Joint \& Cartesian} \\
        OPEN TEACH \cite{openteach}                        & \textcolor{goodgray}{N/A}        & \textcolor{goodgreen}{Available}  & \textcolor{goodgray}{N/A}     & \textcolor{goodred}{Real}          & \textcolor{goodred}{Meta Quest 3} & \textcolor{goodgray}{N/A} & \textcolor{goodgreen}{Joint \& Cartesian} \\
        \midrule
        \rowcolor{yellow!50}
        \textbf{Ours}                                      & \textcolor{goodgreen}{Available} & \textcolor{goodgreen}{Available}  & \textcolor{goodgreen}{Mujoco, CoppeliaSim, IsaacSim} & \textcolor{goodgreen}{Sim \& Real} & \textcolor{goodgreen}{Meta Quest 3, HoloLens 2} & \textcolor{goodgreen}{Available} & \textcolor{goodgreen}{Joint \& Cartesian} \\
        \bottomrule        
        \end{tabular}
    \end{adjustbox}    
    \caption{Comparison of XR-based system. IRIS is compared with related works in seven aspects.}
    \vspace{-0.3cm}
    \label{tab:xr_features}
\end{table*}

%% file: sections/03_system_overview.tex
\section{System Overview}

This section presents the hardware and software architecture of IRIS,
along with several applications which have been explored in this paper.
An overview of its paradigm is shown in Figure \ref{fig:system_overview}.

\subsection{System Architecture}

\input{image/system_overview}

% The IRIS system operates across simulation and/or sensor-processing computers, multiple XR headsets, and other monitoring and control programs, requiring a robust and reliable network connection between them.
\textbf{Node Communication Protocol.}
The IRIS system operates across simulation or sensor-processing computers and XR headsets.
It provides a dedicated Python library, SimPublisher, for use on the computers, in combination with a Unity application deployed on the headsets.
Through this library, users are able to render objects remotely on the headsets and manage groups of headsets without the need to write C\# code.
This approach eliminates the requirement for developing separate Unity projects, thereby substantially reducing the cost and complexity associated with integrating robotic systems into VR applications.
This architecture requires a robust network connection between all components.
While Robot Operating System (ROS) \cite{quigley2009ros} offers a general communication framework and it is possible to connect to Unity and XR development,
it is quite heavy to deploy.
IRIS was intentionally designed to be decoupled from the ROS framework in order to support a broader range of simulators and avoid introducing unnecessary dependencies.
It built an lightweight communication protocol based on ZeroMQ (ZMQ) \cite{zeromqZeroMQ}, and extended it by auto-node discovery features.
This design choice enhances modularity and simplifies integration in diverse environments. Importantly, IRIS remains compatible with ROS through a lightweight and flexible communication interface, allowing seamless integration when needed.
To ensure node discovery, the master node broadcasts UDP messages to the broadcast port on the network at a specific frequency.
The network is built via Wi-Fi or cable.
% When a new XR node is launched, 
% it listens on the broadcast port to receive a broadcast message from the master node. 
% Upon receiving the broadcast message,
% the XR node extracts the master node's details, including its ZMQ socket address and port, to establish a ZMQ connection.
When a new XR node launches,
it listens to the broadcast port and receives messages from the master node.
Then it extracts connection details from these messages and establishes a ZMQ connection.
This protocol achieves \textbf{Cross-User} ability, ensures reliable communication, automatic reconnection, and smooth recovery from disconnections,
making it ideal for dynamic multi-device XR systems.

% To visualize a scene with arbitrary objects, the XR headsets need to receive the scene model from the simulation and reconstruct it.
% However, the XR application and the simulation run on different devices and use different software architectures (the XR application is developed with C\# and Unity, while simulators might be built in Python or C++).
% This makes it impractical to directly transfer the scene from the simulation to the XR environment.
% To address this issue, existing solutions rely on predefined models in the XR application for specific robots and assets, requiring a static set of models to be maintained within the application.
% This approach restricts flexibility and reusability, preventing the support of robots or objects not included in the model set.
% To address this issue, IRIS introduces a novel unified scene specification, which is generated by parsing the scene directly from the simulation.
% This specification is subsequently transmitted to the headsets using the node communication protocol, enabling the XR application to accurately and dynamically recreate the scene in real time.
\textbf{Unified Scene Representation.}
% To visualize simulation scenes and virtual objects in XR headsets,
% Current solutions \cite{jiang2024comprehensive, dexhub-park, arcade, arclfd} use predefined models in XR applications,
% limiting flexibility and support for new objects and robots.
% IRIS solves this by introducing a unified scene representation (USR) which is parsed from simulations.
% Although URDF is mainly used to load scene descriptions,
% each simulator still relies on its own internal data structures at run time,
% and URDF cannot contain mesh and texture information.
% Unity’s GameObject, in turn, is a native construct that Python-based simulation or robotics scripts cannot easily create, and it cannot be serialized directly for streaming to VR headsets.
% To address these limitations, the USR was introduced as a middleware layer between multiple simulators and Unity.
% USR objects can be generated in Python, serialized, and then loaded into Unity without extra conversion.
To visualize simulation scenes and virtual objects in XR headsets,
existing solutions \cite{jiang2024comprehensive, dexhub-park, arcade, arclfd} rely on predefined models,
which limits flexibility for incorporating new objects and robots.
IRIS overcomes this limitation with a Unified Scene Representation (USR) parsed directly from simulations.
Unlike URDF, USR can be serialized with meshes and texture information and is adapted to Unity’s GameObject structure,
allowing scenes to be reconstructed natively in Unity.
Serving as a middleware layer between simulators and Unity,
USR enables objects to be generated in Python, serialized, and seamlessly loaded into Unity without additional conversion.
% and transmits all the data to headsets via node communication.
The USR includes all objects with their geometry, meshes, materials, and textures.
% In the specification, 
% geometry defines the object's shape (e.g., cube, sphere, capsule, cylinder, or mesh).
% The mesh contains vertex lists, face lists, normals, and texture coordinates. 
% Materials define surface properties, including color, emission color, reflectance, and attached textures, 
% and the texture is an image representing the object's surface appearance.
All the objects is loaded in this specification using a kinematic tree structure and serialized into byte format.
IRIS application rebuild an identical scene upon received this specification from simulation node.
% it rebuilds the scene upon receiving the kinematic tree and then requests the meshes and textures from the simulation server in byte format.
% IRIS creates its own Python library to help users generate this specification from simulation automatically.  
% Afterwards, SimPublisher (IRIS Python Server) continually acquires simulation states and forwards them to the XR headset.
IRIS provides a custom Python library named \textit{SimPublisher} that automatically generates specifications from simulation data,
then it continuously collects simulation states and transmits them to headsets at a fixed frequency.
This scene specification enables IRIS to process all kinds of robots and objects in simulation,
facilitating both \textbf{Cross-Scene} and \textbf{Cross-Embodiment} capabilities.
The USR is a general definition that does not rely on any specific simulator.
Hence, IRIS can be easily adapted to various simulators by implementing a new simulation parser to generate a scene specification from the simulator and a new publisher to update the states of the scene.
Currently, IRIS supports scene parsers for \textbf{MuJoCo}, \textbf{IsaacSim}, \textbf{CoppeliaSim}, and \textbf{Genesis},
with the potential to be extended to other simulation engines as desired.
This demonstrates that IRIS can be easily adapted to various benchmarks and simulators, highlighting its \textbf{Cross-Simulator} capability.

% \subsubsection{Simplified API Access to Interaction Data}
% \label{sec:interaction_data_collection}
% IRIS provides user-friendly access to interaction data through the \textit{SimPublisher}, which supports both Meta Quest 3 and HoloLens 2. 
% Users only need to create a new device instance and assign a name to the connection.
% The \textit{ScenePublisher} then waits automatically for the device to launch.
% Once the device is launched, the instance receives messages from the XR headset, enabling users to access various types of interaction data through different methods.
% This process makes it straightforward to retrieve data such as hand tracking, motion controller inputs, and other relevant interaction information.

% \subsection{Hardware Setting}

% \subsubsection{Multiple Headsets Compatibility}
% \textcolor{red}{rewrite this part}

% The basic hardware of IRIS is one PC for running the simulation and XR headsets.
% The whole architecture is extendable so that other hardware is also easy to integrate into IRIS if they use IRIS communication protocol.
% To project the scene to the MR devices,
% a local WIFI network is utilized to build communication between the simulation PC, the headset, and other devices.
% The local network has a shorter communication delay compared to the remote network or the cloud.
% Currently, we tested our system on HoloLens 2 \cite{hololens2_lolambean_2023} and Meta Quest 3.

\textbf{Multiple Headsets Compatibility.}
IRIS implements an XR application using Unity.
% featuring essential capabilities such as node communication protocols, scene construction based on scene specifications, and synchronization with remote simulations.
This application can be directly deployed to other headset platforms using the Unity deployment pipeline, showcasing IRIS's \textbf{Cross-Platform} capability.
% For input data handling (reading and sending), the application requires the use of platform-specific APIs, making it impractical to rely on a generalized framework.
% This approach separates the visualization and interaction components, minimizing the effort needed to transfer IRIS codebase to new platforms.
Currently, IRIS has been tested on HoloLens 2 \cite{hololens2_lolambean_2023} and Meta Quest 3 \cite{wikipediaMetaQuest}.
Due to Meta Quest 3 visualization resolution is better than HoloLens 2, 
this paper conducted experiments and displayed XR scenes using this headset.

% \subsubsection{Intuitive Robot Control Interface}
% \label{sec:intuitive_robot_interface}
% \textcolor{red}{should have some images of each interface}

\textbf{Intuitive Robot Control Interface.}
\label{sec:intuitive_robot_control_interface}
In data collection tasks or robot interaction,
robot control interfaces are used to operate the robot in both simulated and real-world environments.
Based on prior work \cite{jiang2024comprehensive, openteach},
IRIS implemented
Kinesthetic Teaching (KT) and Motion Controller (MC) 
as its default robot controllers.
The details of these two interfaces are in the Appendix. \ref{app:intuitive_robot_interface}.
These two methods were used and evaluated with other interfaces in Sec. \ref{sec:user_study_result}.
IRIS's flexible framework allows users to easily customize and implement additional control interfaces,
including hand tracking, gloves, smartphones, and motion tracking systems.
Fig. \ref{fig:interface} shows how these two interfaces work.

%% file: image/system_overview.tex
% \begin{figure*}[ht]
%     \centering
%     \includegraphics[width=\textwidth]{image/system_overview.png}
%     \caption{The graph of user study}
%     \label{fig:system_overview}
% \end{figure*}

% \begin{figure}
%     \centering
%     \includegraphics[width=\linewidth]{image/system_overview/architecture_sim.png}
%     \caption{Caption}
%     \label{fig:overview_sim}
% \end{figure}

% \begin{figure}
%     \centering
%     \includegraphics[width=\linewidth]{image/system_overview/architecture_real_wider.png}
%     % \includegraphics[width=\linewidth]{image/system_overview/architecture_real.png}
%     \caption{Caption}
%     \label{fig:overview_real}
% \end{figure}

\begin{figure*}[htbp]
    \centering
    \begin{subfigure}[b]{0.45\linewidth}
        \centering
        \includegraphics[width=\linewidth]{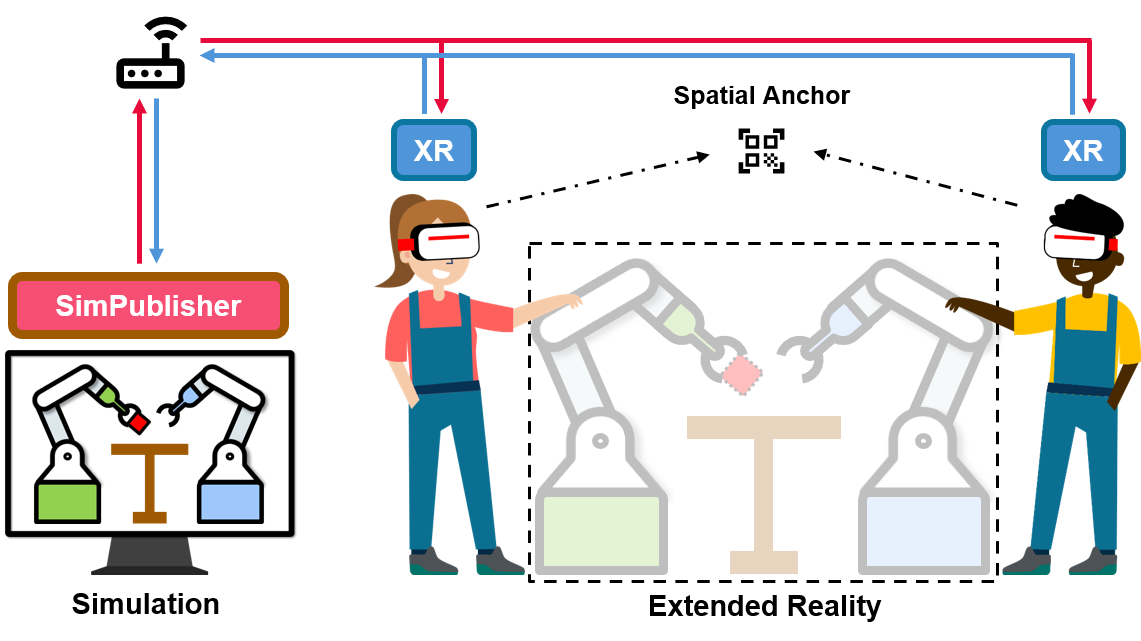}
        \caption{Interact with Robots in Simulation}
        \label{fig:overview_sim}
    \end{subfigure}
    \hfill
    \begin{subfigure}[b]{0.45\linewidth}
        \centering
        \includegraphics[width=\linewidth]{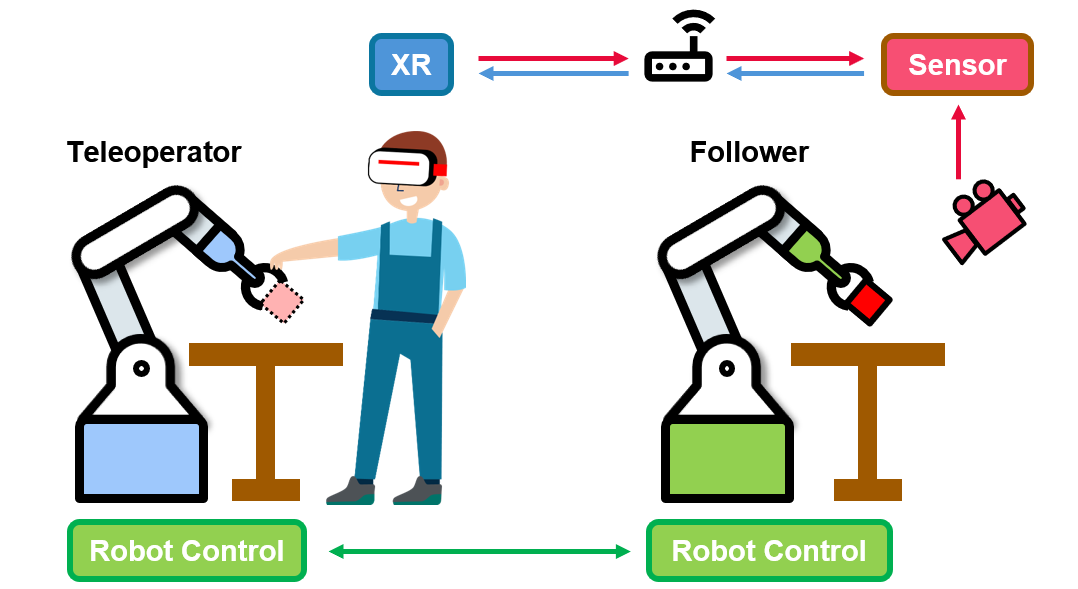}
        \caption{Interact with Robots in Real World}
        \label{fig:overview_real}
    \end{subfigure}
    \caption{
Paradigms of the system architecture in both simulation (left) and real world (right).
All the devices are connected through a Wi-Fi router.
In the left image, the simulation updates the scene to all headsets using the SimPublisher.
A spatial anchor is used to align the virtual scenes across different headsets.
In the right image, a sensor generates a point cloud transmitted to the XR headset,
allowing the operator to clearly observe the manipulated object in front of the follower robot.
}
    \vspace{-0.8cm}
    \label{fig:system_overview}
\end{figure*}

%% file: sections/04_system_application.tex
\subsection{System Application}
\label{sec:system_application}

IRIS is a versatile platform with significant potential for robot learning and robotics research community.
This section outlines several applications that have been explored in this work.

%%%%%%%%%%%%%%%%%%%%%%%%%%%%%%%%
% From RSS
% \subsubsection{Robot Data Collection for Benchmarks}

% Through its flexible framework design, IRIS supports four simulators and various robot manipulation benchmarks.
% Based on the interfaces introduced in Sec. \ref{sec:interaction_data_collection},
% IRIS already provides example controllers for various benchmarks and frameworks, such as robosuite, LIBERO, RoboCasa, mink, Metaworld, Fancy Gym, and so on.
% % Since all the objects in the simulation including robots and other objects are considered as one item, so naturally IRIS supports visualize any type of robot in the XR headsets from the simulation without any configuration.
% Robots from all the frameworks can be controlled by Motion Controller or Kinesthetic Teaching.
% Fig. \ref{fig:front_page} shows the robots controlled by IRIS including Franka Panda, Aloha 2, Barrett Wam Arm, UR5e, iiwa14, Boston Dynamics Spot, Unitree H1, and more.
% From RSS
%%%%%%%%%%%%%%%%%%%%%%%%%%%%%%%%

\textbf{General Manipulation Data Collection.}
Through its flexible framework design, IRIS supports four simulators and various robot manipulation benchmarks.
IRIS has been tested in some MuJoCo-based benchmarks including \textbf{Meta World} \cite{yu2020meta}, \textbf{LIBERO} \cite{liu2024libero}, \textbf{RoboCasa} \cite{nasiriany2024robocasa}, \textbf{robosuite} \cite{zhu2020robosuite}, \textbf{Fancy Gym} \cite{fancy_gym}, and CoppeliaSim-based benchmark like 
\textbf{PyRep} \cite{james2019pyrep},
\textbf{Colosseum} \cite{pumacay2024colosseum}.
% Since all the objects in the simulation including robots and other objects are considered as one item, so naturally IRIS supports visualize any type of robot in the XR headsets from the simulation without any configuration.
Robots can be operated using either our default controllers (Sec. \ref{sec:intuitive_robot_control_interface}) or user-customized controllers.

% \textcolor{red}{picture with all kinds of robot control}

% \input{image/aloha}

\textbf{Deformable Object Manipulation.}
IRIS supports deformable object by dynamically updating the mesh in real time,
making it possible to train and test robotic algorithms for tasks that involve soft objects.
As far as we know, no existing work has explored the manipulation of deformable objects using by rendering them in XR headsets.
This paper conducted an experiment (Sec. \ref{sec:deformable_manipulation_experiment}) to validate the data collected by IRIS based on IsaacSim.

\input{image/collaborative_collection/collaborative_collection}

\textbf{Collaborative Manipulation.}
Collaborative manipulation, where multiple users provide demonstrations simultaneously, is vital for human-robot systems \cite{tung2021learning}.
% Collecting such demonstration data is challenging, requiring synchronization between participants, robots, and virtual environments.
Previous approaches \cite{Qin2023AnyTeleopAG} often use multiple screens,
lacking XR's immersion,
and typically limit control to one person while others merely observe \cite{szczurek2023multimodal}.
IRIS's communication protocol enables seamless integration of devices for controlling multiple robots in shared scenes, supporting additional XR headsets with minimal setup for dynamic collaborative environments.
% This flexibility allows for scaling or adapting the system as needed, facilitating diverse use cases and applications.
% A handover task was chosen as an example, as illustrated in Fig. \textcolor{red}{image}.
% In this scenario, two operators each control two Aloha 2 robotic arms to pass a red board between them.
Fig. \ref{fig:collaborative_collection} shows collaborative manipulation for handover task.
% \textcolor{red}{more pictures!!!!!}

\input{image/interact_with_policy/interact_with_policy}

% Simulation has long been an essential tool for training robot agents, especially for reinforcement learning (RL) policies.
% Its advantages -- such as safety, scalability, and cost-efficiency -- make it widely used in RL training. 
\textbf{High-Dynamic Task Data Collection and Interaction}
% Previous works on robot tasks utilized interfaces such as keyboards \cite{mandlekar2018roboturk}, 3D mouse \cite{luo2024precise, liu2024libero} or smartphone \cite{mandlekar2023human}. However, these methods are insufficient for tasks that require complex motions.
% % IRIS delivers an immersive experience, providing the appearance of \textit{"walking into"} the simulation.
% Motion controllers enable IRIS to capture user movements for complex tasks like competitive sports, where responsiveness and precision are essential.
% To demonstrate IRIS's capabilities,
% this paper designed an experiment where participants played table tennis against an episodic RL agent trained with BBRL \cite{otto2023deep} in simulation.
Prior work used keyboards, 3D mice, or smartphones \cite{mandlekar2018roboturk, luo2024precise, liu2024libero, mandlekar2023human}, which are limited for complex motions.
With the support for motion controllers, IRIS can capture precise user movements.
This paper demonstrates the usage of motion controllers in a simulated table tennis task against an RL agent trained with BBRL \cite{otto2023deep}.
% Table tennis is a challenging task that requires dynamic perspective shifts and precise racket control.
% Using a motion controller and XR headset, the participant can return the ball to the RL agent just like in the real world.
Fig.~\ref{fig:interact_with_policy} presents this application and 
% Apart from that, this paper also leverage collected data to train policies that learn human behavior patterns for ball returns.
an experiment of using collected data to train policies used in Sec. \ref{sec:table_tennis_evaluation}.

\input{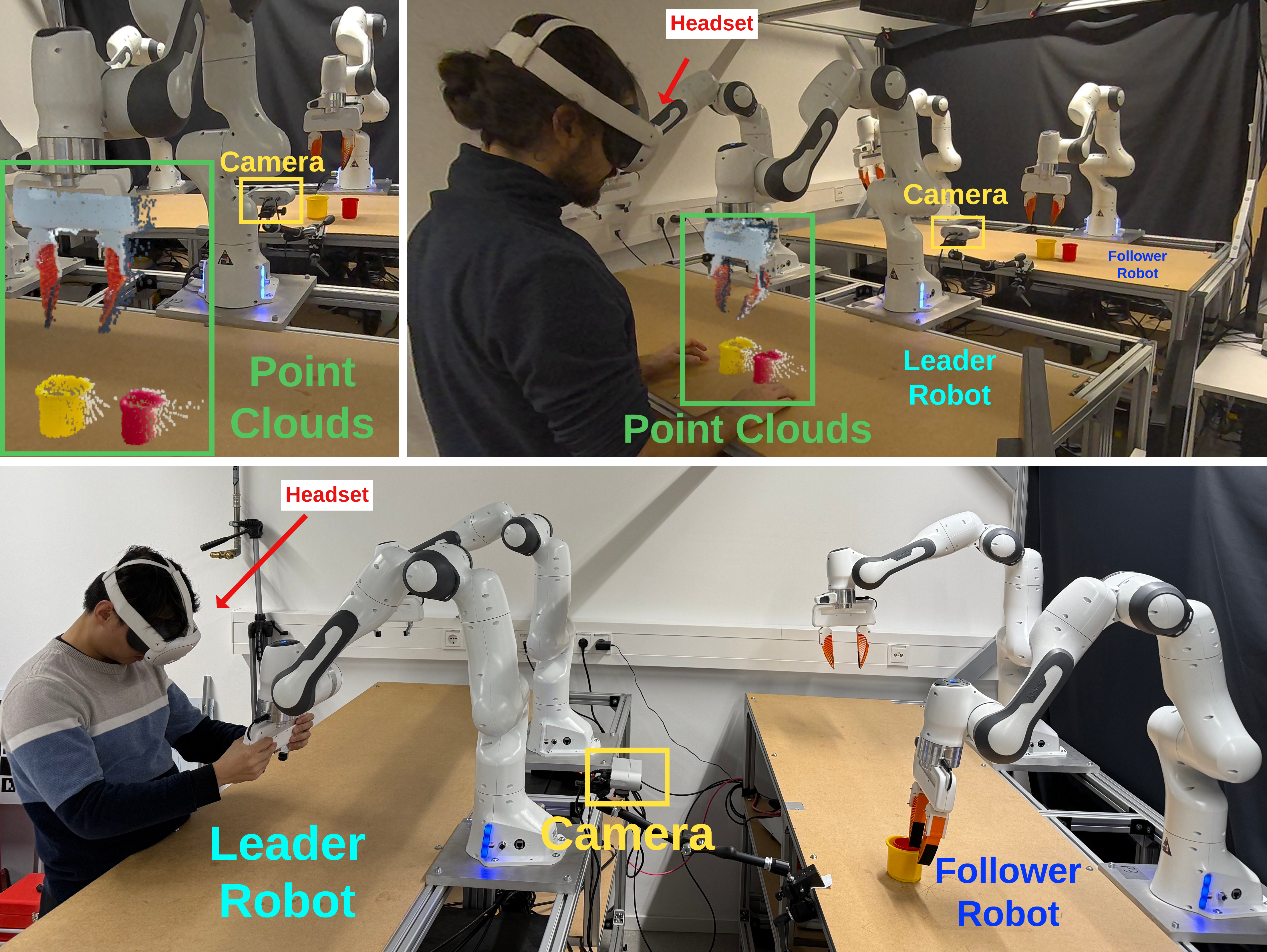}

\textbf{Real World Teleoperation and Data Collection.}
IRIS not only supports simulation but can also be used in the real-world setting.
To minimize physical obstruction from humans, tele-operation using XR is commonly employed for collecting such data.
Current approaches \cite{openteach, opentelevision} that utilize video streaming to XR headsets are restricted to fixed viewpoints.
% Some approaches, such as \cite{openteach}, utilize cameras to stream videos to XR headsets.
% However, this method lacks depth perception.
% Other works, like \cite{opentelevision}, employ movable platforms to track the movement of users' heads, allowing for active scene observation by adjusting the viewing perspective.
% While effective, this approach requires significant effort and resources to install and maintain the necessary hardware.
IRIS overcomes this limitation by projecting point cloud from depth cameras to XR headsets, ensuring both immersion and interactivity.
Fig. \ref{fig:real_robot_point_cloud_setup} shows the real-world application, and the details are in the Appendix. \ref{appendix:Point Cloud Processing Pipeline}

%% file: image/collaborative_collection/collaborative_collection.tex
\begin{figure}[htbp]
    \centering
    \includegraphics[width=0.9\linewidth]{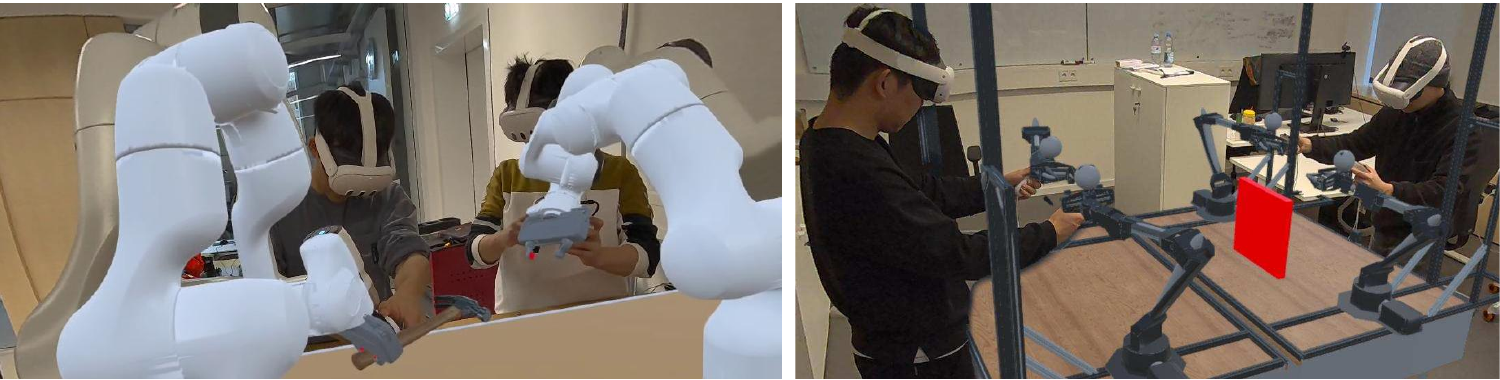}
    % \begin{subfigure}[b]{0.49\linewidth}
    %     \centering
    %     \includegraphics[width=\linewidth]{image/collaborative_collection/kt_collaborative_same_size.png}
    %     % \caption{Collaborative manipulation with two Franka Panda robots by Kinesthetic Teaching}
    %     % \label{fig:overview_sim}
    % \end{subfigure}
    % \hfill
    % \begin{subfigure}[b]{0.49\linewidth}
    %     \centering
    %     \includegraphics[width=\linewidth]{image/collaborative_collection/aloha_two_player_same_size.png}
    %     % \caption{Collaborative manipulation with two Aloha 2 by Motion Controller}
    %     % \label{fig:overview_real}
    % \end{subfigure}
    \caption{
Collaborative manipulation in simulation via IRIS. The left image shows the collaborative manipulation for handing over a hammer between two Franka Panda robots by KT, and the right image shows that collaborative manipulation for handing over a red board between two Aloha 2 Arms by MC.
}
    \vspace{-0.3cm}
    \label{fig:collaborative_collection}
\end{figure}

%% file: image/interact_with_policy/interact_with_policy.tex
\begin{figure}[t!]
    \centering
    \includegraphics[width=0.9\linewidth]{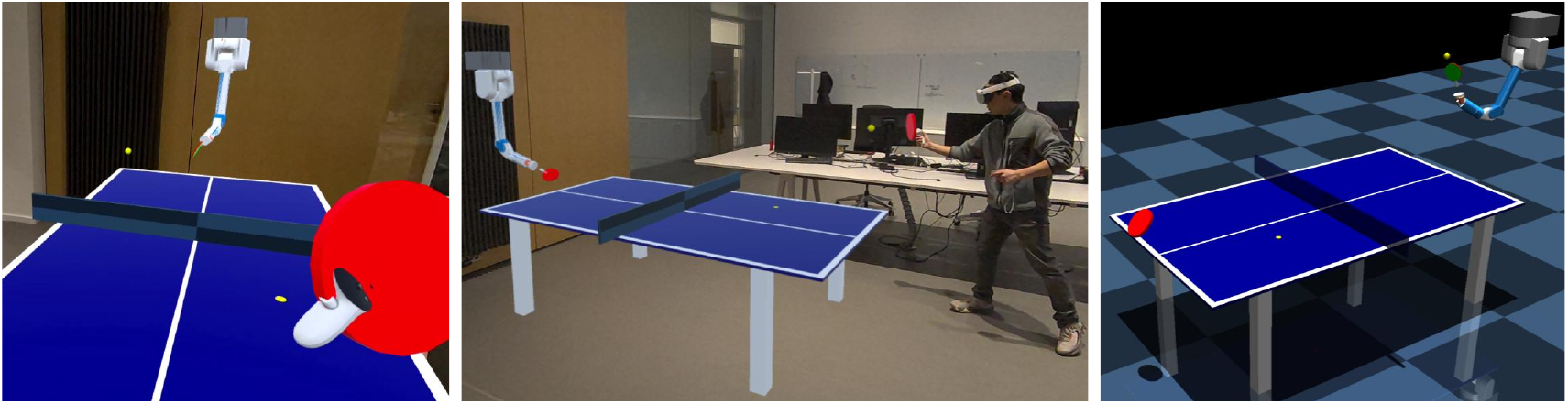}
    % \begin{subfigure}[b]{0.7\textwidth}
    %     \centering
    %     \includegraphics[width=\linewidth]{image/interact_with_policy/first_person_view.pdf}
    %     % \input{image/interact_with_policy/first_person_view.pdf}
    %     \caption{Interact with RL Agent in the first-person view (left) and simulation (right)}
    %     \label{fig:interact_with_policy_first_person}
    % \end{subfigure}
    % \vskip 1em
    % \begin{subfigure}[b]{0.7\textwidth}
    %     \centering
    %     \includegraphics[width=\linewidth]{image/interact_with_policy/third_person_view.png}
    %     \caption{Interact with RL Agent in the third-person view}
    %     \label{fig:interact_with_policy_third_person}
    % \end{subfigure}
    \caption{
        Playing table tennis with RL agent in Fancy Gym environment. The RL agent policy is trained with \textit{Deep Black-Box Reinforcement Learning} (BBRL) \cite{otto2023deep, otto2023mp3} 
    }
    \vspace{-0.3cm}
    \label{fig:interact_with_policy}
\end{figure}

%% file: image/Real_robot/real_robot_img.tex
\begin{figure}[t!]
    \centering
    \includegraphics[width=0.9\textwidth]{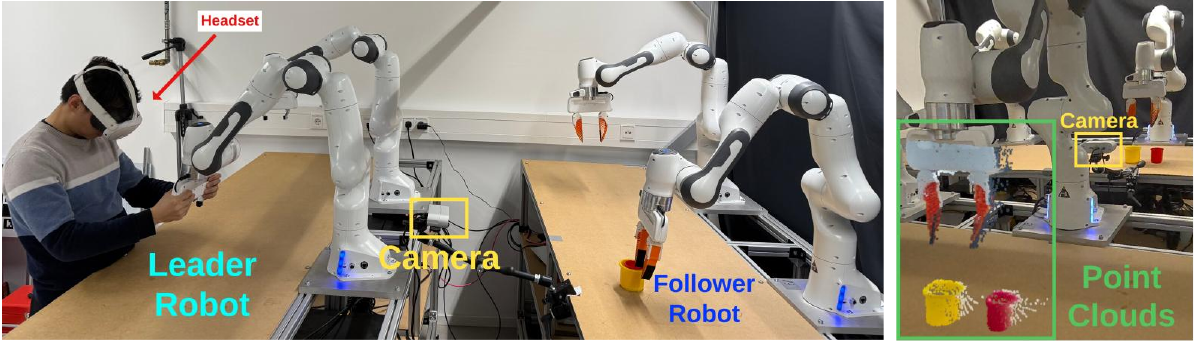}
    \caption{IRIS Real-world Application.
    This setup features two Franka robots: a leader robot controlled by a user wearing a Meta Quest 3 headset and a follower robot that mirrors its movements.
    A depth camera captures the environment for real-time point cloud visualization in XR.
    % The bottom image presents the physical setup, where the Leader Robot is directly controlled while the Follower Robot autonomously replicates actions, enabling precise, immersive remote operation.
    }
    \vspace{-0.6cm}
    \label{fig:real_robot_point_cloud_setup}
\end{figure}

%% file: sections/05_experiments.tex
\section{Experiment}

% This section evaluates IRIS's capability to create demonstrations, focusing on the efficiency and intuitiveness of data collection.
% It is followed by a performance profile analysis.
% This section addresses the following questions:
In this section, we focus on three questions to demonstrate the effectiveness and scalability of IRIS:
(1) How effective and intuitive is the IRIS system for data collection in terms of user experience?
(2) Can data collected by IRIS be effectively used for policy training in simulation?
(3) Is IRIS suitable for data collection in real-world scenarios?
To address these questions, we evaluate the performance of IRIS across three groups of tasks, with detailed experimental settings presented in Appendix~\ref{abb:exp}.
% \subsection{Robot Manipulation in Simulation}
% IRIS is promising to collect demonstration in the simualtion,
% To verify the data collection efficiency of IRIS,
% a study was conducted to evaluate it.
% LIBERO was selected and 4 tasks is picked which represent four different movement.
% which includes
% \textit{close the microwave},
% \textit{turn off the stove},
% \textit{pick up the book in the middle and place it on the cabinet shelf},
% and
% \textit{turn on the stove and put the frying pan on it}.
% To make a fair comparison, we use the task from LIBERO and the data rather than designing other task.
% The baselines are two control interfaces from LIBERO (keyboards and 3D mouse),
% From the prior study from \cite{jiang2024comprehensive} \textcolor{red}{find another one support research} and our try,
% the hand tracking is not stable comparing to motion controller. so we choose KT and MC to conduct the user study.

% There are 8 participants in this user study, we evaluate each interface by using objective metrics ans subjective metrics.
% The objective metrics includes success rate and time consumed in each tasks,
% and the subjective metrics are conducted by questionnaires includes 
% \textit{Experience}, \textit{Usefulness}, \textit{Intuitiveness}, and \textit{Efficiency}.
% Every participant will use each interface to collect demonstration five times for each task,
% they will give a score in these four dimensions from 1 to 7.

% \subsection{User Study}
\subsection{User Experience Evaluation}
\label{sec:user_study}

\input{table/success_rate_table}

To assess the efficiency and intuitiveness of IRIS data collection application, a pilot study was conducted by collecting demonstrations for LIBERO benchmark tasks \cite{liu2024libero}.
Four tasks (Fig. \ref{fig:images_grid} in the Appendix) were selected in the dimension of translation, rotation, and compound movement, including
\textit{close the microwave},
\textit{turn off the stove},
\textit{pick up the book in the middle and place it on the cabinet shelf},
and \textit{turn on the stove and put the frying pan on it}.
% To ensure a fair comparison, the tasks and data were taken directly from LIBERO \cite{liu2024libero} without any modification.
The control interface baselines for this study are two control interfaces from LIBERO: the Keyboard (KB) and the 3D Mouse (3M).
% Based on findings from prior research \cite{jiang2024comprehensive},
% hand tracking was found to be less stable than motion controllers.
% Therefore, we selected Kinesthetic Teaching and Motion Controller as the interfaces for the user study.

\textbf{Study Design}
This study involved eight participants with no prior experience using IRIS or XR headsets.
They evaluated each interface through both objective and subjective metrics.
Objective measurements included success rate and average time per task, while subjective assessments were gathered via a questionnaire based on the UMUX framework \cite{finstad2010usability}.
The questionnaire evaluates each interface by a 7-point Likert scale in four dimensions including \textit{Experience}, \textit{Usefulness}, \textit{Intuitiveness}, and \textit{Efficiency}.
This paper employed the Kruskal-Wallis test \cite{wikipediaKruskalWallisTest} for a better and more robust statistical analysis for the study.

\textbf{Study Result}
\label{sec:user_study_result}
In the result of objective metrics, Tab. \ref{tab:success_rate} and Fig. \ref{fig:user_study_result} present the success rates and the average time consumed for each interface across the tasks.
The result shows a success rate of over $90\%$ across all four tasks when using the KT and MC interfaces from IRIS,
and these XR-based interface significantly outperformed the non-XR interface ($p < 0.05$ 
\footnote{level of statistical significance by Kruskal-Wallis test}
) in all conditions except MC in Task 2.
% level of statistical significance
% For example, the 3D Mouse interface achieved only a $37.5\%$ success rate on task 3.
% Figure \ref{fig:objective_study} shows the average time consumed for each task across four interfaces.
The KT and MC interfaces consistently demonstrate lower task completion times than KB and 3M ($p < 0.05$) particularly for Task 3 and Task 4, indicating higher efficiency.
% These results align with the observed lower success rates for these interfaces, highlighting their inefficiency in time-critical tasks.
The result of subjective result is shown in Fig. \ref{fig:user_study_result}.
The KT and MC interfaces consistently receive significant ($p < 0.05$) higher scores than KB and 3M across all criteria, indicating positive user perception and ease of use.
% In contrast, the Keyboard and 3D Mouse interfaces receive significantly lower ratings, particularly in intuitiveness and efficiency, reflecting the participants' difficulties in using these methods to control the robot.
This study demonstrates that IRIS outperformed baseline interfaces in both objective and subjective metrics,
indicating it provides a more intuitive and efficient approach for data collection.

\input{image/objective_study}

\subsection{Policy Evaluation in Simulation}

\input{table/libero_data_policy_evaluation}

\textbf{General Manipulation}
To evaluate the quality of data collected using IRIS, we employ two standard imitation learning algorithms: BC-Transformer \cite{jia2024towards} and BESO \cite{reuss2023goal}. These models are trained separately on datasets collected with IRIS and on the original LIBERO dataset, with results shown in Figure~\ref{fig:libero_policy_evaluation_fig}.
To ensure a fair comparison, we collect the same number of trajectories using IRIS as in LIBERO, using only the MC interface (instead of KT), since LIBERO operates in Cartesian space rather than joint space. Each model is trained for 50 epochs with three random seeds to capture performance variance, and all experiments use identical training parameters.
% The results demonstrate that models trained on IRIS-collected data achieve comparable performance to those trained on the original dataset.
The results demonstrate that the quality of data collected by IRIS is on par with the original LIBERO dataset.

\textbf{Deformable Objects}
\label{sec:deformable_manipulation_experiment}
% These collected demonstrations can then be used to train models for recorded tasks.
% Example results include \textit{Fold Cloth}: $0.97 \pm 0.018$, \textit{Lift Teddy}: $0.90 \pm 0.035$, and \textit{Stow Teddy}: $0.85 \pm 0.053$.
We also evaluate the data collected by IRIS from deformable object manipulation.
Three tasks were designed to evaluate the data including \textit{Fold Cloth}, \textit{Lift Teddy}, and \textit{Stow Teddy}.
The policy used in this experiment is the U-Net diffusion model \cite{chi2024diffusionpolicy}.
% with reduced down dimension sizes of $[256, 512, 1024]$,
% and a diffusion step embedding dimension size of $256$.
Observations are robot EEF pose, depth, and image data.
The success rate of each task are \textit{Fold Cloth}: $0.97 \pm 0.018$, \textit{Lift Teddy}: $0.90 \pm 0.035$, and \textit{Stow Teddy}: $0.85 \pm 0.053$. 
% The details of this experiment are in the Appendix. \ref{app:deformable}.

\textbf{Dynamic Task Data Collection}
\label{sec:table_tennis_evaluation}
\input{table/table_tennis}
To evaulate the data quality of highly dynamic task collected by IRIS,
this paper uses the table tennis from Fancy\_Gym \cite{fancy_gym} by motion controllers.
The observation includes bat proprioceptive state and dual camera images,
and the action is the desired bat position and orientation in task space. 
Fig. \ref{fig:table_tennis_figure} shows the performance of imitation learning models \cite{reuss2024multimodal, zhou2025beast} trained on the collected data,
using ball interception rate and successful return rate as evaluation metrics.
% The trained policy should first reach the incoming ball with natural stroke movement as human players and then hit the ball to opposite side table.
% The details are in the Appendix. \ref{app:table_tennis_evaluation}
% From all the three experiments,
% this paper conducted three experiments of the possible application in the Sec. \ref{sec:system_application}
% to valid the data collection in Simulation.
% With result of the user study (\ref{sec:user_study}),
% the data collected by IRIS have a compariable quality,
% but with more efficient data collection.
These experiments validate the data from three data collection scenarios in
Section \ref{sec:system_application}.
The results demonstrate that IRIS collects data of comparable quality to traditional methods, while offering significantly greater efficiency.

% In this study, we evaluated two manipulation tasks: Cup Inserting and Pick Up Lego. In the Cup Inserting task, the goal was to insert a smaller cup into a larger one to successfully complete the task. In the Pick Up Lego task, the objective was to pick up a Lego piece and place it into a cup. For each task, we collected 30 demonstrations using both the Tele-operation (Tele-op) interface and the IRIS system. Using these collected demonstrations, we trained two separate policies with our proposed X model—one trained on Tele-op data, and the other on IRIS data. The resulting policies were then evaluated on 40 trials per task.

\input{table/real_robot_experiment}

\subsection{Real World Evaluation}
% \subsection{Policy Evaluation in Simulation}
This experiment evaluates the effectiveness of IRIS for real-world data collection through two tasks: Cup Inserting and Picking Up Lego.
IRIS was compared against Tele-Op, a widely used method for real robot data collection.
In Tele-Op \cite{jia2024towards, opentelevision, openteach}, users observe the scene via a monitor or from a distance, which is less comfortable and narrows the operator’s field of view.
For each task, 30 successful demonstrations were collected from both methods.
These two datasets were used to train two BC-Transformer policies with identical hyperparameters.
The metrics include data collection success rate (the percentage of successful attempts during data collection) and policy success rate after training.
The results (Fig. \ref{fig:combined} b) show that IRIS achieves a higher data collection success rate, and that policies trained with IRIS exhibit better quality than those trained with Tele-Op.
% The details of this experiment is in the Appdenix. \ref{app:real_robot_experiment}.

\input{image/soft_and_real_experiments}

% \textcolor{red}{should be a table of performance here}

%% file: table/success_rate_table.tex
% \begin{table*}[t]
%     \centering
%     \newcolumntype{Y}{>{\centering\arraybackslash}X}
%     \begin{tabularx}{0.7\textwidth}{lXXXX}
%         \hline
%         \toprule
%         Interface & \textbf{Task 1} & \textbf{Task 2} & \textbf{Task 3} & \textbf{Task 4} \\
%         \midrule
%         KB & $0.90$ & $0.725$ & $0.750$ & $0.500$ \\
%         3M & \textbf{$1.00$}  & $0.950$ & $0.375$ & $0.900$ \\
%         KT (Ours)       & \textbf{$1.00$}  & \textbf{$1.00$}  & \textbf{$0.975$} & $0.950$ \\
%         MC (Ours)      &\textbf{ $1.00$}  & $0.900$ &\textbf{ $0.975$} & \textbf{$1.00$}  \\
%         \bottomrule
%     \end{tabularx}
%     \caption{
%     Success rate of four different interfaces.
%     KB and 3M represents Keyboard and 3D Mouse control interface from LIBERO.
%     KT and MC are two control interface used by IRIS.
%     }
%     \label{tab:success_rate}
% \end{table*}

\begin{table*}[t]
    \centering
    \newcolumntype{Y}{>{\centering\arraybackslash}X}
    \begin{tabularx}{0.7\textwidth}{lYYYY}
        \toprule
        Interface & \textbf{Task 1} & \textbf{Task 2} & \textbf{Task 3} & \textbf{Task 4} \\
        \midrule
        KB (LIBERO) & 0.90 & 0.725 & 0.750 & 0.500 \\
        3M (LIBERO) & \textbf{1.00} & 0.950 & 0.375 & 0.900 \\
        KT (Ours) & \textbf{1.00} & \textbf{1.00} & \textbf{0.975} & 0.950 \\
        MC (Ours) & \textbf{1.00} & 0.900 & \textbf{0.975} & \textbf{1.00} \\
        \bottomrule
    \end{tabularx}
    \vspace{-0.2cm}
    \caption{Success rate of four interfaces. KT and MC lead to higher success rate across four tasks.
    % KB and 3M represent Keyboard and 3D Mouse control interfaces from LIBERO. KT and MC are two control interfaces used by IRIS.
    }
    \label{tab:success_rate}
    \vspace{-0.3cm}
\end{table*}

%% file: image/objective_study.tex
\begin{figure}[t!]
    \centering
    \begin{subfigure}[b]{0.8\textwidth}
        \centering
        \includegraphics[width=\textwidth]{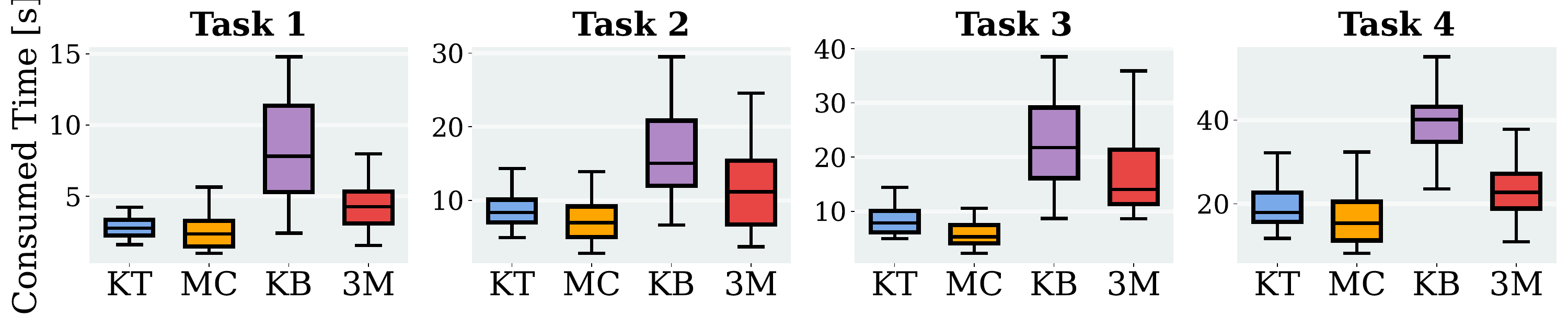}
        % \caption{}
        % \label{fig:objective_study}
        \vspace{-1em} % Reduce space below caption
    \end{subfigure}
    % \hfill
    \begin{subfigure}[b]{0.8\textwidth}
        \centering
        \includegraphics[width=\textwidth]{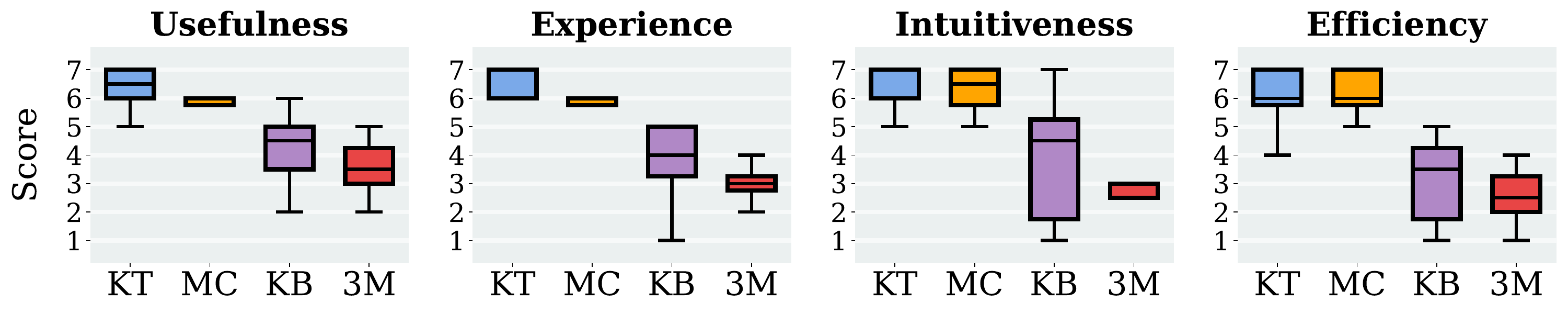}
        % \caption{}
        % \label{fig:subjective_study}
    \end{subfigure}

    \caption{\textbf{The first row}: average task completion time for each interface  across tasks. \textbf{The second row}: subjective evaluation scores of four metrics.
    Our interfaces, KT and MC in XR setting, are faster in data collection with better user experience.
    }

    % \caption{(a) This graph shows the average task completion time (in seconds) for each interface across tasks. The KT and MC interfaces consistently perform more efficiently, while the Keyboard and 3D Mouse interfaces result in longer completion times, particularly on more complex tasks. (b) Subjective evaluation scores for usefulness, experience, intuitiveness, and efficiency across four interfaces. The KT and MC interfaces perform favorably in all categories, while the Keyboard and 3D Mouse interfaces receive lower ratings, particularly in intuitiveness and efficiency.}
    
    \label{fig:user_study_result}
    \vspace{-0.6cm}
\end{figure}

%% file: table/libero_data_policy_evaluation.tex
% \begin{table*}[htbp]
%   \centering
%   \begin{tabularx}{1.0\textwidth}{c|XXXXX}
%     \hline
%     \toprule
%     \multirow{1}{*}{Policy}
%     & Dataset & Task1 & Task2 & Task3 & Task4 \\
%     \midrule
%     \multirow{2}{*}{\textbf{BC-Transformer}}
%     &\textit{LIBERO} & $\mathbf{0.48\pm0.01}$  & $\mathbf{0.48\pm0.01}$ & $\mathbf{0.46\pm0.01}$ & $0.45\pm0.01$ \\
%     &\textit{IRIS} & $0.46\pm0.02$ & $0.43\pm0.03$ & $0.43\pm0.01$ & $\mathbf{0.50\pm0.03}$ \\
%     \midrule
%     \multirow{2}{*}{\textbf{BESO}}
%     & \textit{LIBERO} & $\mathbf{0.48\pm0.01}$ & $0.45\pm0.04$ & $0.46\pm0.04$ & $0.43\pm0.05$ \\
%     & \textit{IRIS} & $0.47\pm0.01$ & $\mathbf{0.50\pm0.03}$ & $\mathbf{0.47\pm0.01}$ & $\mathbf{0.50\pm0.01}$ \\
%     \bottomrule
%   \end{tabularx}
%   \caption{Transcription of the hand-drawn table}
%   \label{tab:libero_policy_evaluation}
% \end{table*}

\begin{figure}[htbp]
\centering
\includegraphics[width=0.8\textwidth]{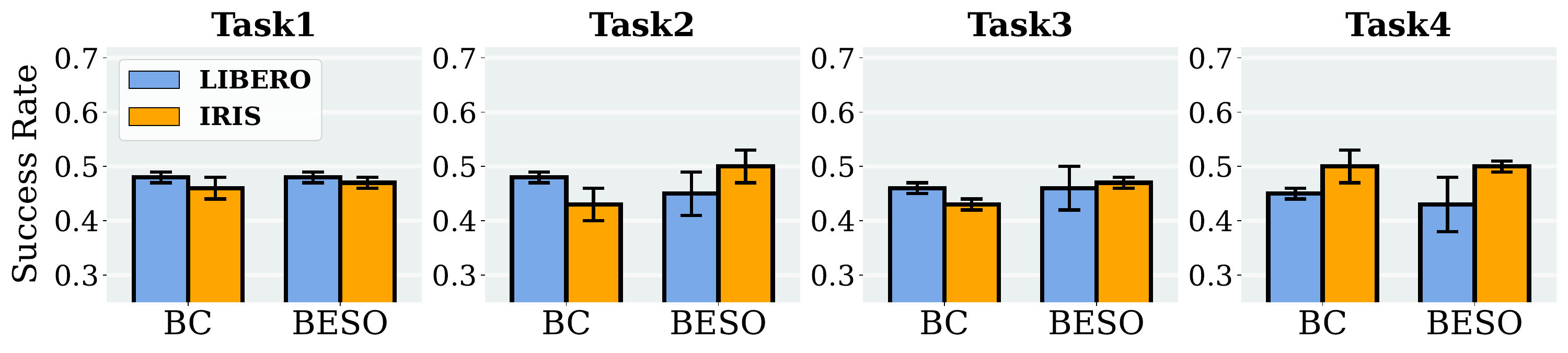}
\caption{Performance comparison of policies trained on different datasets across LIBERO tasks}
\vspace{-0.3cm}
\label{fig:libero_policy_evaluation_fig}
\end{figure}

%% file: table/real_robot_experiment.tex
% \begin{table}[h]
% \centering
% \renewcommand{\arraystretch}{1.3}
% \begin{tabular}{|c|c|c|c|c|c|}
% \hline
% \multirow{2}{*}{\textbf{Tasks}} & \multicolumn{2}{c|}{\textbf{Success Rate}} & \multicolumn{2}{c|}{\textbf{Policy Success Rate}} \\
% \cline{2-5}
%  & \textbf{Tele-op} & \textbf{IRIS} & \textbf{Tele-op} & \textbf{IRIS} \\
% \hline
% Cup Inserting & 96.77 & 100.00 & 80.00 & 85.00 \\
% \hline
% Pick Up Lego & 88.23 & 96.77 & 65.00 & 82.50 \\
% \hline
% \end{tabular}
% \caption{Success rates and policy success rates for the Cup Inserting and Pick Up Lego tasks. Success rates indicate the proportion of successful demonstrations collected via each interface (Tele-operation and IRIS), while policy success rates reflect the performance of the trained policies when evaluated on the corresponding tasks.}
% \label{tab:real_robot_exp_table}
% \end{table}

\begin{figure}[htbp]
\centering
\begin{subfigure}[b]{0.42\textwidth}
\includegraphics[width=\textwidth]{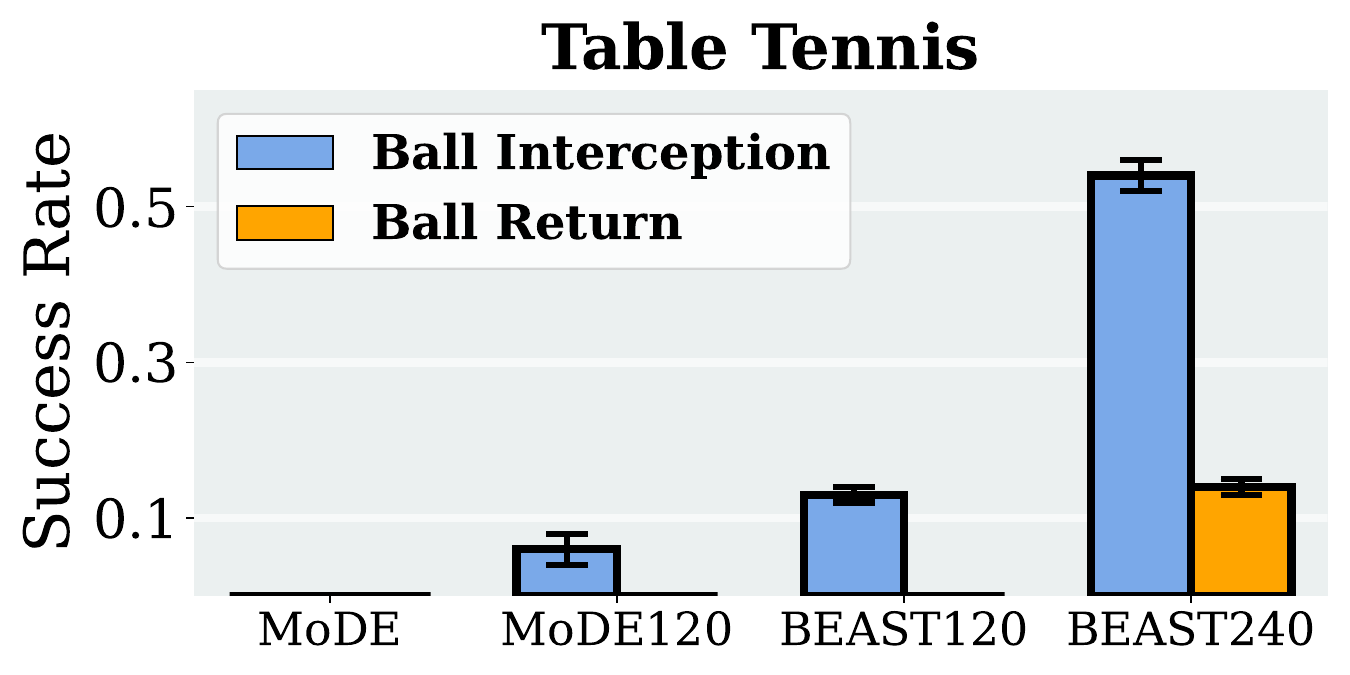}
\caption{Policy Performance of Table Tennis Task}
\label{fig:table_tennis_figure}
\end{subfigure}
\hfill
\begin{subfigure}[b]{0.46\textwidth}
\includegraphics[width=\textwidth]{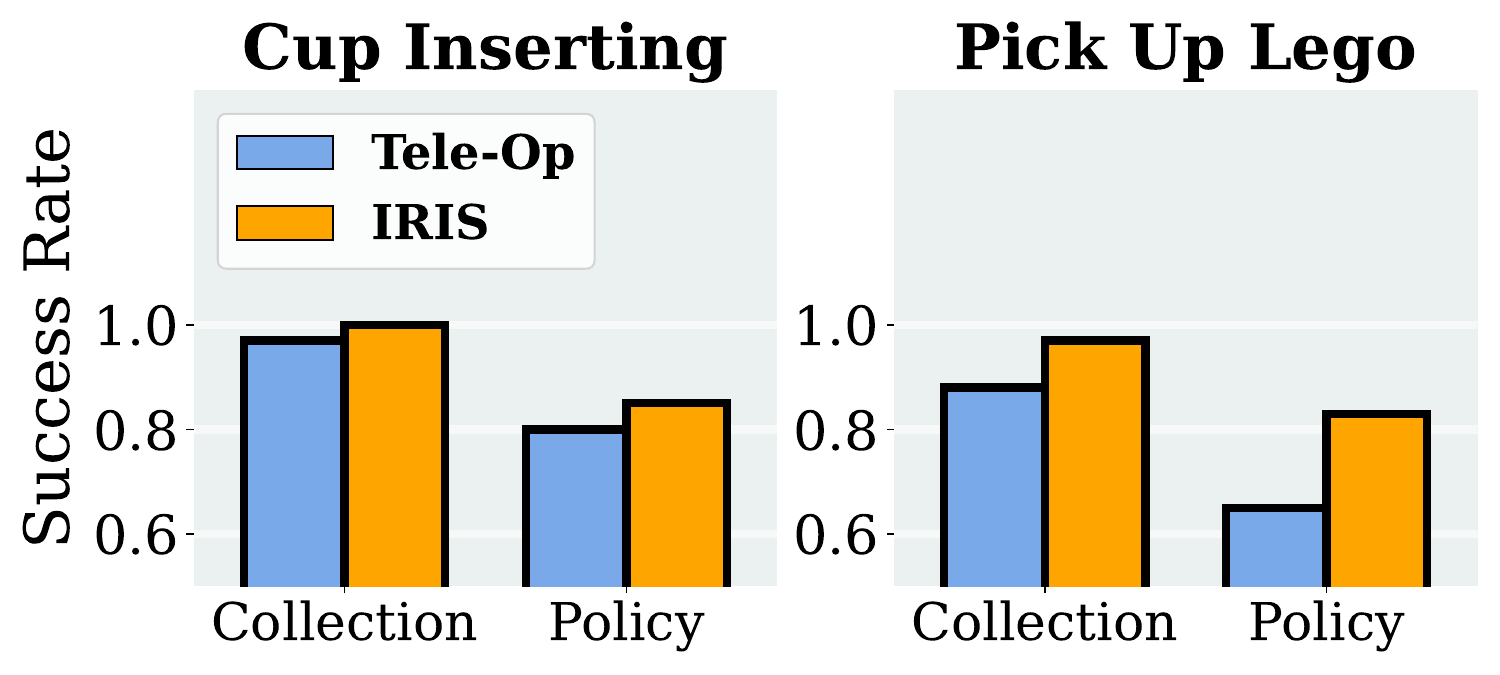}
\caption{Policy Performance of Real World Experiment}
\label{fig:real_robot_exp_figure}
\end{subfigure}
\caption{Performance evaluation of policies trained on IRIS-collected data across diverse scenarios}
\vspace{-0.5cm}
\label{fig:combined}
\end{figure}

%% file: image/soft_and_real_experiments.tex
\begin{figure}[htbp]
\centering
\begin{subfigure}[b]{0.37\textwidth}
\includegraphics[width=0.9\textwidth]{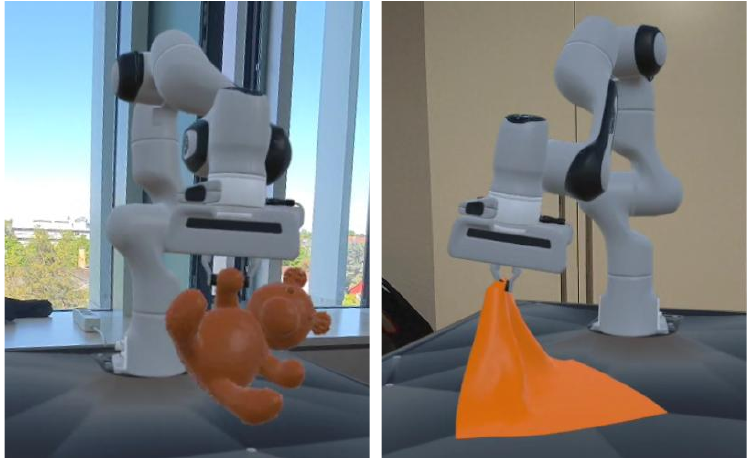}
\caption{Soft object manipulation in IsaacSim}
\label{fig:soft_body_experiment_figure}
\end{subfigure}
\hfill
\begin{subfigure}[b]{0.612\textwidth}
\includegraphics[width=0.9\textwidth]{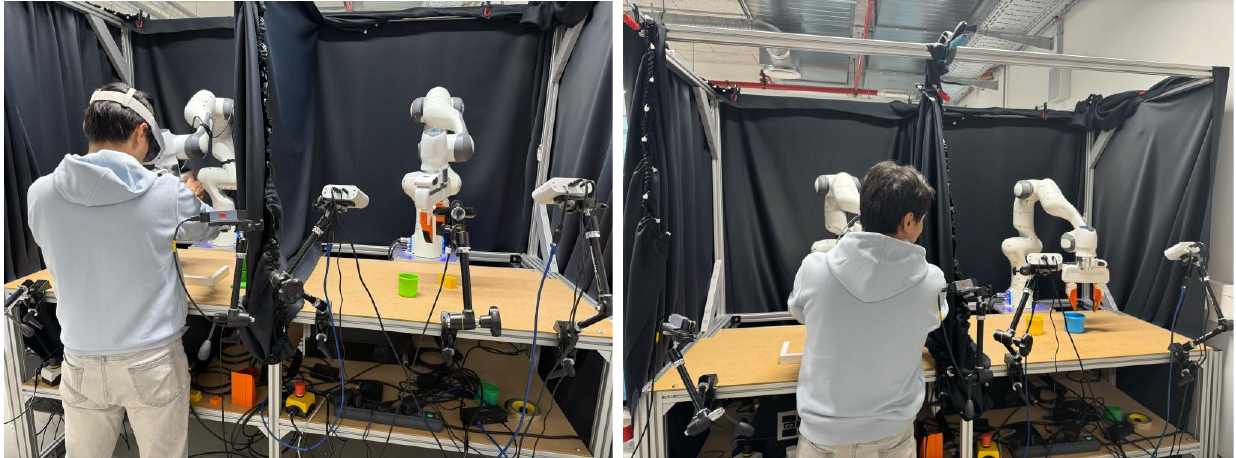}
\caption{Real-world data collection by IRIS(left) and Tele-Op(right)}
\label{fig:real_world_experiment_figure}
\end{subfigure}
\caption{The experiment of deformable manipulation and real-world data collection}
\vspace{-0.6cm}
\label{fig:soft_and_real_experiment_figure}
\end{figure}

%% file: sections/06_limitation_and_conclusion.tex
% The visualization effect is not perfect in XR headsets, since we simply used RGB texture for the materials,
% making the robots from different simulators look the same.
% There should be some rendering configuration in the simulator but we only use RGB texture from them.
% Secondly, IRIS's performance is restricted by the number of simulated objects.
% When the quality of simulated objects in the simulation exceeds some limitation. the performance will be significantly reduced.

% Although the IRIS have 

\section{Conclusion}

% \textcolor{red}{TODO}

% In this work, we introduced IRIS, an innovative Immersive Robot Interaction System that bridges the gap between Extended Reality (XR) technologies and robotics data collection. By addressing the challenges of reproducibility and reusability prevalent in current XR-based systems, IRIS offers a flexible and extendable framework that can be used in multiple simulators, benchmarks, real-world scenarios, and multiple-user use case.
% The user study show that IRIS outperforms previous benchmark data collection method, making it as a potential option for future data collection pipeline.
% The IRIS codebase is open-source, and it will facilitate further research and make IRIS adapted to more use case and more hardware platform.

In this work, we introduced IRIS, an innovative framework that seamlessly integrates Extended Reality (XR) technologies with robotics data collection. IRIS addresses key challenges in reproducibility and reusability that are common in current XR-based systems. Its flexible, extendable design supports multiple simulators, benchmarks, real-world applications, and multi-user use cases.
User studies demonstrate that IRIS outperforms previous data collection methods, positioning it as a promising solution for future data collection pipelines.
As an open-source project, IRIS codebase promotes further research and adaptation across diverse use cases and hardware platforms.

%% file: sections/07_limitations.tex
\section*{Limitations}

Based on our development and usage experience, IRIS has three main limitations.
First, the visualization in XR headsets is limited to basic RGB textures for materials.
IRIS currently supports deformable objects exclusively in IsaacSim, as other simulators either lack robust deformable object capabilities or haven't implemented them.
This paper selected IsaacSim for initial testing due to its superior deformable object simulation.
Finally, IRIS has only been tested on Meta Quest 3 and HoloLens 2,
and further evaluation on additional devices would be beneficial.

%% file: appendix/all_the_appendix.tex
\input{appendix/00_Appendix_Technical_Details}

\input{appendix/01_Appendix_Application_Details}

\input{appendix/02_Appendix_Experiment}

%% file: appendix/00_Appendix_Technical_Details.tex
\appendix
\section{IRIS Framework}

% In this section,
% We separate the physical simulation and rendering into a simulation PC and MR headsets
% ScenePublisher

\subsection{Node Communication Protocol}

The IRIS system operates across simulation and/or sensor-processing computers, multiple XR headsets, and other monitoring and control programs, requiring a robust and reliable network connection between them.
All devices are part of the same subnet, with all the devices connected via Wi-Fi or cable.
Different from Robot Operating System (ROS \cite{quigley2009ros}), 
IRIS leverages a local network across multiple hosts instead of using a single host, 
while using both Request-Response and Publish-Subscribe patterns for data transmission.
This protocol follows a master-node architecture, where the simulation PC serves as the master node, and the XR headsets and other devices act as XR nodes.
Communication is achieved through a combination of UDP sockets and ZeroMQ (ZMQ).

\input{image/protocol}

To ensure node discovery, the master node broadcasts UDP messages at 5 Hz to a fixed broadcast port on the network.
When a new XR node is launched, 
it listens on the broadcast port to receive a broadcast message from the master node. 
%It extracts the master node's details, 
%including its ZMQ socket address and port, to establish a stable ZMQ connection.
Upon receiving the broadcast message, the XR node extracts the master node's details, including its ZMQ socket address and port, to establish a reliable ZMQ connection.
If the master node goes offline, XR nodes continue listening on the discovery port, allowing automatic reconnection when the master node relaunches.
This protocol (Fig. \ref{fig:protocol}) achieves \textbf{Cross-User} ability of IRIS, ensures reliable communication, automatic reconnection, and smooth recovery from disconnections,
making it ideal for dynamic multi-device XR systems.

\subsection{Unified Scene Specification}
% To visualize a scene with arbitrary objects in headsets from the simulation,
% the headsets need to receive the scene model from the simulation and rebuild them.
% The XR application running in the headsets is developed by C\# and Unity,
% and simulation is running by Python or C++.
% From related works, they only support specific robots and assets,
% since they need to create some identical predefined models in a model set for the XR application,
% and they don't support robots or objects which is not in the models set.
% This mechanics highly restrict the flexibility and reusability of their system.
% To fully support all the objects, robots, and assets from simulation,
% we defined a unified scene model format and directly read the scene from simulation.
% Then this information will be sent to the headset by the node communication protocol for rebuilding an identical scene in our XR application.

To visualize a scene with arbitrary objects, the XR headsets need to receive the scene model from the simulation and reconstruct it.
However, the XR application and the simulation run on different devices and use different software architectures (the XR application is developed with C\# and Unity, while simulators might be built in Python or C++).
This makes it impractical to directly transfer the scene from the simulation to the XR environment.
To address this issue, existing solutions rely on predefined models in the XR application for specific robots and assets, requiring a static set of models to be maintained within the application.
This approach restricts flexibility and reusability, preventing the support of robots or objects not included in the model set.
To address this issue, IRIS introduces a novel unified scene specification, which is generated by parsing the scene directly from the simulation.
This specification is subsequently transmitted to the headsets using the node communication protocol, enabling the XR application to accurately and dynamically recreate the scene in real time.

\begin{figure}[t]
    \centering
    \includegraphics[width=0.5\linewidth]{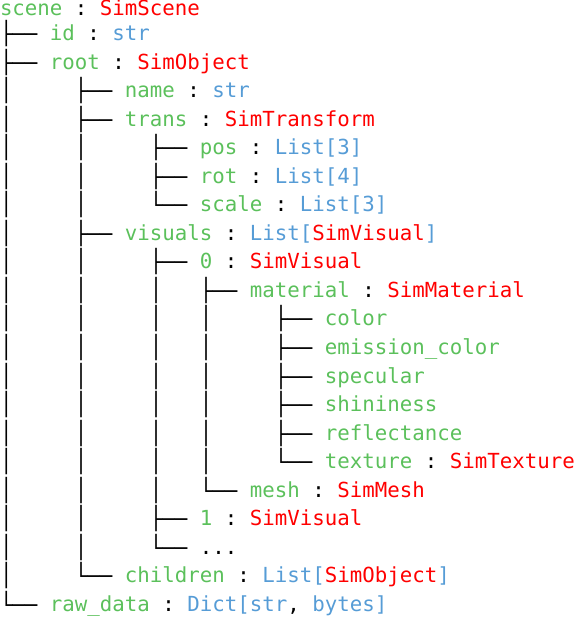}
    \caption{
The hierarchical structure of the Scene Specification begins with a root SimObject, which contains all objects in the scene. Each SimObject has a name, a list of child SimObjects, and a list of visuals. Each visual represents a geometric element attached to the object.
Within each geometric element, materials define properties such as color and texture, while meshes determine the shape. The scene's raw data includes the byte streams of meshes and textures. Since these streams are extensive, they are sent to XR headsets sequentially after the initial scene specification is transmitted.
}
    \label{fig:scene_specification}
\end{figure}

% The unified scene specification includes environment setting (e.g., lighting) and all the objects as well as their geometry elements, meshes, materials and textures attached to them.
% IRIS provides a Python Library named ScenePublisher for parsering a simulation scene to a scenespevification.
% In the scene specification, geometry element is the shape of object (e.g., Cube, sphere, capsule, cylinder, or mesh).
% The mesh contains the vertex list, faces list, normals list, and texture coordinate list.
% material is the color and eemissionColor, reflectance or texture attached to them.
% Texture is an image which repesetnt texture.
% The objects is stacked by the kinematic tree and they will be parsed into json format by the ScenePublisher.
% Since the mesh and texuture contain too much data, and they will be sent to clients at beginning because it is rather big.
% So the visual and material which have mesh and texture will hold an hash code, and wait the request from XR node to retrieve it later.

The unified scene specification (shown in Fig. \ref{fig:scene_specification}) includes all objects with their geometry, meshes, materials, and textures. 
IRIS provides a Python library called \textit{ScenePublisher} to parse a simulation scene into this scene specification.
In the specification, 
geometry defines the object's shape (e.g., cube, sphere, capsule, cylinder, or mesh).
The mesh contains vertex lists, face lists, normals, and texture coordinates. 
Materials define surface properties, including color, emission color, reflectance, and attached textures, 
and the texture is an image representing the object's surface appearance.
Objects are organized using a kinematic tree structure and are serialized into JSON by the \textit{ScenePublisher}.

Since meshes and textures contain large amounts of data, 
geometry and materials store only their hash code to reduce the transmission load. 
The XR application rebuilds the scene upon receiving the kinematic tree and then requests the meshes and textures from the simulation server in byte format.
In some cases, textures can be quite large (e.g., the textures for the RoboCasa \cite{nasiriany2024robocasa} scene exceed 700 MB). To ensure the scene loads within an acceptable time for users, we compress the textures. This compression reduces the loading time to a few seconds.
Afterwards, the \textit{ScenePublisher} continually acquires simulation states and forwards them to the XR headset. This way the positions and rotations of all the objects are updated at a fixed frequency.

The scene specification enables IRIS to support a wide range of robots and objects in simulation, facilitating both \textbf{Cross-Scene} and \textbf{Cross-Embodiment} capabilities.

\subsection{Extendable and Flexible Framework Support}
The unified scene specification is a general definition that does not rely on any specific simulator, providing an extensible mechanism for scene loading and updating. IRIS can be easily adapted to various simulation engines and frameworks by implementing a new simulation parser to generate the unified specification from the simulation scene and a new publisher to update the states of scene.

Currently, IRIS supports scene parsers for MuJoCo, IsaacSim, CoppeliaSim, and Genesis, with the potential to be extended to other simulation engines as desired.
IRIS has been tested in some MuJoCo-based benchmarks including \textbf{Meta World} \cite{yu2020meta}, \textbf{LIBERO} \cite{liu2024libero}, \textbf{RoboCasa} \cite{nasiriany2024robocasa}, \textbf{robosuite} \cite{zhu2020robosuite}, \textbf{Fancy Gym} \cite{fancy_gym}, and CoppeliaSim-based benchmark like 
\textbf{PyRep} \cite{james2019pyrep},
\textbf{Colosseum} \cite{pumacay2024colosseum}.
This demonstrates that IRIS can be easily adapted to various benchmarks and simulators, highlighting its \textbf{Cross-Simulator} capability.

IRIS provides a user-friendly API.
For each environment or framework,
a single line of code suffices to visualize and update the simulation in the XR headset.
Here is a short example of how to use it in the MuJoCo Simulation:
\begin{lstlisting}[language=Python]
# import scenepub
from scenepub.sim.mj_publisher import MujocoPublisher
# define the mujoco environment
model = mujoco.MjModel.from_xml_path(xml_path)
data = mujoco.MjData(model)

# define the ScenePublisher for mujoco; only this line needs to be added
publisher = MujocoPublisher(model, data, host)

# run simulation
while True:
    # run simulation step logic
    pass
\end{lstlisting}

The MujocoPublisher instance only needs to access the model and data from the MuJoCo simulation. It then creates a separate thread to run the communication protocol, automatically connecting and communicating with all available XR headsets.
IRIS provides various Simulation Publishers for different environments, all with a consistent and easy-to-use interface. This mechanism ensures a seamless experience, making the system very user-friendly.

\subsection{Real Scene Loading}
\label{sec:real_world_teleop}
% The loading and updating of real scene is similar to the simulation.
% IRIS use cameras and calibrate them for merging multiple point clouds into one point cloud.
% Then the point cloud will be cropped and down-sampled,
% each point will be sent to headsets and IRIS use particle system in Unity to visualize and update the point cloud.

% We merge and down-sample the point clouds from multiple cameras by first applying extrinsic transformations to each point cloud. 
% Then, we map the XYZ coordinates of the both pointclouds to voxel-grid indices.
% Next, we blend the colors of all points within a voxel and compute the voxel's centroid XYZ coordinates, returning both as output.
% This entire process uses Thrust \cite{bell2012thrust} and runs on the GPU to achieve low latency.

% The loading and updating of real-world scenes in IRIS follows a process similar to that used for simulation scenes, making the feature of Cross-Reality.
The loading and updating of real-world scenes in IRIS follow a process similar to that of simulation scenes, demonstrating its \textbf{Cross-Reality} capability.
It processes point clouds from one or multiple RGB-D cameras, which are extrinsically calibrated to a fiducial marker in the scene. A point cloud processor applies the extrinsic transformation to each point cloud before merging them. The merged point cloud is then cropped and downsampled using a voxel-grid filter.
This filter maps all 3D points to voxel-grid indices, blends the colors of points within the same voxel, and calculates the voxels centroid.
The output is a reduced point cloud that retains both color and position data.
To ensure low latency, this process runs on the GPU using Thrust \cite{bell2012thrust}.
Finally, the processed point cloud is transmitted to the headsets, where IRIS uses a particle system in Unity to visualize and dynamically update the scene in real-time.

\section{Simulator Support and Real-world Data Collection}

\subsection{Mujoco}
\label{app:mujoco}

MuJoCo \cite{todorov2012mujoco} is a fast, accurate physics engine ideal for simulating robots with complex joint structures and contact dynamics.
It supports advanced robot simulations with features like customizable actuators, collision detection, and friction modeling, enabling realistic testing of robotic control, manipulation, and learning algorithms in dynamic environments.

When starting a MuJoCo simulation, MuJoCo provides a model instance and a data instance.
The scene specification of IRIS can be generated from the model, while the simulation states can be retrieved from the data instance.
The model instance contains all the necessary assets for the scene specification, including meshes, textures, and materials. 
These assets are stored as NumPy arrays and need to be converted to byte streams with data types such as \textit{float32} or \textit{int8} for transmission.

MuJoCo uses a standard robotic coordinate system, where X is forward, Y is left, and Z is up. This differs from Unity's coordinate system, where X is right, Y is up, and Z is forward.
Therefore, object transforms, mesh vertices, faces, and normal data must be adapted for Unity using the following code:

\begin{lstlisting}[language=Python]
def mj2unity_pos(pos: List[float]) -> List[float]:
    return [-pos[1], pos[2], pos[0]]

def mj2unity_quat(quat: List[float]) -> List[float]:
    return [quat[2], -quat[3], -quat[1], quat[0]]
\end{lstlisting}

Since MuJoCo includes visual group settings that specify which object groups can be visualized, IRIS provides an API to support this feature through the \textit{MujocoPublisher} definition, as shown below:

\begin{lstlisting}[language=Python]
class MujocoPublisher(ScenePublisher):
    def __init__(
        self,
        mj_model,
        mj_data,
        host: str = "127.0.0.1",
        no_rendered_objects: Optional[List[str]] = None,
        no_tracked_objects: Optional[List[str]] = None,
        visible_geoms_groups: Optional[List[int]] = None,
    ) -> None:
\end{lstlisting}

\subsection{IsaacSim}

IsaacSim~\cite{nvidia_isaac_sim} is a robotics simulation environment based on the NVIDIA Omniverse platform~\cite{nvidia_omniverse}.
Several frameworks \cite{mittal2023orbit,gong2023arnold} are built on top of it, sharing the same underlying data structures.
IsaacSim supports ray-tracing for realistic rendering, and batched physics simulation on GPUs, significantly accelerating the training of models.

\begin{figure}[h]
    \centering
    \includegraphics[width=0.5\linewidth]{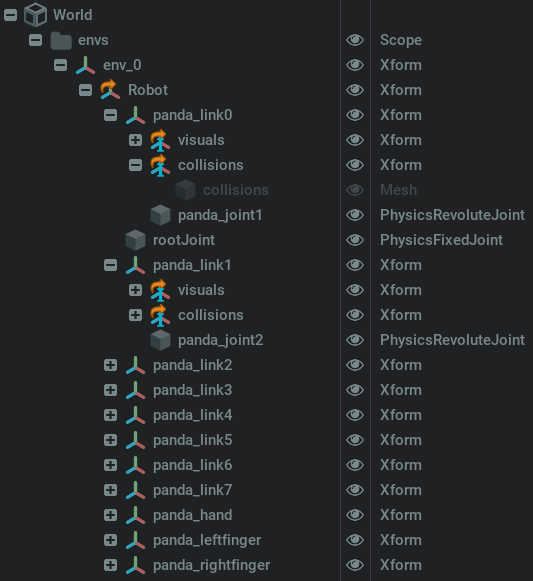}
    \caption{An example of USD scene hierarchy for Franka Panda robot arm in IsaacSim.}
    \label{fig:isaacsim-tree}
\end{figure}

The scenes in IsaacSim are organized in a Universal Scene Description (USD) format, and data can be accessed directly with the OpenUSD API \cite{pixarUSD}.
The scene hierarchy consists of transformations (Xform), meshes, articulations (joints) among other elements. Fig.~\ref{fig:isaacsim-tree} illustrates a sample scene hierarchy.

Besides the low-level OpenUSD API, IsaacSim also provides some utility APIs for accessing scene data. In our system, we use a mix of both APIs. A (simplified) code snippet for accessing parsing the tree structure of the USD scene is given below. We use the OpenUSD API \texttt{GetChildren()} to retrieve child primitives here.

\begin{lstlisting}[language=Python]
# compute local transforms of the current prim
trans, rot, scale = self.compute_local_trans(root)
# parse material and geometry
mat_info = self.parse_prim_material(prim=root)
self.parse_prim_geometries(
    prim=root,
    prim_path=prim_path,
    sim_obj=sim_object,
    mat_info=mat_info or inherited_material,
)
# parse children of the current prim
for child in root.GetChildren():
    if obj := self.parse_prim_tree(
        root=child,
        parent_path=prim_path,
        inherited_material=mat_info or inherited_material,
    ):
        sim_object.children.append(obj)
\end{lstlisting}

Here is another snippet for computing the world transform of a primitive in the hierarchy. Instead of computing the world transform with OpenUSD API, the class in IsaacSim API \texttt{XFormPrim} is used to simplify the code.

\begin{lstlisting}[language=Python]
from pxr import Usd
from omni.isaac.core.prims import XFormPrim

def compute_world_trans(self, prim: Usd.Prim):
    prim = XFormPrim(str(prim.GetPath()))
    assert prim.is_valid()

    pos, quat = prim.get_world_pose()
    scale = prim.get_world_scale()

    return (
        pos.cpu().numpy(),
        quat_to_rot_matrix(quat.cpu().numpy()),
        scale.cpu().numpy(),
    )
\end{lstlisting}

\subsection{CoppeliaSim}
\begin{figure}[h]
    \centering
    \includegraphics[width=0.5\linewidth]{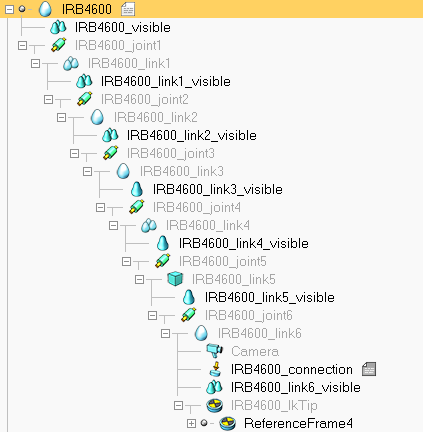}
    \caption{An example of a tree structure for building an ABB IRB4600 manipulator.}
    \label{fig:coppliasim-tree}
\end{figure}

CoppeliaSim, formerly known as V-REP, is a versatile and widely used robot simulation software that supports various physics engine backbones. Scenes in CoppeliaSim are constructed using a tree structure, which includes objects such as visual shapes, dynamic shapes, and joints, as illustrated in Fig.~\ref{fig:coppliasim-tree}.

In our system, we interact with CoppeliaSim using Python APIs. Depending on the version of CoppeliaSim, we utilize two groups of Python API functions:

\textbf{ZeroMQ Remote API} \footnote{https://manual.coppeliarobotics.com/en/apiFunctions.htm}. This is an official API introduced in version 4.4. It facilitates fast and straightforward communication between the simulator and a Python script, enabling users to efficiently build scenarios. For example, to retrieve mesh information for all shape objects in a CoppeliaSim scene, one can use the following call:
\begin{lstlisting}[language=Python, caption=Example ZeroMQ Remote API.]
from coppeliasim_zmqremoteapi_client import RemoteAPIClient
client = RemoteAPIClient()
sim = client.require('sim')

list_vertices, list_indices, list_normals = [], [], []

objects_id_list = sim.getObjectsInTree(sim.handle_scene, 
                                       sim.handle_all, 0)

for idx in objects_id_list:
    if sim.getObjectType(idx) == sim.sceneobject_shape:
        # We assume all the shapes are primary shapes
        vertices, indices, normals = sim.getShapeMesh(idx)
        
        list_vertices.append(vertices)
        list_indices.append(indices)
        list_list_normals.append(normals)
\end{lstlisting}

\textbf{PyRep API}\cite{james2019pyrep} \footnote{https://github.com/stepjam/PyRep}. This is a widely-used third-party Python communication interface under the CoppeliaSim version 4.1. Several notable CoppeliaSim projects and benchmarks, such as RLBench \cite{james2020rlbench}, are built on this interface. Similar to the official interface, PyRep provides various API functions and properties. For instance, to retrieve all object handles from a CoppeliaSim scene, one can use the following call:
\begin{lstlisting}[language=Python, caption=Example PyRep API.]
from pyrep.backend.sim import simGetObjectsInTree
from pyrep.backend.sim import simGetObjectType
from pyrep.backend.sim import simGetShapeMesh
from pyrep.backend.simConst import sim_handle_scene
from pyrep.backend.simConst import sim_handle_all
from pyrep.backend.simConst import sim_object_shape_type

list_vertices, list_indices, list_normals = [], [], []

objects_id_list = simGetObjectsInTree(sim_handle_scene, 
                                      sim_handle_all, 0)

for idx in objects_id_list:
    if simGetObjectType(idx) == sim_object_shape_type:
        # We assume all the shapes are primary shapes
        vertices, indices, normals = simGetShapeMesh(idx)
        
        list_vertices.append(vertices)
        list_indices.append(indices)
        list_list_normals.append(normals)

\end{lstlisting}

\subsection{Genesis}

Genesis \cite{Genesis} is a versatile physics platform designed for robotics, embodied AI, and physical AI applications. It combines a universal, re-engineered physics engine capable of simulating diverse materials and phenomena with a lightweight, ultra-fast, and user-friendly robotics simulation environment. It also features a powerful, photorealistic rendering system and a generative data engine that transforms natural language prompts into multi-modal data. By integrating various physics solvers within a unified framework, Genesis supports automated data generation through a generative agent framework, with its physics engine and simulation platform now open-source and further expansions planned.

IRIS supports Genesis by providing a Genesis parser designed to translate simulation data from the Genesis physics platform into the IRIS framework's internal representation.
It manages this by reading rigid entities, links, and geometries from a \textit{gs.Scene} object and converting them into equivalent IRIS structures such as \textit{SimScene}, \textit{SimObject}, \textit{SimVisual}, and \textit{SimMaterial}. Here's a detailed breakdown of the parsing process and its components:

The main function initializes a \textit{SimScene} and iterates through all entities in the Genesis scene.
It builds a hierarchical scene graph based on parent-child relationships between entities and links. The graph is stored in the hierarchy dictionary, which tracks each object's parent and child nodes.

Each rigid entity (\textit{RigidEntity}) from Genesis is processed to create a \textit{SimObject}. The position and rotation of the entity are retrieved and transformed is similart to Mujoco (\ref{app:mujoco}) to ensure compatibility with Unity's coordinate system.
If the Genesis scene is already built, the function pulls actual position and rotation data; otherwise, it defaults to identity transforms.

Links (RigidLink) represent individual parts of a rigid entity's structure. Each link is converted into a \textit{SimObject} and positioned within the scene based on its parent link's position and orientation.
If the link has associated visual geometries (\textit{vgeoms}), the parser processes each visual element to create \textit{SimVisual} objects, which store geometry and material data. Links without visual elements are simply added to the scene hierarchy without rendering.

Visual geometries (\textit{RigidGeom}) from Genesis are mapped to IRIS visual representations. The parser constructs a \textit{SimVisual} object, including position, rotation, and mesh data.
The mesh generation function extracts vertex and face data from the Genesis mesh and constructs a \textit{SimMesh} object for the IRIS scene, defining both the physical and visual structure of the geometry. Genesis leverages the \textit{Trimesh} \cite{trimeshTrimeshTrimesh} library to handle mesh processing, which simplifies the conversion to the IRIS mesh format.

After processing all entities, the parser organizes the parsed objects into a tree structure. Objects without parents are attached to the root of the scene, while others are linked to their respective parents based on the hierarchy dictionary.

\subsection{Real World}

\input{appendix/real_world_appendix}

%% file: image/protocol.tex
\begin{figure}[ht]
    \centering
    \includegraphics[width=0.7\textwidth]{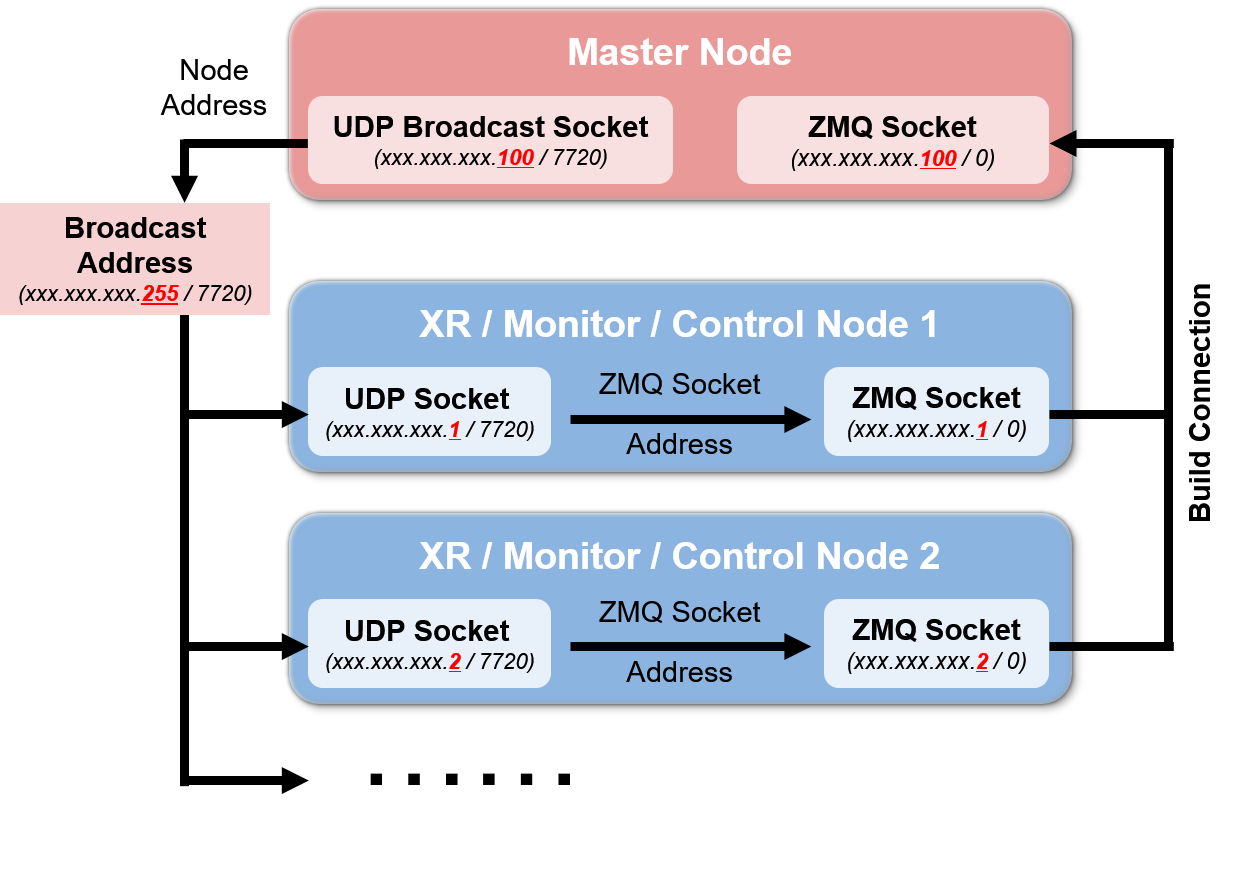}
    \caption{The master node broadcasts UDP messages containing its details to the broadcast address (e.g., \textit{192.168.0.255}) at port 7720. Each IP address in the diagram (e.g., \textit{192.168.0.100/0}) represents a device's unique address on the network, where '0' indicates a dynamically assigned port provided by the operating system. XR nodes, upon startup, listen on the broadcast port (7720) to receive these messages, extract the master node's IP and ZMQ socket address, and build a stable connection. This architecture supports both request-response and publish-subscribe communication patterns, ensuring robust, multi-device connectivity with automatic reconnection capabilities.}
    \label{fig:protocol}
\end{figure}

%% file: appendix/real_world_appendix.tex
The point cloud-based XR teleoperation system enables immersive teleoperation and data collection by allowing users to interact with a real-world robotic setup through an Extended Reality (XR) interface. Unlike conventional video-streaming methods, this approach provides spatial awareness and depth perception by rendering real-time 3D reconstructions of the environment in XR. The system consists of an ORBBEC Femto Bolt depth camera positioned in front of the robot to capture RGB and depth data in real time. A QR code on the robot’s end effector serves as a reference marker for pose estimation, aligning point clouds with the robot’s frame. A C++-based point cloud processing pipeline converts raw data into XYZ-RGB point clouds at 30Hz, generating approximately 2 million points per frame. To optimize computational efficiency, GPU-accelerated voxel grid downsampling is applied before transmission. The processed point clouds are sent to a main process that handles transformation, cropping, and communication with the XR system. A ZeroMQ (ZMQ) messaging protocol is used to enable efficient, low-latency communication between system components, ensuring real-time transmission of point clouds and control commands. A Meta Quest 3 headset renders the processed point clouds in real time, enabling users to visualize and interact with the scene in a fully immersive 3D environment.

\subsubsection{Camera-to-Robot Base Transformation}

\paragraph{Pose Estimation and Homogeneous Transformation}
To ensure accurate spatial alignment, the system transforms the camera’s coordinate frame into the robot’s base coordinate frame using a homogeneous transformation matrix, computed as follows:

\begin{itemize}
    \item The QR code on the robot’s end effector provides a fixed reference point for tracking position and orientation in the camera's space.
    \item Using Forward Kinematics (FK), the end effector’s pose relative to the robot base is determined.
    \item The camera-to-robot base transformation is derived as:
    \begin{equation}
        T_{\mathrm{Robot Base}}^{\mathrm{Camera}} = T_{\mathrm{Robot Base}}^{\mathrm{End Effector}} \cdot T_{\mathrm{End Effector}}^{\mathrm{Camera}}
    \end{equation}
    where:
    \begin{itemize}
        \item $T_{\mathrm{Robot Base}}^{\mathrm{End Effector}}$ is obtained from the robot’s FK.
        \item $T_{\mathrm{End Effector}}^{\mathrm{Camera}}$ is estimated using the QR code tracking system.
    \end{itemize}
\end{itemize}

This ensures that the captured point clouds align precisely with the robot’s coordinate frame, eliminating drift and inconsistencies in XR visualization.

\subsubsection{Point Cloud Processing Pipeline}
\label{appendix:Point Cloud Processing Pipeline}
\paragraph{Raw Data Acquisition}
The ORBBEC Femto Bolt camera captures:
\begin{itemize}
    \item Depth images (encoded as a depth map),
    \item RGB images (color information),
    \item Camera intrinsic parameters (for depth-to-3D conversion).
\end{itemize}

\paragraph{Conversion to XYZ-RGB Point Cloud}
Using the camera intrinsics, each depth pixel is converted into 3D world coordinates $(X, Y, Z)$ using:
\begin{equation}
    X = (u - c_x) \frac{Z}{f_x}, \quad Y = (v - c_y) \frac{Z}{f_y}, \quad Z = D(u,v)
\end{equation}
where:
\begin{itemize}
    \item $(u, v)$ are pixel coordinates,
    \item $(c_x, c_y)$ are the camera's principal point offsets,
    \item $(f_x, f_y)$ are focal lengths,
    \item $D(u,v)$ is the depth value at pixel $(u, v)$.
\end{itemize}
Each $(X, Y, Z)$ point is assigned an $(R, G, B)$ value from the color image, forming the XYZ-RGB point cloud.

\paragraph{Cropping and Filtering}
To reduce noise and retain only relevant portions of the workspace, the system:
\begin{itemize}
    \item Crops unnecessary regions using bounding box constraints,
    \item Applies statistical outlier removal to eliminate noise points.
\end{itemize}

\paragraph{Voxel Grid Downsampling (GPU-Accelerated)}
Since the raw point cloud consists of approximately 2 million points per frame, direct transmission is computationally expensive. To optimize efficiency, voxel grid downsampling is applied, which:
\begin{itemize}
    \item Divides the workspace into 3D voxels,
    \item Averages all points within each voxel to generate a single representative point,
    \item Reduces the point cloud size while preserving structural details.
\end{itemize}

\subsubsection{XR Visualization and Teleoperation}

\paragraph{Real-Time Communication with XR System}
The processed point cloud is transmitted to a main process, which then relays the data to the Meta Quest 3 headset using ZeroMQ. ZeroMQ ensures efficient, asynchronous, and low-latency messaging between the processing node and the XR system, allowing:  
\begin{itemize}  
    \item Real-time rendering of the scene in XR,  
    \item Full 3D immersion with accurate depth perception,  
    \item Reliable transmission of point cloud data and control commands.  
\end{itemize}  

\paragraph{Leader-Follower Teleoperation Framework}
The teleoperation system consists of:
\begin{itemize}
    \item A leader robot controlled by a human wearing the Meta Quest 3 headset,
    \item A follower robot that replicates the leader’s movements, including gripper actions.
\end{itemize}

A networked control architecture ensures:
\begin{itemize}
    \item Low-latency synchronization between the leader and follower,
    \item Seamless remote manipulation, removing the need for physical human presence.
\end{itemize}

\subsubsection{Calibration for XR-Robot Alignment}
To ensure that XR visualization aligns precisely with real-world objects, the system undergoes a calibration process consisting of:
\begin{itemize}
    \item Robot workspace alignment: The follower robot’s workspace is adjusted to match the leader’s XR-rendered environment,
    \item Point cloud adjustment: The camera-to-robot transformation is fine-tuned,
    \item User feedback correction: Users verify object positions in XR and real-world views.
\end{itemize}

\subsubsection{Future Extensions: Multi-Camera and ICP-Based Fusion}
Future implementations will:
\begin{itemize}
    \item Incorporate multiple cameras for wider coverage,
    \item Perform point cloud fusion using Iterative Closest Point (ICP) for improved spatial accuracy,
    \item Enhance depth perception by reconstructing high-fidelity 3D representations of the workspace.
\end{itemize}

%% file: appendix/01_Appendix_Application_Details.tex
\section{Intuitive Robot Control Interface}
\label{app:intuitive_robot_interface}
% \textcolor{red}{should have some images of each interface}

In data collection tasks, robot control interfaces are used to operate the robot in both simulated and real-world environments.
Based on research in teleoperation and robot data collection \cite{jiang2024comprehensive}, Kinesthetic Teaching and Motion Controllers have been identified as the most intuitive and effective control interfaces.
Hence, we ensured that IRIS supports these two methods.
Thanks to IRIS's flexible framework, it is possible to easily customize and implement additional alternative control interfaces,
such as hand tracking, gloves, smartphones, or motion tracking systems.
This adaptability enables tailored solutions to meet specific requirements, enhancing both the usability and versatility of the system for various applications.
Fig. \ref{fig:interface} shows how these two interfaces work in IRIS. The implementation of these interfaces is outlined below.

\input{image/interface/interface}

\subsection{Kinesthetic Teaching}
\label{sec:kt}
% Kinesthetic teaching is an intuitive interface controlling robots by physically moving the real robot.
% The real robot transmit joints position and velocity to the controlled robot in the real time,
% and the controlled robot could be real robot or virtual robot.
% In both real world and simulation data collection,
% the key of Kinesthetic Teaching is the alignment of the real robot with the corresponding virtual robot in the XR headsets,
% so that usrs feel an immersive control. 
% By using spatial anchor, the robot objects in the XR and points of cloud could be perfectly match to the real robot.

Kinesthetic teaching is an intuitive interface that allows users to control robots by physically moving the real robot \cite{wrede2013user, sukkar2023guided, jiang2024comprehensive}.
The real robot transmits joint positions and velocities in real time to the controlled robot, which can be either a virtual or another physical robot.
In both real-world and simulation data collection,
the key aspect of kinesthetic teaching is ensuring alignment between the real robot and its virtual counterpart in the XR headsets.
By utilizing \textit{Spatial Anchors} (\ref{sec:spatial_anchor}), the virtual robot in the XR headsets can be perfectly aligned with the real robot, which provide users with an intuitive and immersive experience.

% In addition, our Kinesthetic Teaching interface provides force feedback for users,
% which means that users feel the resistance force from real robot when controlled robots contacts with objects,
% this feature also applied to real world data and virtual data collection.

\subsection{Motion Controller}

% Motion controllers are widely used in robot data collection.
% \textcolor{red}{some papers}
% They offer stable tracking and a flexible approach for various applications.
% While some XR headsets, such as the HoloLens 2, do not natively support motion controllers, this limitation can be addressed by integrating third-party motion controllers compatible with platforms like SteamVR. \textcolor{red}{citation}
% Motion controllers use inverse kinematics to control robots in Cartesian space. The movement of the robot’s end effector is dictated by the motion controller's trigger. 
% IRIS supports retrieve motion controller data from Meta Quest 3.
% For more complex scenario, users could also design their own controller based on the our IRIS.
% From our usage, motion controllers are only suitable for two-gripper end effectors, and it is challenging to use them with robotic hands to perform complex movements.

Motion controller Interfaces are commonly used in robot data collection and teleoperation \cite{pettinger2020reducing, lin2022comparison}, providing stable tracking and flexibility for various applications. 
Although some XR headsets, such as HoloLens 2, do not natively support motion controllers, this limitation can be resolved by integrating third-party controllers compatible with platforms such as SteamVR \cite{steampoweredSteamVR}.
Motion controllers utilize inverse kinematics to control robots in Cartesian space, with the movement of the robot’s end effector controlled by the controller's trigger. 
IRIS supports retrieval of motion controller data from devices like the Meta Quest 3. 
For more complex scenarios, users can design custom controllers using IRIS' flexible framework.
% Based on our experience, motion controllers are well-suited for two-gripper end effectors, 
% but are less effective for controlling robotic hands in tasks requiring complex movements.

% \subsubsection{Hand Tracking}

% Hand tracking is an intuitive method for controlling robots, especially humanoid robots or robotic arms with hand-like end effectors.
% The inside-out hand tracking (HT) interface uses the cameras of XR headsets to track hand movements and recognize gestures, 
% which is widely supported across different types of XR devices.
% This method is particularly useful for data collection tasks, as it closely mimics robotic hand movements, providing a natural and immersive control experience.
% \textcolor{red}{not sure about this part}
% IRIS provides an easy-to-use Python API (\ref{sec:interaction_data_collection}) to access hand tracking data. 
% Although this interface has only been tested with two-finger end effectors rather than humanoid robots. 
% users can easily leverage IRIS to develop custom controllers for their own robotic systems.
% However, hand tracking has some limitations.
% It requires the user's hands to stay within the headset's field of view, and tracking can be disrupted by occlusions, poor lighting, or fast hand movements, reducing its reliability for precise control.

% \textcolor{red}{should be a table of interfaces here}

\subsection{Affiliated Monitor Tools}
\begin{figure}[h!]
    \centering
    \includegraphics[width=1.0\textwidth]{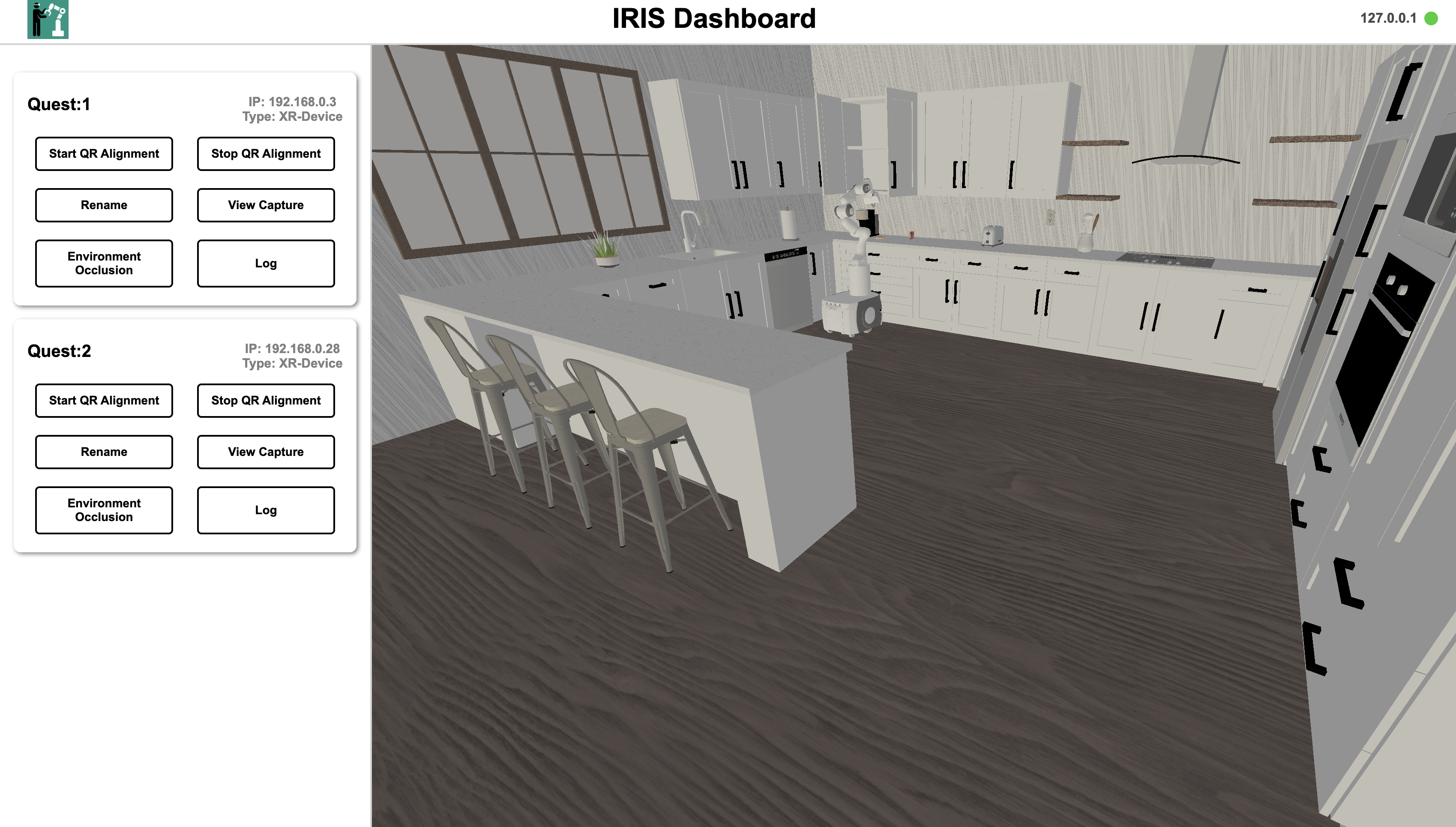}
    \caption{
The IRIS Dashboard, accessible via a web interface, 
allows the control and monitoring of all connected nodes. 
The scene streamed by IRIS is rendered on the right-hand side. 
The left panel displays the connected XR devices, 
providing an interface through which users can control all services made available by each device.
    }
    \label{fig:webapp}
\end{figure}

The extensibility of IRIS opens up numerous possibilities for creating new applications. IRIS includes a web-based monitoring tool for managing all XR headsets. This tool allows users to easily start and stop alignment processes, as well as rename devices.
Additionally, the tool supports real-time scene visualization using \textit{three.js} \cite{threejsThreejsDocs}, using the unified scene specification. An example screenshot is shown in Fig. \ref{fig:webapp}.

\subsection{Spatial Anchor}
\label{sec:spatial_anchor}
% \textcolor{red}{rewrite this part}

% Spatial anchor serves as a reference in a 3D environment to accurately position virtual objects within physical space. 
% It allows virtual objects to maintain their position, orientation, and alignment even as users move around or leave and return to the environment.
% IRIS use QR code or motion controller (only for Meta Quest 3) as spatial anchor
% Alignment between the virtual environment and the physical world is achieved by using a trackable QR code.
% The QR code serves as the coordinate frame of the world for the virtual environment,
% and users can provide the offset.
% This method could be used for align all the virtual scene from multiple headsets and makes them look like they are sharing the same scene in the real world.
% facilitates seamless alignment of virtual and physical elements, 
% This method could also be used for synchronizing a virtual robot with its real-world counterpart in the Kinesthetic Teaching interface and multiple-collector view alignment.
% The implementation of QR tracking varies in different headsets,
% and this function is supported by all the headsets in IRIS.

A spatial anchor serves as a reference point in a 3D environment to accurately position virtual objects within physical space.
It enables virtual objects to maintain their position, orientation, and alignment, even as users move around or leave and return to the environment.
IRIS utilizes either QR codes or motion controllers (currently only implemented for the Meta Quest 3) as spatial anchors.
This approach allows for the alignment of augmented scenes across multiple headsets in the real world, making it appear as though all users are sharing the same scene.
Additionally, this method can synchronize a virtual robot with its real-world counterpart in the {Kinesthetic Teaching} (\ref{sec:kt}) interface and facilitate multiple-collector view alignment.
Fig. \ref{fig:alignment} shows the alignment between real robot and virtual robot.

\input{image/alignment/alignment}

%% file: image/interface/interface.tex
\begin{figure}[h!]
    \centering
    \includegraphics[width=\linewidth]{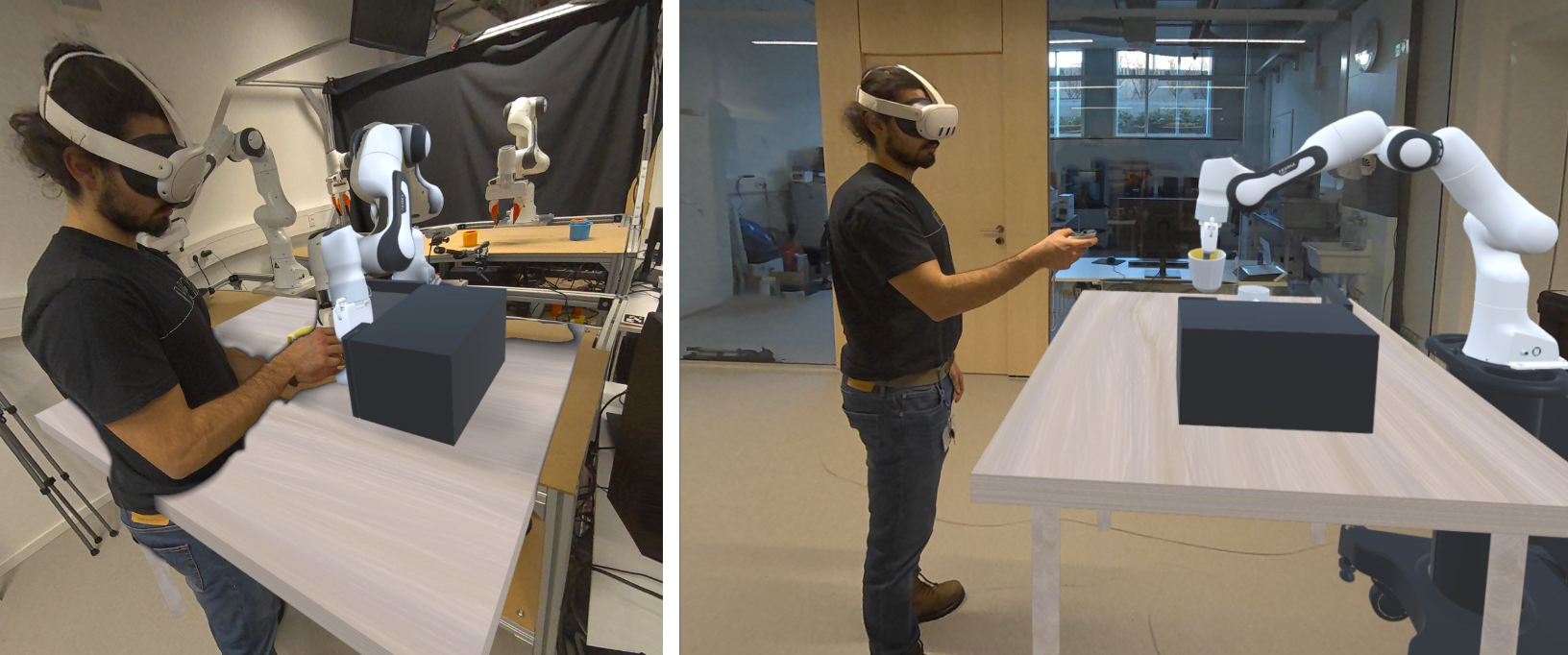}
    \caption{This image illustrates examples of using two interfaces to control robots in simulation: Kinesthetic Teaching (left) and Motion Controller (right), shown from a third-person perspective.}
    \label{fig:interface}
\end{figure}

%% file: image/alignment/alignment.tex
\begin{figure}[h!]
    \centering
    \includegraphics[width=\linewidth]{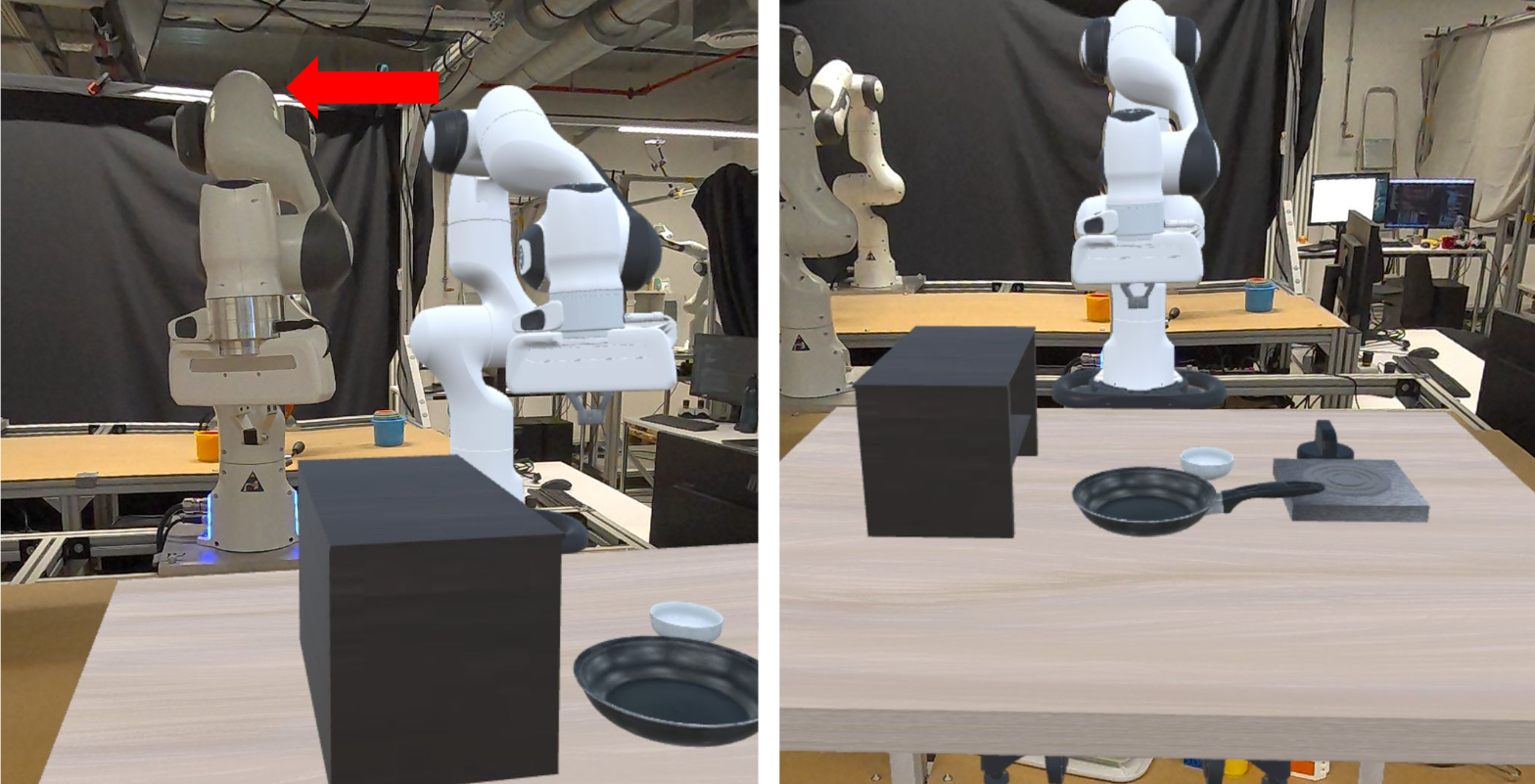}
    \caption{Alignment between real robot and virtual robot with a spatial anchor. It is used for controlling virtual robot by real world kinesthetic teaching or scene sharing among multiple users.}
    \label{fig:alignment}
\end{figure}

%% file: appendix/02_Appendix_Experiment.tex
\section{Experiment Details}
\label{abb:exp}
\subsection{User Study}

\input{image/user_study_task/user_studt_task_fig}

In this section, we provide more details of the user study we conducted, which assesses different data collection interfaces.
As mentioned in the main paper, tasks from the LIBERO benchmark \cite{liu2024libero} were chosen due to their diversity in various movement patterns.
Four representative tasks (see Fig.~\ref{fig:images_grid}) were selected in the dimension of translation, rotation, and compound movement:
\begin{itemize}
\item[-] Task 1: \textit{close the microwave}
\item[-] Task 2: \textit{turn off the stove}
\item[-] Task 3: \textit{pick up the book in the middle and place it on the cabinet shelf}
\item[-] Task 4: \textit{turn on the stove and put the frying pan on it}.
\end{itemize}

To ensure a fair comparison, the tasks and data were taken directly from LIBERO \cite{liu2024libero} without any modification.
The baselines for this study were two standard control interfaces provided by LIBERO: the Keyboard (KB) and the 3D Mouse (3M).
Based on findings from prior research \cite{jiang2024comprehensive},
hand tracking was found to be less stable than motion controllers. Therefore, we selected Kinesthetic Teaching and Motion Controller as the interfaces for the user study.

The study involved eight participants who evaluated the efficiency and intuitiveness of each interface for collecting demonstrations using both objective and subjective metrics.
The objective metrics included the success rate and the average time taken per task.
To ensure successful demonstrations, 
participants executed tasks at a very slow pace, which diluted efficiency measurements (as all interfaces appeared efficient when tasks were performed slowly), which biased participants against later interfaces \cite{jiang2024comprehensive}. 
To mitigate these biases and ensure high-quality data collection, a time limit per task was introduced,
and the time limits for four tasks are 20s, 20s, 30s, and 40s, which are quite enough for finishing the task.
The subjective metrics were assessed through a questionnaire evaluating four dimensions: \textit{Experience}, \textit{Usefulness}, \textit{Intuitiveness}, and \textit{Efficiency}.
Each participant performed each task five times using all interfaces. After finishing using one interface, they provided ratings on the subjective dimensions using a 7-point Likert scale.

For the objective metrics, Table \ref{tab:success_rate} presents the success rates for each interface across the tasks. 
A trial was considered unsuccessful if the participant failed to complete the task or exceeded the time limit. The time limits were set based on task difficulty: 20 seconds for Task 1 and Task 2, 30 seconds for Task 3, and 40 seconds for Task 4.
The data shows a success rate of over $90\%$ across all four tasks when using the KT and MC interfaces from IRIS. In contrast, the Keyboard and 3D Mouse methods from LIBERO often resulted in failures. For example, the 3D Mouse interface achieved only a $37.5\%$ success rate on Task 3.

% Figure \ref{fig:objective_study} shows the average time consumed for each task across four interfaces. The KT and MC interfaces consistently demonstrate lower task completion times, indicating higher efficiency. In contrast, the Keyboard and 3D Mouse interfaces show significantly higher completion times, particularly for Task 3 and Task 4, where the 3D Mouse method approaches or exceeds the task's time limit. These results align with the observed lower success rates for these interfaces, highlighting their inefficiency in time-critical tasks.

% Figure \ref{fig:subjective_study} compares subjective scores for four interfaces based on usefulness, experience, intuitiveness, and efficiency.
% The KT and MC interfaces consistently receive high scores across all criteria, indicating positive user perception and ease of use. 
% In contrast, the Keyboard and 3D Mouse interfaces receive significantly lower ratings, particularly in intuitiveness and efficiency, reflecting the participants' difficulties in using these methods to control the robot.

As is clear in Fig. \ref{fig:user_study_result}, the results of this user study show that IRIS received higher scores than the baseline interfaces across both objective and subjective metrics.
This indicates that the system offers a more intuitive and efficient approach for data collection.

\subsection{Deformable Object Data Collection}
\label{app:deformable}
Demonstrations involving deformable objects and cloth as seen in Fig. \ref{fig:deformable_examples} can be recorded when using IsaacLab for physics simulation. Each point of a simulated deformable object is transmitted to IRIS. This allows for an accurate representation of object deformation during simulation. Transmitting each point individually adds a performance penalty compared to rigid objects. IRIS provides hand-tracking information that is used to control simulated robots and manipulate objects.
The success criteria for the three example tasks as seen in Sec. \ref{sec:deformable_manipulation_experiment} are as follows. \textit{Fold Cloth}: The distance between two of the corner particles is below a threshold. This allows for folding either front-to-back or side-to-side. \textit{Lift Teddy}: The center of the teddy must be lifted above a certain threshold. \textit{Stow Teddy}: The center of the teddy must be inside the box and below a certain threshold. The initial object positions are chosen randomly without rotation. The box position for the \textit{Stow Teddy} task is fixed.
All example policies are using a U-Net diffusion model with reduced down dimension sizes of $[256, 512, 1024]$, and a diffusion step embedding dimension size of $256$. Other parameters are at their default values. The models are trained for 100 to 120 epochs. Observations are robot EEF information, and a mix of depth and image data. The last two observations are stacked and used as input. The models predict 16 actions at a time. Between 60 and 90 demonstrations are used for training.
\input{image/deformable/deformable}

\subsection{High Dynamic Data Collection}
\label{app:table_tennis_evaluation}
Traditional data collection methods struggle to acquire successful demonstrations in dynamic tasks that require high action timeliness and precision. 
To validate the practicality and effectiveness of IRIS, we applied it to a custom-designed table tennis task in MuJoCo simulation environment. IRIS was used to collect high-frequency strike data, and the collected demonstrations were evaluated using state-of-the-art imitation learning algorithms.
In the task, a human demonstrator controls a virtual bat via a handheld controller to return randomly generated incoming balls, closely mimicking real bat manipulation. 
The system captures 73 around 5-second demonstrations at 100 Hz, where each demonstration movement consists of a forward stroke and natural backward release.
The observation includes bat proprioceptive state and dual camera images, and the action is the desired bat position and orientation in task space. 
Fig. \ref{fig:table_tennis_figure} shows the performance of imitation learning models trained on this data, using ball interception rate and successful return rate as evaluation metrics. The trained policy should first reach the incoming ball with natural stroke movement as human players and then hit the ball to opposite side table.

\subsection{Real Robot Experiment}
\label{app:real_robot_experiment}

\input{image/real_world_experiment_appendix}

To evaluate the real-world applicability of IRIS, we conducted a comparative study against a standard teleoperation baseline (Tele-Op) using two real-world manipulation tasks: Cup Inserting and Picking Up Lego (Fig.~\ref{fig:real_world_exp_appendix_combined}). For each task, 30 demonstrations were collected using both IRIS and Tele-Op, under identical experimental conditions. These datasets were then used to train two separate BC-Transformer policies, using consistent hyperparameters. We assessed both the data collection success rate (i.e., the percentage of successful demonstrations during collection) and the policy success rate (i.e., performance of trained policies in executing the tasks). As illustrated in Fig. \ref{fig:real_robot_exp_figure}, IRIS not only achieved a higher success rate during demonstration collection, but also resulted in better-performing policies compared to those trained on Tele-Op data. These results indicate that IRIS facilitates higher-quality real-world data collection, which in turn leads to more effective imitation learning outcomes.

\section{System Performance Analysis}

The performance analysis focuses on two key aspects: network latency and headset FPS (frames per second).
Since IRIS uses asynchronous bidirectional data transfer, network latency is low, averaging around 20-30 ms. Transmission bandwidth depends on Wi-Fi capacity. Even for large scenes from RoboCasa, which contain over 200 MB of compressed assets, it takes no more than 5 seconds to transfer and generate a full scene with more than 300 objects. For real robot teleoperation, the system handles point cloud data efficiently, achieving a transmission speed of 10,000 points at 60 Hz—exceeding the camera’s frame rate.

The FPS performance depends on the hardware. On the Meta Quest 3, a scene with one robot runs at approximately 70 FPS, while on HoloLens 2, it runs around 40 FPS. As the scene size increases, FPS gradually decreases. With around 200 objects, the Meta Quest 3 headsets struggle to keep up with head movement, causing virtual objects to lag or become stuck. However, performance is additionally influenced by the complexity of the meshes and textures, as these require significant computational resources from the XR headset.

In our experiments by using Meta Quest 3, IRIS successfully handled all benchmark scenarios listed in the paper, except for some scenes from RoboCasa \cite{nasiriany2024robocasa}. These scenes have over 700 MB of assets in one single instance, which is closed to the Meta Quest 3 RAM limit. Nonetheless, IRIS was able to manage most of scenes from RoboCasa without any significant performance issues.
For real robot data collection, the optimal point cloud size is around 10,000 points, which achieves a balance between point cloud quality and FPS, maintaining a frame rate of approximately 40 FPS.

%% file: image/user_study_task/user_studt_task_fig.tex
\begin{figure}[h]
    \centering
    \setlength{\tabcolsep}{2pt} % Reduce space between columns
    \renewcommand{\arraystretch}{.5} % Adjust row spacing
    \begin{tabular}{cccc}
        \includegraphics[width=0.2\textwidth]{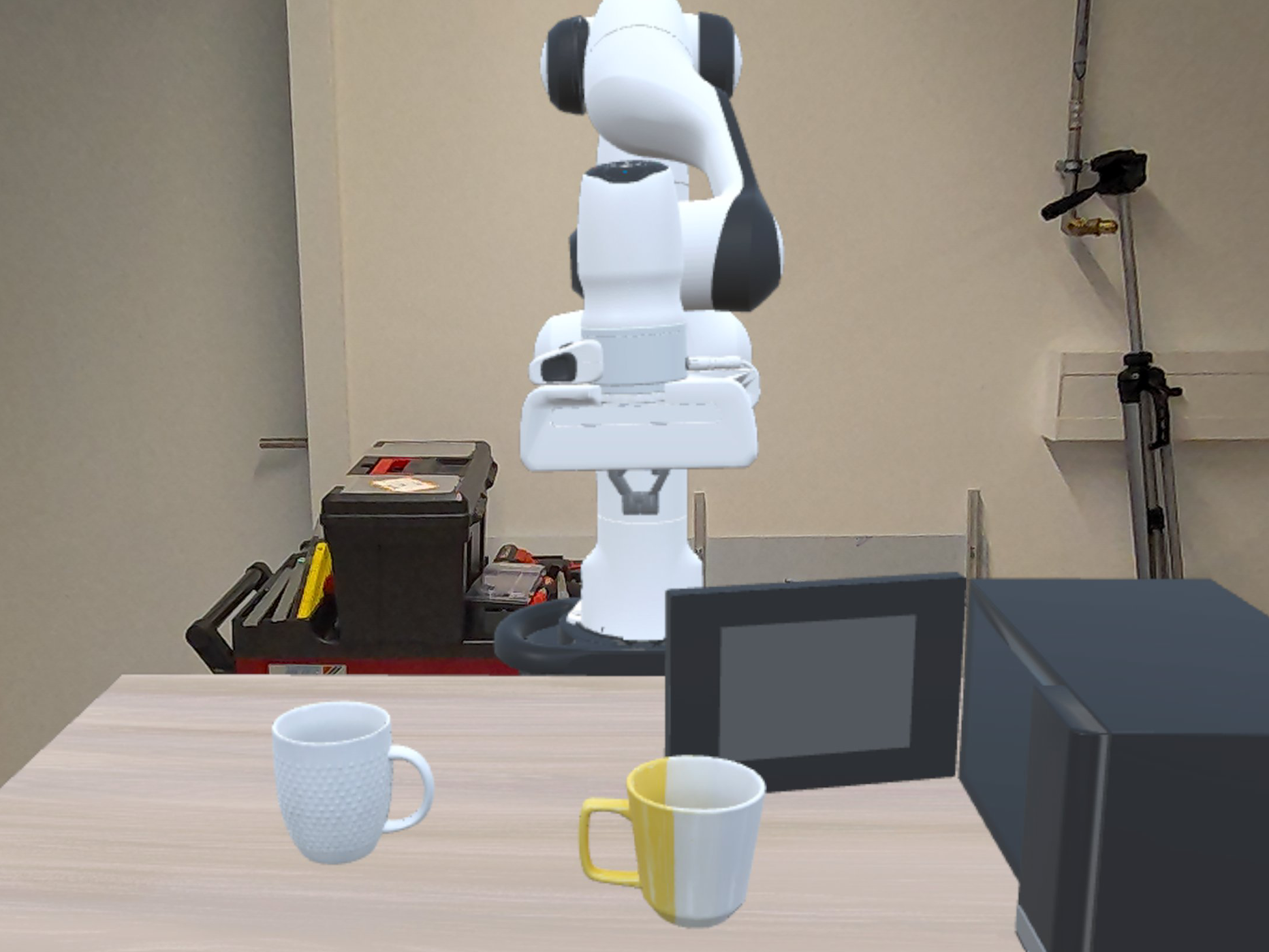} & 
        \includegraphics[width=0.2\textwidth]{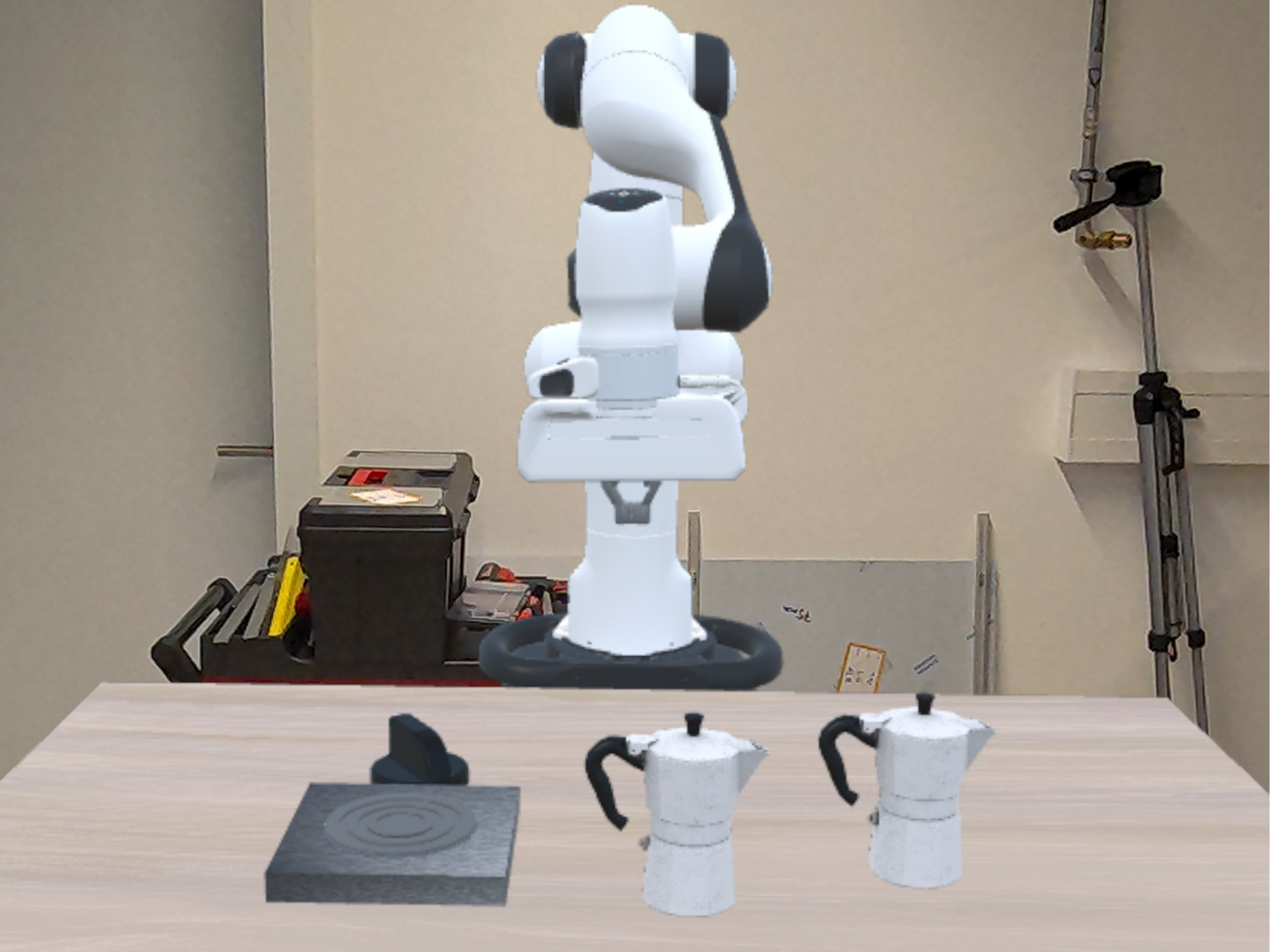} & 
        \includegraphics[width=0.2\textwidth]{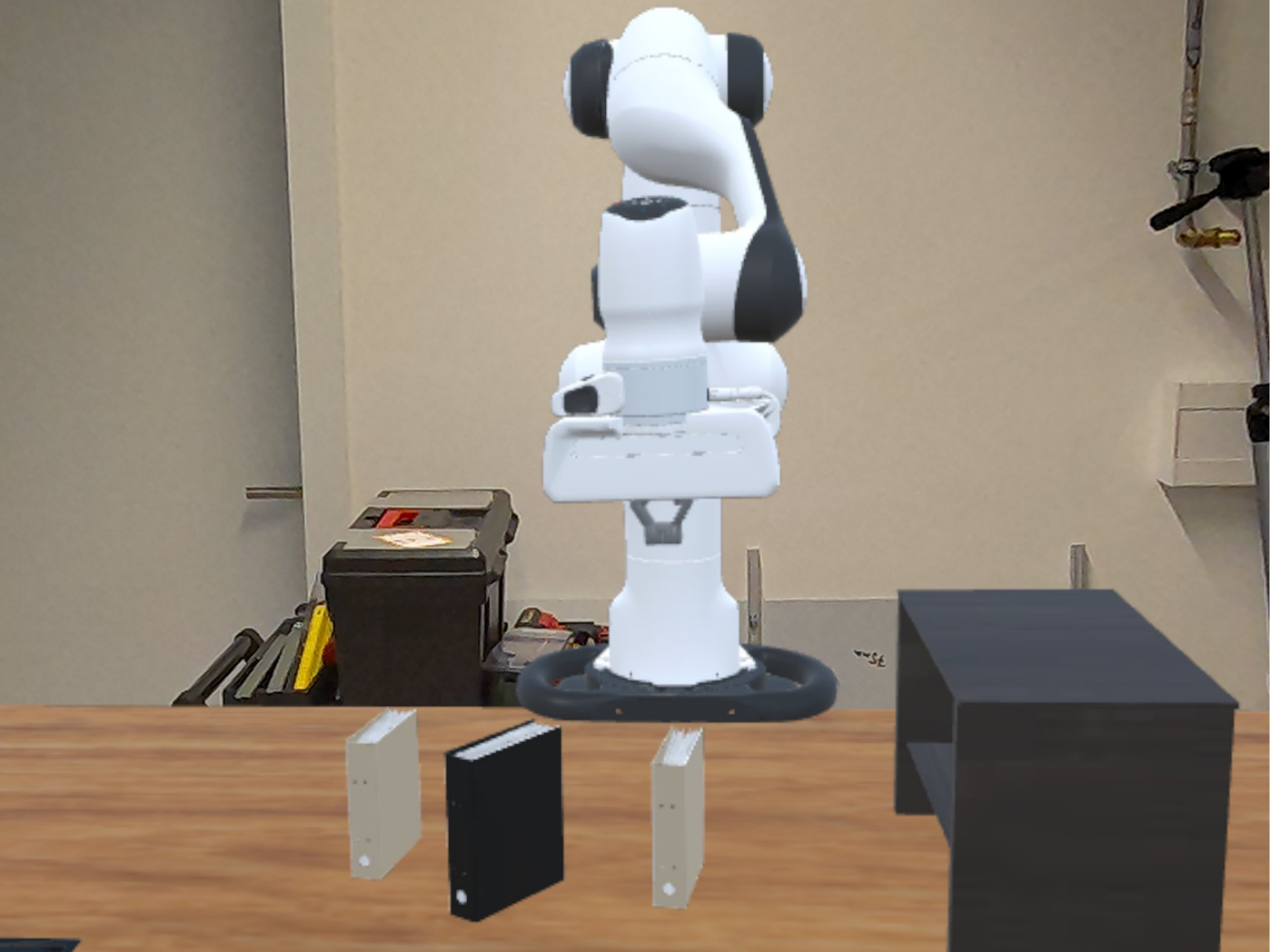} & 
        \includegraphics[width=0.2\textwidth]{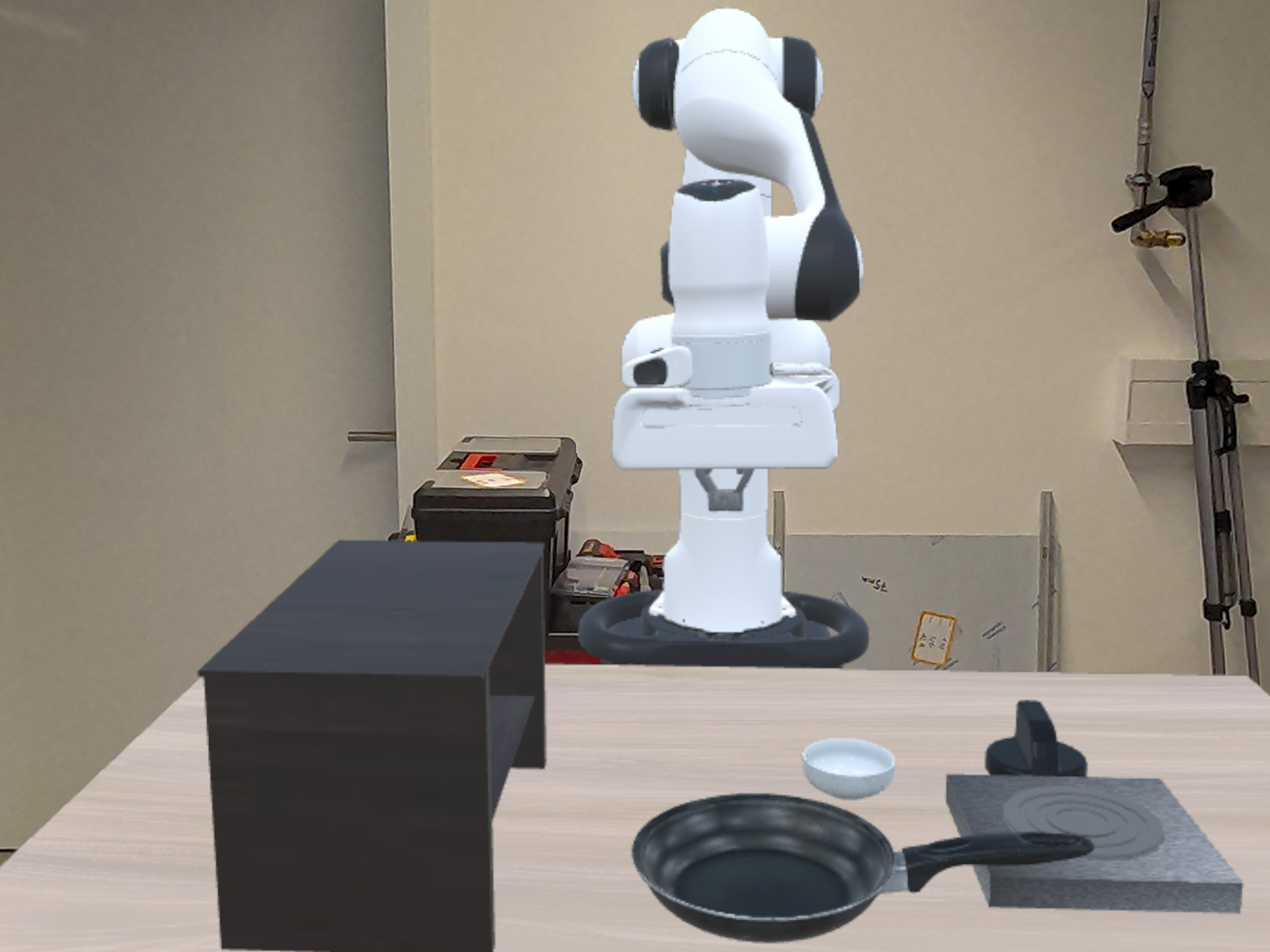} \\
        \scriptsize (a1) & 
        \scriptsize (b1) & 
        \scriptsize (c1) & 
        \scriptsize (d1) \\
        \includegraphics[width=0.2\textwidth]{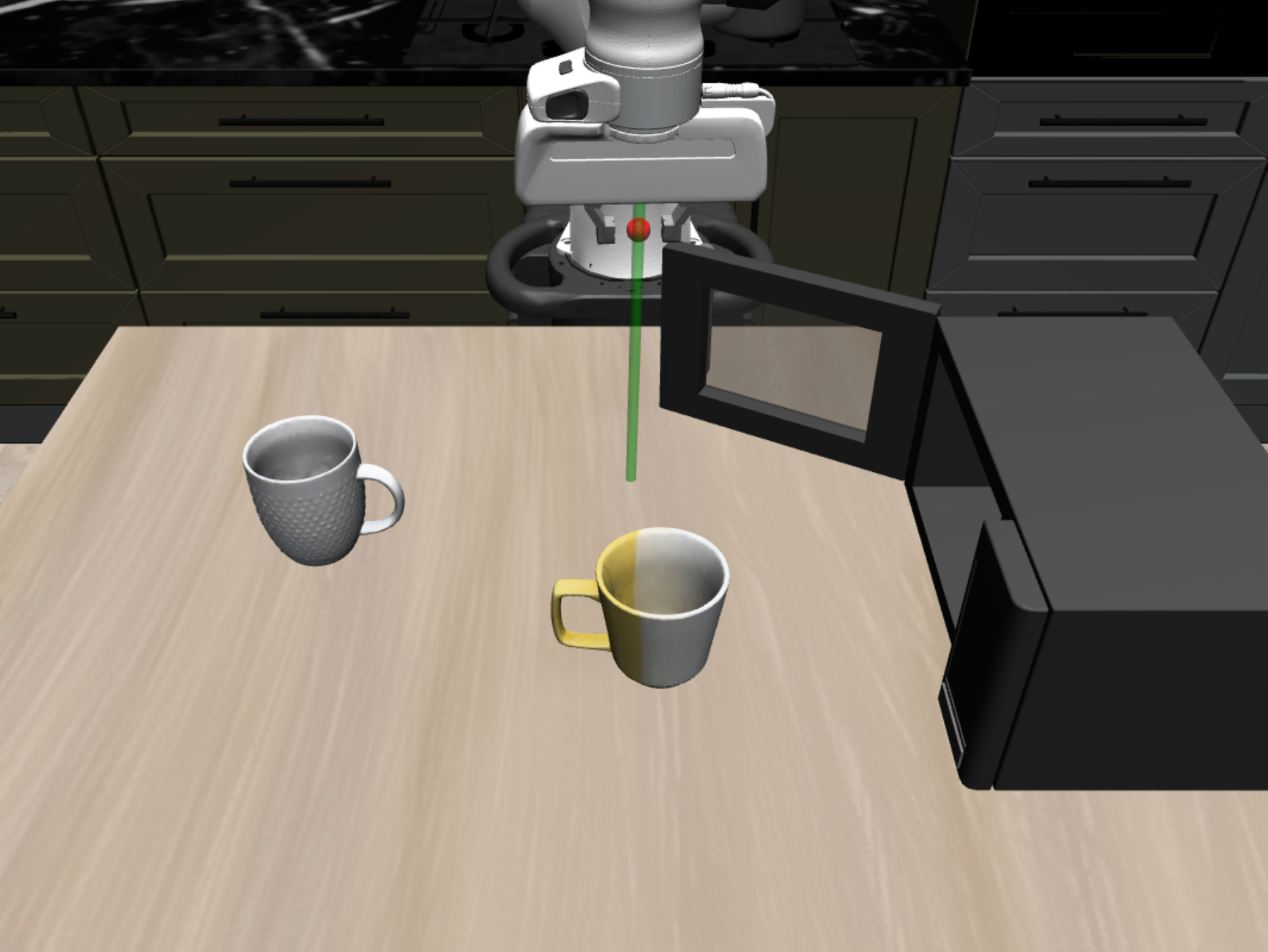} & 
        \includegraphics[width=0.2\textwidth]{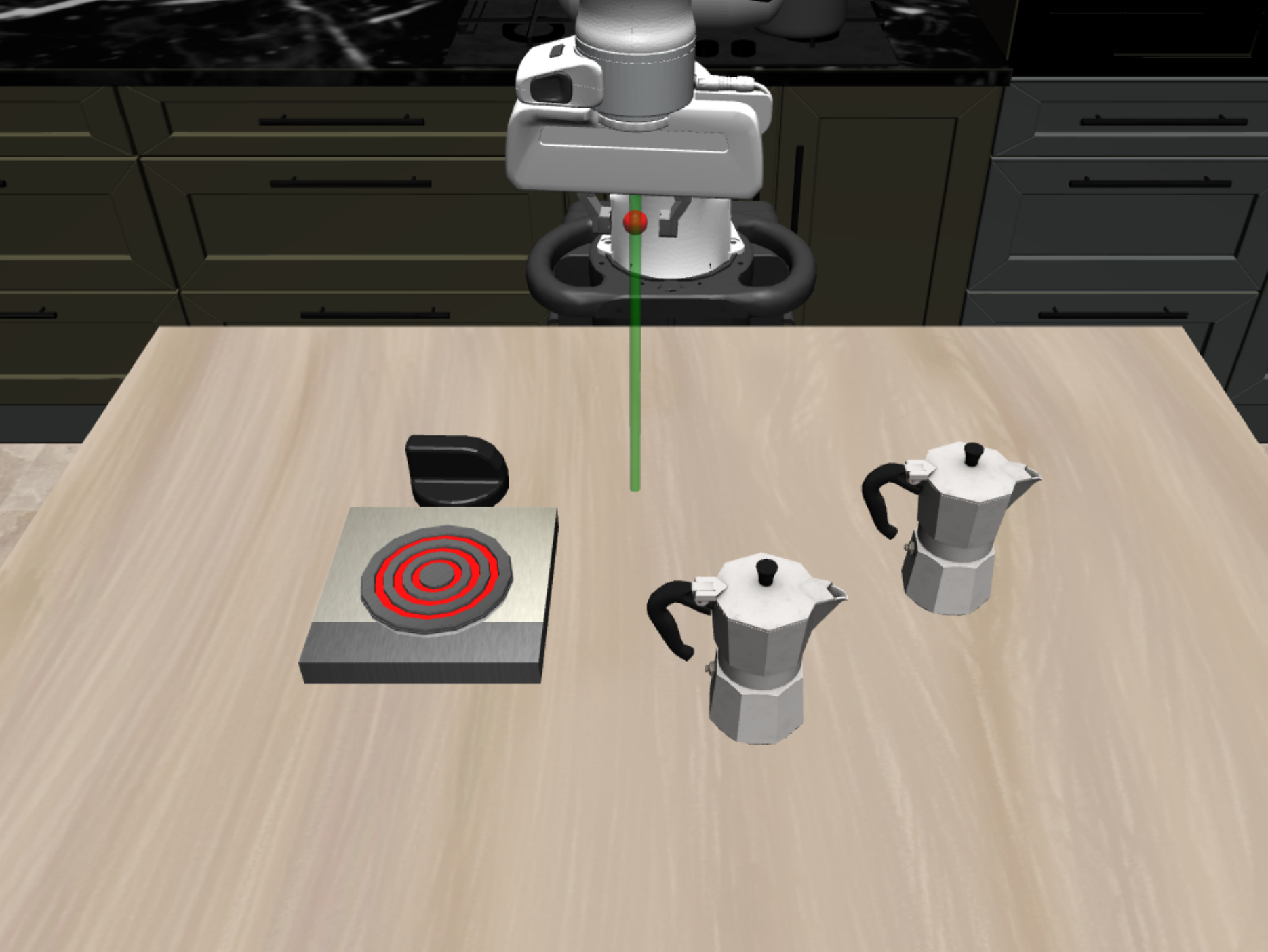} & 
        \includegraphics[width=0.2\textwidth]{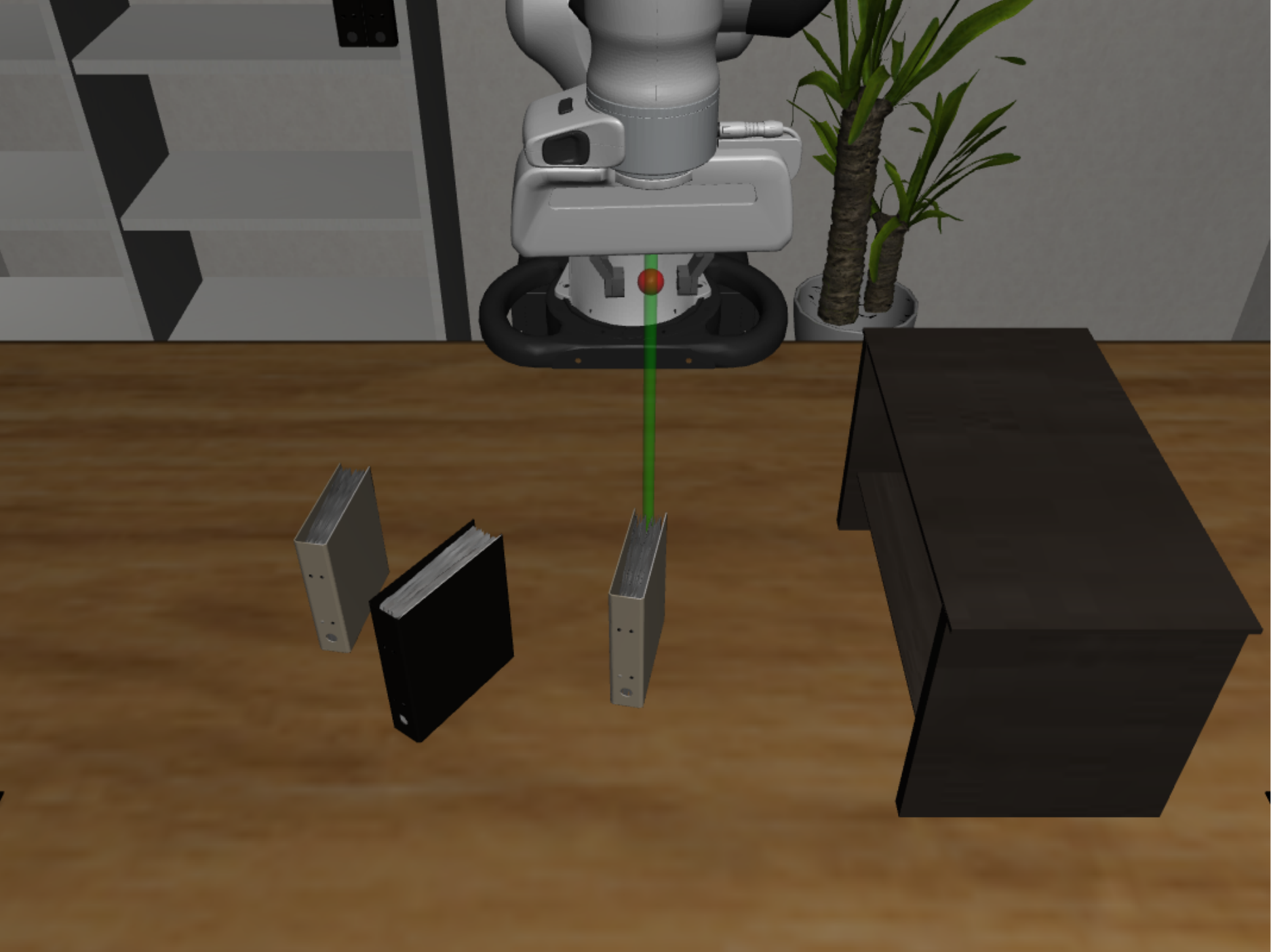} & 
        \includegraphics[width=0.2\textwidth]{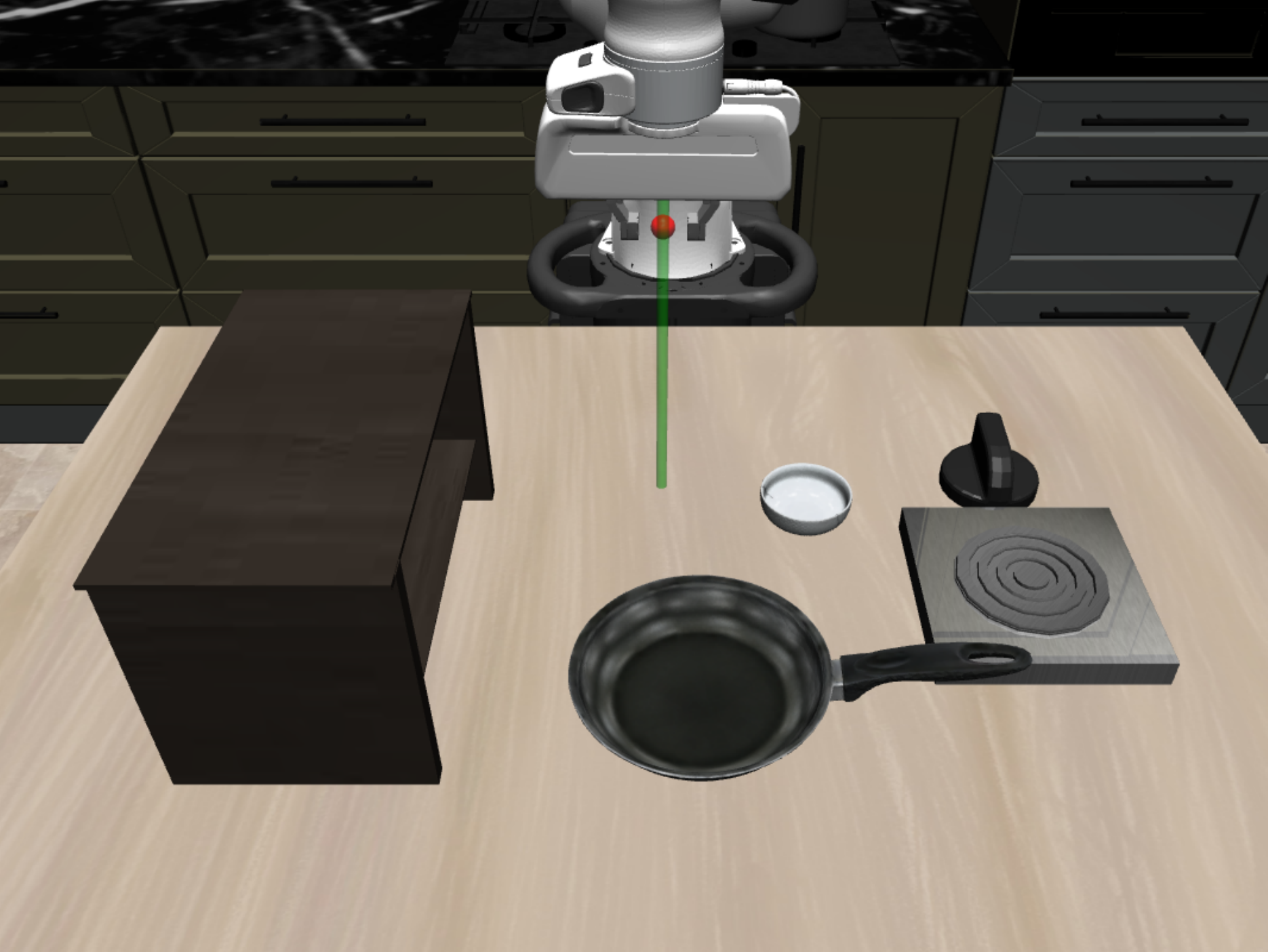} \\
        \scriptsize (a2) & 
        \scriptsize (b2) & 
        \scriptsize (c2) & 
        \scriptsize (d2) \\
    \end{tabular}
    \caption{
    Four tasks from LIBERO in simulation (top row: a1, b1, c1, d1) and corresponding view from Meta Quest 3 (bottom row: a2, b2, c2, d2): (a) Close the microwave, (b) Turn off the stove, (c) Pick up the book and place it on the shelf, (d) Turn on the stove and place the frying pan on it.}
    \label{fig:images_grid}
\end{figure}

%% file: image/deformable/deformable.tex
\begin{figure}[h]
\includegraphics[width=0.24\linewidth]{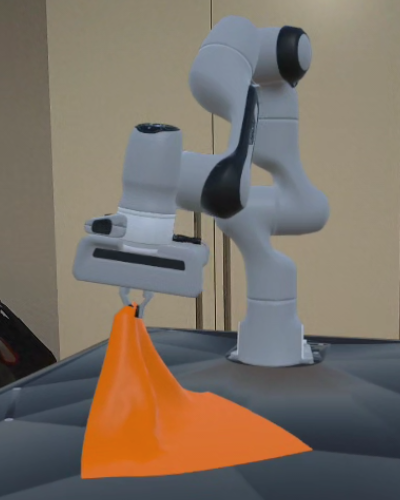}
\includegraphics[width=0.24\linewidth]{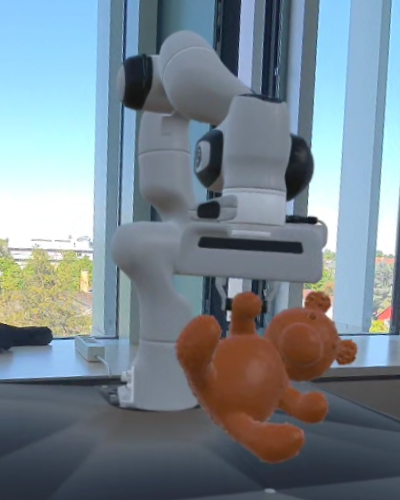}
\includegraphics[width=0.48\linewidth]{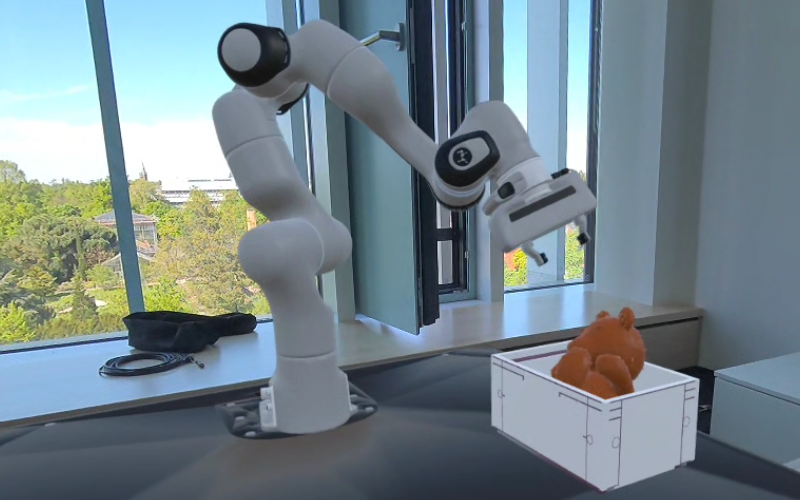}
\caption{Views wearing a VR headset and collecting task demonstrations using IRIS with IsaacLab. Tasks from left to right are: folding a cloth in half, lifting a deformable teddy, and stowing a deformable teddy in a slightly undersized box.}
\label{fig:deformable_examples}
\end{figure}

%% file: image/real_world_experiment_appendix.tex
\begin{figure}[t]
\centering
\begin{subfigure}[b]{0.45\textwidth}
\includegraphics[width=\textwidth]{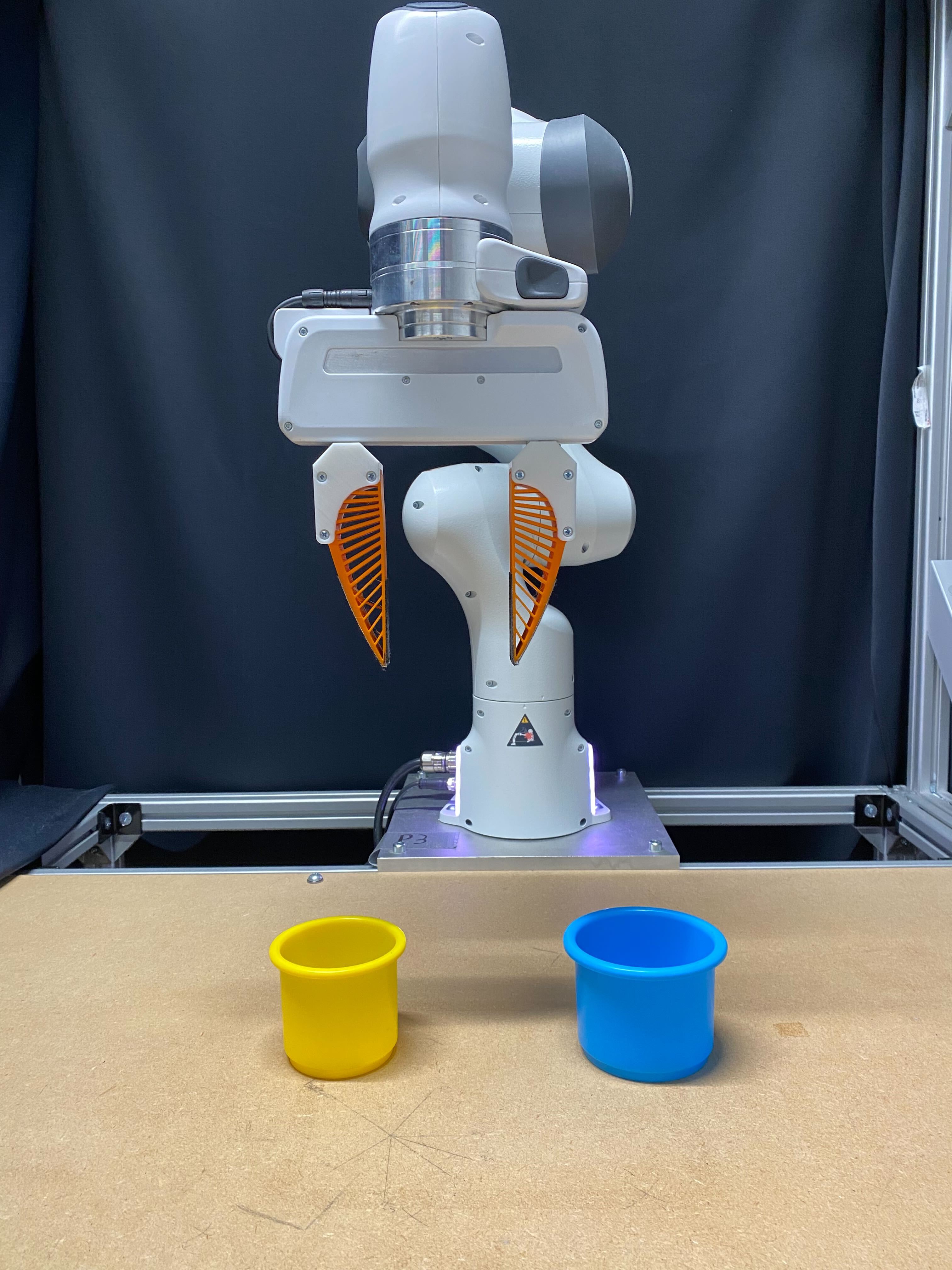}
% \caption{Policy Performance of Table Tennis Task}
% \label{fig:real_world_exp_appendix_1}
\end{subfigure}
% \hfill
\hspace{0.2cm}
\begin{subfigure}[b]{0.45\textwidth}
\includegraphics[width=\textwidth]{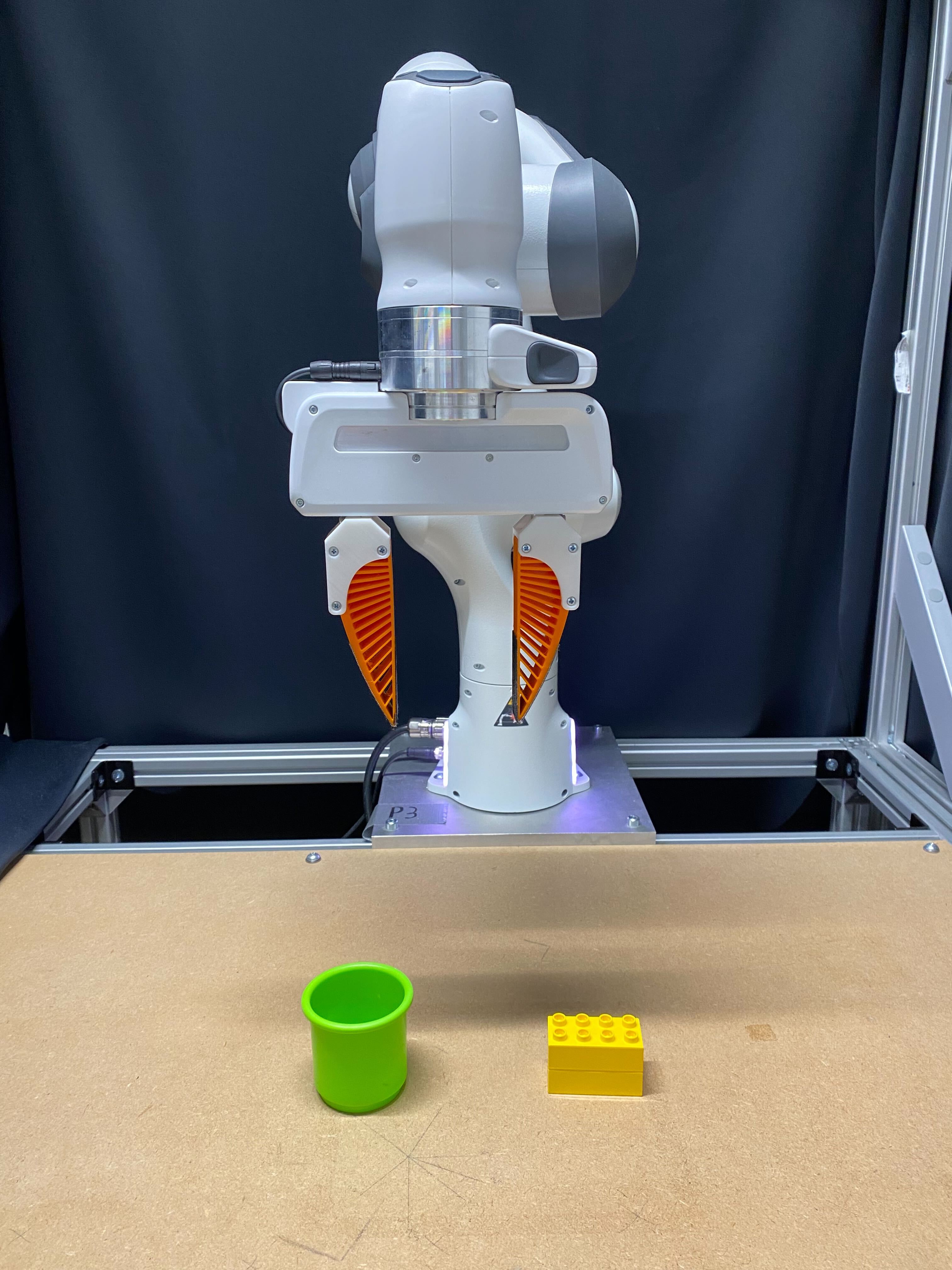}
% \caption{Policy Performance of Real World Experiment}
% \label{fig:real_world_exp_appendix_2}
\end{subfigure}
\caption{Two real-world manipulation tasks: Cup Inserting and Picking Up Lego.}
\vspace{-0.3cm}
\label{fig:real_world_exp_appendix_combined}
\end{figure}

%% file: example.bbl
\begin{thebibliography}{76}
\providecommand{\natexlab}[1]{#1}
\providecommand{\url}[1]{\texttt{#1}}
\expandafter\ifx\csname urlstyle\endcsname\relax
  \providecommand{\doi}[1]{doi: #1}\else
  \providecommand{\doi}{doi: \begingroup \urlstyle{rm}\Url}\fi

\bibitem[Aldaco et~al.(2024)Aldaco, Armstrong, Baruch, Bingham, Chan, Draper, Dwibedi, Finn, Florence, Goodrich, et~al.]{aldaco2024aloha}
J.~Aldaco, T.~Armstrong, R.~Baruch, J.~Bingham, S.~Chan, K.~Draper, D.~Dwibedi, C.~Finn, P.~Florence, S.~Goodrich, et~al.
\newblock Aloha 2: An enhanced low-cost hardware for bimanual teleoperation.
\newblock \emph{arXiv preprint arXiv:2405.02292}, 2024.

\bibitem[Wang et~al.(2024)Wang, Shi, Wang, Zhang, Fei-Fei, and Liu]{wang2024dexcap}
C.~Wang, H.~Shi, W.~Wang, R.~Zhang, L.~Fei-Fei, and C.~K. Liu.
\newblock Dexcap: Scalable and portable mocap data collection system for dexterous manipulation.
\newblock \emph{arXiv preprint arXiv:2403.07788}, 2024.

\bibitem[Mann et~al.(2020)Mann, Ryder, Subbiah, Kaplan, Dhariwal, Neelakantan, Shyam, Sastry, Askell, Agarwal, et~al.]{mann2020language}
B.~Mann, N.~Ryder, M.~Subbiah, J.~Kaplan, P.~Dhariwal, A.~Neelakantan, P.~Shyam, G.~Sastry, A.~Askell, S.~Agarwal, et~al.
\newblock Language models are few-shot learners.
\newblock \emph{arXiv preprint arXiv:2005.14165}, 1, 2020.

\bibitem[Radford et~al.(2021)Radford, Kim, Hallacy, Ramesh, Goh, Agarwal, Sastry, Askell, Mishkin, Clark, et~al.]{radford2021learning}
A.~Radford, J.~W. Kim, C.~Hallacy, A.~Ramesh, G.~Goh, S.~Agarwal, G.~Sastry, A.~Askell, P.~Mishkin, J.~Clark, et~al.
\newblock Learning transferable visual models from natural language supervision.
\newblock In \emph{International conference on machine learning}, pages 8748--8763. PMLR, 2021.

\bibitem[Gao et~al.(2024)Gao, Xie, Xiao, Finn, and Sadigh]{gao2024efficient}
J.~Gao, A.~Xie, T.~Xiao, C.~Finn, and D.~Sadigh.
\newblock Efficient data collection for robotic manipulation via compositional generalization.
\newblock \emph{arXiv preprint arXiv:2403.05110}, 2024.

\bibitem[wik()]{wikipediaExtendedReality}
{E}xtended reality - {W}ikipedia --- en.wikipedia.org.
\newblock \url{https://en.wikipedia.org/wiki/Extended_reality}.
\newblock [Accessed 11-01-2025].

\bibitem[Jiang et~al.(2024)Jiang, Mattes, Jia, Schreiber, Neumann, and Lioutikov]{jiang2024comprehensive}
X.~Jiang, P.~Mattes, X.~Jia, N.~Schreiber, G.~Neumann, and R.~Lioutikov.
\newblock A comprehensive user study on augmented reality-based data collection interfaces for robot learning.
\newblock In \emph{Proceedings of the 2024 ACM/IEEE International Conference on Human-Robot Interaction}, pages 333--342, 2024.

\bibitem[Yang et~al.(2024)Yang, Ikeda, Bertasius, and Szafir]{arcade}
Y.~Yang, B.~Ikeda, G.~Bertasius, and D.~Szafir.
\newblock Arcade: Scalable demonstration collection and generation via augmented reality for imitation learning.
\newblock In \emph{2024 IEEE/RSJ International Conference on Intelligent Robots and Systems (IROS)}, pages 2855--2861. IEEE, 2024.

\bibitem[Park et~al.(2024)Park, Bhatia, Ankile, and Agrawal]{dexhub-park}
Y.~Park, J.~S. Bhatia, L.~Ankile, and P.~Agrawal.
\newblock Dexhub and dart: Towards internet scale robot data collection.
\newblock \emph{arXiv preprint arXiv:2411.02214}, 2024.

\bibitem[Iyer et~al.(2024)Iyer, Peng, Dai, Guzey, Haldar, Chintala, and Pinto]{openteach}
A.~Iyer, Z.~Peng, Y.~Dai, I.~Guzey, S.~Haldar, S.~Chintala, and L.~Pinto.
\newblock Open teach: A versatile teleoperation system for robotic manipulation.
\newblock \emph{arXiv preprint arXiv:2403.07870}, 2024.

\bibitem[Cheng et~al.()Cheng, Li, Yang, Yang, and Wang]{opentelevision}
X.~Cheng, J.~Li, S.~Yang, G.~Yang, and X.~Wang.
\newblock Open-television: Teleoperation with immersive active visual feedback.
\newblock In \emph{8th Annual Conference on Robot Learning}.

\bibitem[Luebbers et~al.(2021)Luebbers, Brooks, Mueller, Szafir, and Hayes]{arclfd}
M.~B. Luebbers, C.~Brooks, C.~L. Mueller, D.~Szafir, and B.~Hayes.
\newblock Arc-lfd: Using augmented reality for interactive long-term robot skill maintenance via constrained learning from demonstration.
\newblock In \emph{2021 IEEE International Conference on Robotics and Automation (ICRA)}, pages 3794--3800. IEEE, 2021.

\bibitem[George et~al.(2025)George, Bartsch, and Farimani]{george2025openvr}
A.~George, A.~Bartsch, and A.~B. Farimani.
\newblock Openvr: Teleoperation for manipulation.
\newblock \emph{SoftwareX}, 29:\penalty0 102054, 2025.

\bibitem[Naceri et~al.(2021)Naceri, Mazzanti, Bimbo, Tefera, Prattichizzo, Caldwell, Mattos, and Deshpande]{vicarios}
A.~Naceri, D.~Mazzanti, J.~Bimbo, Y.~T. Tefera, D.~Prattichizzo, D.~G. Caldwell, L.~S. Mattos, and N.~Deshpande.
\newblock The vicarios virtual reality interface for remote robotic teleoperation: Teleporting for intuitive tele-manipulation.
\newblock \emph{Journal of Intelligent \& Robotic Systems}, 101:\penalty0 1--16, 2021.

\bibitem[Mosbach et~al.(2022)Mosbach, Moraw, and Behnke]{mosbach2022accelerating}
M.~Mosbach, K.~Moraw, and S.~Behnke.
\newblock Accelerating interactive human-like manipulation learning with gpu-based simulation and high-quality demonstrations.
\newblock In \emph{2022 IEEE-RAS 21st International Conference on Humanoid Robots (Humanoids)}, pages 435--441. IEEE, 2022.

\bibitem[Lipton et~al.(2017)Lipton, Fay, and Rus]{lipton2017baxter}
J.~I. Lipton, A.~J. Fay, and D.~Rus.
\newblock Baxter's homunculus: Virtual reality spaces for teleoperation in manufacturing.
\newblock \emph{IEEE Robotics and Automation Letters}, 3\penalty0 (1):\penalty0 179--186, 2017.

\bibitem[Nechyporenko et~al.(2024)Nechyporenko, Hoque, Webb, Sivapurapu, and Zhang]{armada}
N.~Nechyporenko, R.~Hoque, C.~Webb, M.~Sivapurapu, and J.~Zhang.
\newblock Armada: Augmented reality for robot manipulation and robot-free data acquisition.
\newblock \emph{arXiv preprint arXiv:2412.10631}, 2024.

\bibitem[Meng et~al.(2023)Meng, Liu, Chai, Wang, and Meng]{meng2023virtual}
L.~Meng, J.~Liu, W.~Chai, J.~Wang, and M.~Q.-H. Meng.
\newblock Virtual reality based robot teleoperation via human-scene interaction.
\newblock \emph{Procedia Computer Science}, 226:\penalty0 141--148, 2023.

\bibitem[Todorov et~al.(2012)Todorov, Erez, and Tassa]{todorov2012mujoco}
E.~Todorov, T.~Erez, and Y.~Tassa.
\newblock Mujoco: A physics engine for model-based control.
\newblock In \emph{2012 IEEE/RSJ international conference on intelligent robots and systems}, pages 5026--5033. IEEE, 2012.

\bibitem[Mittal et~al.(2023)Mittal, Yu, Yu, Liu, Rudin, Hoeller, Yuan, Singh, Guo, Mazhar, Mandlekar, Babich, State, Hutter, and Garg]{mittal2023orbit}
M.~Mittal, C.~Yu, Q.~Yu, J.~Liu, N.~Rudin, D.~Hoeller, J.~L. Yuan, R.~Singh, Y.~Guo, H.~Mazhar, A.~Mandlekar, B.~Babich, G.~State, M.~Hutter, and A.~Garg.
\newblock Orbit: A unified simulation framework for interactive robot learning environments.
\newblock \emph{IEEE Robotics and Automation Letters}, 8\penalty0 (6):\penalty0 3740--3747, 2023.
\newblock \doi{10.1109/LRA.2023.3270034}.

\bibitem[Rohmer et~al.(2013)Rohmer, Singh, and Freese]{coppeliaSim}
E.~Rohmer, S.~P.~N. Singh, and M.~Freese.
\newblock Coppeliasim (formerly v-rep): a versatile and scalable robot simulation framework.
\newblock In \emph{Proc. of The International Conference on Intelligent Robots and Systems (IROS)}, 2013.
\newblock www.coppeliarobotics.com.

\bibitem[Authors(2024)]{Genesis}
G.~Authors.
\newblock Genesis: A universal and generative physics engine for robotics and beyond, December 2024.
\newblock URL \url{https://github.com/Genesis-Embodied-AI/Genesis}.

\bibitem[Technologies()]{unity3dUnityManual}
U.~Technologies.
\newblock {U}nity - {M}anual: {U}nity 6 {U}ser {M}anual --- docs.unity3d.com.
\newblock \url{https://docs.unity3d.com/6000.0/Documentation/Manual/UnityManual.html}.
\newblock [Accessed 31-01-2025].

\bibitem[Fan et~al.(2023)Fan, Guo, Feng, Lin, Wang, Liang, Garrad, Rossiter, Zhang, Lepora, et~al.]{digitaltwinmr}
W.~Fan, X.~Guo, E.~Feng, J.~Lin, Y.~Wang, J.~Liang, M.~Garrad, J.~Rossiter, Z.~Zhang, N.~Lepora, et~al.
\newblock Digital twin-driven mixed reality framework for immersive teleoperation with haptic rendering.
\newblock \emph{IEEE Robotics and Automation Letters}, 2023.

\bibitem[Zhu et~al.(2023)Zhu, Jiang, Chen, Aoyama, and Hasegawa]{sharedctlframework}
Y.~Zhu, B.~Jiang, Q.~Chen, T.~Aoyama, and Y.~Hasegawa.
\newblock A shared control framework for enhanced grasping performance in teleoperation.
\newblock \emph{IEEE Access}, 2023.

\bibitem[Arunachalam et~al.(2023)Arunachalam, G{\"u}zey, Chintala, and Pinto]{holodex}
S.~P. Arunachalam, I.~G{\"u}zey, S.~Chintala, and L.~Pinto.
\newblock Holo-dex: Teaching dexterity with immersive mixed reality.
\newblock In \emph{2023 IEEE International Conference on Robotics and Automation (ICRA)}, pages 5962--5969. IEEE, 2023.

\bibitem[Ding et~al.(2024)Ding, Qin, Zhu, Jia, Yang, Yang, Qi, and Wang]{bunnyvisionpro}
R.~Ding, Y.~Qin, J.~Zhu, C.~Jia, S.~Yang, R.~Yang, X.~Qi, and X.~Wang.
\newblock Bunny-visionpro: Real-time bimanual dexterous teleoperation for imitation learning.
\newblock \emph{arXiv preprint arXiv:2407.03162}, 2024.

\bibitem[Audonnet et~al.(2024)Audonnet, Ramirez-Alpizar, and Aragon-Camarasa]{immertwin}
F.~P. Audonnet, I.~G. Ramirez-Alpizar, and G.~Aragon-Camarasa.
\newblock Immertwin: A mixed reality framework for enhanced robotic arm teleoperation.
\newblock \emph{arXiv preprint arXiv:2409.08964}, 2024.

\bibitem[Szczurek et~al.(2023)Szczurek, Prades, Matheson, Rodriguez-Nogueira, and Di~Castro]{szczurek2023multimodal}
K.~A. Szczurek, R.~M. Prades, E.~Matheson, J.~Rodriguez-Nogueira, and M.~Di~Castro.
\newblock Multimodal multi-user mixed reality human--robot interface for remote operations in hazardous environments.
\newblock \emph{IEEE Access}, 11:\penalty0 17305--17333, 2023.

\bibitem[Zhao et~al.(2023)Zhao, Kumar, Levine, and Finn]{zhao2023learning}
T.~Z. Zhao, V.~Kumar, S.~Levine, and C.~Finn.
\newblock Learning fine-grained bimanual manipulation with low-cost hardware.
\newblock \emph{arXiv preprint arXiv:2304.13705}, 2023.

\bibitem[Fu et~al.(2024)Fu, Zhao, and Finn]{fu2024mobile}
Z.~Fu, T.~Z. Zhao, and C.~Finn.
\newblock Mobile aloha: Learning bimanual mobile manipulation with low-cost whole-body teleoperation.
\newblock In \emph{{Conference on Robot Learning (CoRL)}}, 2024.

\bibitem[Wu et~al.(2023)Wu, Shentu, Yi, Lin, and Abbeel]{wu2023gello}
P.~Wu, Y.~Shentu, Z.~Yi, X.~Lin, and P.~Abbeel.
\newblock Gello: A general, low-cost, and intuitive teleoperation framework for robot manipulators.
\newblock \emph{arXiv preprint arXiv:2309.13037}, 2023.

\bibitem[Qin et~al.(2023)Qin, Yang, Huang, Wyk, Su, Wang, Chao, and Fox]{Qin2023AnyTeleopAG}
Y.~Qin, W.~Yang, B.~Huang, K.~V. Wyk, H.~Su, X.~Wang, Y.-W. Chao, and D.~Fox.
\newblock Anyteleop: A general vision-based dexterous robot arm-hand teleoperation system.
\newblock \emph{ArXiv}, abs/2307.04577, 2023.
\newblock URL \url{https://api.semanticscholar.org/CorpusID:259367735}.

\bibitem[Duan et~al.(2023)Duan, Wang, Shridhar, Fox, and Krishna]{ar2-d2-pmlr-v229-duan23a}
J.~Duan, Y.~R. Wang, M.~Shridhar, D.~Fox, and R.~Krishna.
\newblock Ar2-d2: Training a robot without a robot.
\newblock In J.~Tan, M.~Toussaint, and K.~Darvish, editors, \emph{Proceedings of The 7th Conference on Robot Learning}, volume 229 of \emph{Proceedings of Machine Learning Research}, pages 2838--2848. PMLR, 06--09 Nov 2023.
\newblock URL \url{https://proceedings.mlr.press/v229/duan23a.html}.

\bibitem[Wang et~al.(2024)Wang, Chang, Duan, Fox, and Krishna]{eve}
J.~Wang, C.-C. Chang, J.~Duan, D.~Fox, and R.~Krishna.
\newblock Eve: Enabling anyone to train robots using augmented reality.
\newblock In \emph{Proceedings of the 37th Annual ACM Symposium on User Interface Software and Technology}, UIST '24, New York, NY, USA, 2024. Association for Computing Machinery.
\newblock ISBN 9798400706288.
\newblock \doi{10.1145/3654777.3676413}.
\newblock URL \url{https://doi.org/10.1145/3654777.3676413}.

\bibitem[Arevalo~Arboleda et~al.(2021)Arevalo~Arboleda, R{\"u}cker, Dierks, and Gerken]{augmentedvisualcues}
S.~Arevalo~Arboleda, F.~R{\"u}cker, T.~Dierks, and J.~Gerken.
\newblock Assisting manipulation and grasping in robot teleoperation with augmented reality visual cues.
\newblock In \emph{Proceedings of the 2021 CHI conference on human factors in computing systems}, pages 1--14, 2021.

\bibitem[Wang et~al.(2024)Wang, Guo, Xu, Zhang, Sun, and Xu]{wang2024robotic}
X.~Wang, S.~Guo, Z.~Xu, Z.~Zhang, Z.~Sun, and Y.~Xu.
\newblock A robotic teleoperation system enhanced by augmented reality for natural human--robot interaction.
\newblock \emph{Cyborg and Bionic Systems}, 5:\penalty0 0098, 2024.

\bibitem[Arunachalam et~al.(2023)Arunachalam, G{\"u}zey, Chintala, and Pinto]{arunachalam2023holo}
S.~P. Arunachalam, I.~G{\"u}zey, S.~Chintala, and L.~Pinto.
\newblock Holo-dex: Teaching dexterity with immersive mixed reality.
\newblock In \emph{2023 IEEE International Conference on Robotics and Automation (ICRA)}, pages 5962--5969. IEEE, 2023.

\bibitem[Chen et~al.(2024)Chen, Wang, Nguyen, Fei-Fei, and Liu]{chen2024arcap}
S.~Chen, C.~Wang, K.~Nguyen, L.~Fei-Fei, and C.~K. Liu.
\newblock Arcap: Collecting high-quality human demonstrations for robot learning with augmented reality feedback.
\newblock \emph{arXiv preprint arXiv:2410.08464}, 2024.

\bibitem[Quigley et~al.(2009)Quigley, Conley, Gerkey, Faust, Foote, Leibs, Wheeler, Ng, et~al.]{quigley2009ros}
M.~Quigley, K.~Conley, B.~Gerkey, J.~Faust, T.~Foote, J.~Leibs, R.~Wheeler, A.~Y. Ng, et~al.
\newblock Ros: an open-source robot operating system.
\newblock In \emph{ICRA workshop on open source software}, volume~3, page~5. Kobe, Japan, 2009.

\bibitem[zer()]{zeromqZeroMQ}
{Z}ero{M}{Q} --- zeromq.org.
\newblock \url{https://zeromq.org/}.
\newblock [Accessed 01-05-2025].

\bibitem[lolambean(2023)]{hololens2_lolambean_2023}
lolambean.
\newblock Hololens 2 hardware, Mar. 2023.
\newblock URL \url{https://learn.microsoft.com/en-us/hololens/hololens2-hardware}.

\bibitem[wik()]{wikipediaMetaQuest}
{M}eta {Q}uest 3 - {W}ikipedia --- en.wikipedia.org.
\newblock \url{https://en.wikipedia.org/wiki/Meta_Quest_3}.
\newblock [Accessed 01-05-2025].

\bibitem[Yu et~al.(2020)Yu, Quillen, He, Julian, Hausman, Finn, and Levine]{yu2020meta}
T.~Yu, D.~Quillen, Z.~He, R.~Julian, K.~Hausman, C.~Finn, and S.~Levine.
\newblock Meta-world: A benchmark and evaluation for multi-task and meta reinforcement learning.
\newblock In \emph{Conference on robot learning}, pages 1094--1100. PMLR, 2020.

\bibitem[Liu et~al.(2024)Liu, Zhu, Gao, Feng, Liu, Zhu, and Stone]{liu2024libero}
B.~Liu, Y.~Zhu, C.~Gao, Y.~Feng, Q.~Liu, Y.~Zhu, and P.~Stone.
\newblock Libero: Benchmarking knowledge transfer for lifelong robot learning.
\newblock \emph{Advances in Neural Information Processing Systems}, 36, 2024.

\bibitem[Nasiriany et~al.(2024)Nasiriany, Maddukuri, Zhang, Parikh, Lo, Joshi, Mandlekar, and Zhu]{nasiriany2024robocasa}
S.~Nasiriany, A.~Maddukuri, L.~Zhang, A.~Parikh, A.~Lo, A.~Joshi, A.~Mandlekar, and Y.~Zhu.
\newblock Robocasa: Large-scale simulation of everyday tasks for generalist robots.
\newblock \emph{arXiv preprint arXiv:2406.02523}, 2024.

\bibitem[Zhu et~al.(2020)Zhu, Wong, Mandlekar, Mart{\'\i}n-Mart{\'\i}n, Joshi, Nasiriany, and Zhu]{zhu2020robosuite}
Y.~Zhu, J.~Wong, A.~Mandlekar, R.~Mart{\'\i}n-Mart{\'\i}n, A.~Joshi, S.~Nasiriany, and Y.~Zhu.
\newblock robosuite: A modular simulation framework and benchmark for robot learning.
\newblock \emph{arXiv preprint arXiv:2009.12293}, 2020.

\bibitem[Otto et~al.()Otto, Celik, Roth, and Zhou]{fancy_gym}
F.~Otto, O.~Celik, D.~Roth, and H.~Zhou.
\newblock Fancy gym.
\newblock URL \url{https://github.com/ALRhub/fancy_gym}.

\bibitem[James et~al.(2019)James, Freese, and Davison]{james2019pyrep}
S.~James, M.~Freese, and A.~J. Davison.
\newblock Pyrep: Bringing v-rep to deep robot learning.
\newblock \emph{arXiv preprint arXiv:1906.11176}, 2019.

\bibitem[Pumacay et~al.(2024)Pumacay, Singh, Duan, Krishna, Thomason, and Fox]{pumacay2024colosseum}
W.~Pumacay, I.~Singh, J.~Duan, R.~Krishna, J.~Thomason, and D.~Fox.
\newblock The colosseum: A benchmark for evaluating generalization for robotic manipulation.
\newblock \emph{arXiv preprint arXiv:2402.08191}, 2024.

\bibitem[Tung et~al.(2021)Tung, Wong, Mandlekar, Mart{\'\i}n-Mart{\'\i}n, Zhu, Fei-Fei, and Savarese]{tung2021learning}
A.~Tung, J.~Wong, A.~Mandlekar, R.~Mart{\'\i}n-Mart{\'\i}n, Y.~Zhu, L.~Fei-Fei, and S.~Savarese.
\newblock Learning multi-arm manipulation through collaborative teleoperation.
\newblock In \emph{2021 IEEE International Conference on Robotics and Automation (ICRA)}, pages 9212--9219. IEEE, 2021.

\bibitem[Otto et~al.(2023{\natexlab{a}})Otto, Celik, Zhou, Ziesche, Ngo, and Neumann]{otto2023deep}
F.~Otto, O.~Celik, H.~Zhou, H.~Ziesche, V.~A. Ngo, and G.~Neumann.
\newblock Deep black-box reinforcement learning with movement primitives.
\newblock In \emph{Conference on Robot Learning}, pages 1244--1265. PMLR, 2023{\natexlab{a}}.

\bibitem[Otto et~al.(2023{\natexlab{b}})Otto, Zhou, Celik, Li, Lioutikov, and Neumann]{otto2023mp3}
F.~Otto, H.~Zhou, O.~Celik, G.~Li, R.~Lioutikov, and G.~Neumann.
\newblock Mp3: Movement primitive-based (re-) planning policy.
\newblock \emph{arXiv preprint arXiv:2306.12729}, 2023{\natexlab{b}}.

\bibitem[Mandlekar et~al.(2018)Mandlekar, Zhu, Garg, Booher, Spero, Tung, Gao, Emmons, Gupta, Orbay, et~al.]{mandlekar2018roboturk}
A.~Mandlekar, Y.~Zhu, A.~Garg, J.~Booher, M.~Spero, A.~Tung, J.~Gao, J.~Emmons, A.~Gupta, E.~Orbay, et~al.
\newblock Roboturk: A crowdsourcing platform for robotic skill learning through imitation.
\newblock In \emph{Conference on Robot Learning}, pages 879--893. PMLR, 2018.

\bibitem[Luo et~al.(2024)Luo, Xu, Wu, and Levine]{luo2024precise}
J.~Luo, C.~Xu, J.~Wu, and S.~Levine.
\newblock Precise and dexterous robotic manipulation via human-in-the-loop reinforcement learning.
\newblock \emph{arXiv preprint arXiv:2410.21845}, 2024.

\bibitem[Mandlekar et~al.(2023)Mandlekar, Garrett, Xu, and Fox]{mandlekar2023human}
A.~Mandlekar, C.~R. Garrett, D.~Xu, and D.~Fox.
\newblock Human-in-the-loop task and motion planning for imitation learning.
\newblock In \emph{Conference on Robot Learning}, pages 3030--3060. PMLR, 2023.

\bibitem[Finstad(2010)]{finstad2010usability}
K.~Finstad.
\newblock The usability metric for user experience.
\newblock \emph{Interacting with computers}, 22\penalty0 (5):\penalty0 323--327, 2010.

\bibitem[wik()]{wikipediaKruskalWallisTest}
{K}ruskal–{W}allis test - {W}ikipedia --- en.wikipedia.org.
\newblock \url{https://en.wikipedia.org/wiki/Kruskal%E2%80%93Wallis_test}.
\newblock [Accessed 27-04-2025].

\bibitem[Jia et~al.(2024)Jia, Blessing, Jiang, Reuss, Donat, Lioutikov, and Neumann]{jia2024towards}
X.~Jia, D.~Blessing, X.~Jiang, M.~Reuss, A.~Donat, R.~Lioutikov, and G.~Neumann.
\newblock Towards diverse behaviors: A benchmark for imitation learning with human demonstrations.
\newblock \emph{arXiv preprint arXiv:2402.14606}, 2024.

\bibitem[Reuss et~al.(2023)Reuss, Li, Jia, and Lioutikov]{reuss2023goal}
M.~Reuss, M.~Li, X.~Jia, and R.~Lioutikov.
\newblock Goal-conditioned imitation learning using score-based diffusion policies.
\newblock \emph{arXiv preprint arXiv:2304.02532}, 2023.

\bibitem[Chi et~al.(2024)Chi, Xu, Feng, Cousineau, Du, Burchfiel, Tedrake, and Song]{chi2024diffusionpolicy}
C.~Chi, Z.~Xu, S.~Feng, E.~Cousineau, Y.~Du, B.~Burchfiel, R.~Tedrake, and S.~Song.
\newblock Diffusion policy: Visuomotor policy learning via action diffusion.
\newblock \emph{The International Journal of Robotics Research}, 2024.

\bibitem[Reuss et~al.(2024)Reuss, Ya{\u{g}}murlu, Wenzel, and Lioutikov]{reuss2024multimodal}
M.~Reuss, {\"O}.~E. Ya{\u{g}}murlu, F.~Wenzel, and R.~Lioutikov.
\newblock Multimodal diffusion transformer: Learning versatile behavior from multimodal goals.
\newblock \emph{arXiv preprint arXiv:2407.05996}, 2024.

\bibitem[Zhou et~al.(2025)Zhou, Liao, Huang, Tang, Otto, Jia, Jiang, Hilber, Li, Wang, et~al.]{zhou2025beast}
H.~Zhou, W.~Liao, X.~Huang, Y.~Tang, F.~Otto, X.~Jia, X.~Jiang, S.~Hilber, G.~Li, Q.~Wang, et~al.
\newblock Beast: Efficient tokenization of b-splines encoded action sequences for imitation learning.
\newblock \emph{arXiv preprint arXiv:2506.06072}, 2025.

\bibitem[Bell and Hoberock(2012)]{bell2012thrust}
N.~Bell and J.~Hoberock.
\newblock Thrust: A productivity-oriented library for cuda.
\newblock In \emph{GPU computing gems Jade edition}, pages 359--371. Elsevier, 2012.

\bibitem[{NVIDIA Corporation}(2024{\natexlab{a}})]{nvidia_isaac_sim}
{NVIDIA Corporation}.
\newblock {NVIDIA Isaac Sim}, 2024{\natexlab{a}}.
\newblock URL \url{https://developer.nvidia.com/isaac-sim}.

\bibitem[{NVIDIA Corporation}(2024{\natexlab{b}})]{nvidia_omniverse}
{NVIDIA Corporation}.
\newblock {NVIDIA Omniverse}, 2024{\natexlab{b}}.
\newblock URL \url{https://www.nvidia.com/en-us/omniverse/}.

\bibitem[Gong et~al.(2023)Gong, Huang, Zhao, Geng, Gao, Wu, Ai, Zhou, Terzopoulos, Zhu, et~al.]{gong2023arnold}
R.~Gong, J.~Huang, Y.~Zhao, H.~Geng, X.~Gao, Q.~Wu, W.~Ai, Z.~Zhou, D.~Terzopoulos, S.-C. Zhu, et~al.
\newblock Arnold: A benchmark for language-grounded task learning with continuous states in realistic 3d scenes.
\newblock In \emph{Proceedings of the IEEE/CVF International Conference on Computer Vision (ICCV)}, 2023.

\bibitem[Studios(2016)]{pixarUSD}
P.~A. Studios.
\newblock Universal scene description (usd).
\newblock \url{https://github.com/PixarAnimationStudios/USD}, 2016.
\newblock Accessed: 2025-01-29.

\bibitem[James et~al.(2020)James, Ma, Arrojo, and Davison]{james2020rlbench}
S.~James, Z.~Ma, D.~R. Arrojo, and A.~J. Davison.
\newblock Rlbench: The robot learning benchmark \& learning environment.
\newblock \emph{IEEE Robotics and Automation Letters}, 5\penalty0 (2):\penalty0 3019--3026, 2020.

\bibitem[tri()]{trimeshTrimeshTrimesh}
trimesh - trimesh 4.6.1 documentation --- trimesh.org.
\newblock \url{https://trimesh.org/trimesh.html}.
\newblock [Accessed 01-02-2025].

\bibitem[Wrede et~al.(2013)Wrede, Emmerich, Gr{\"u}nberg, Nordmann, Swadzba, and Steil]{wrede2013user}
S.~Wrede, C.~Emmerich, R.~Gr{\"u}nberg, A.~Nordmann, A.~Swadzba, and J.~Steil.
\newblock A user study on kinesthetic teaching of redundant robots in task and configuration space.
\newblock \emph{Journal of Human-Robot Interaction}, 2\penalty0 (1):\penalty0 56--81, 2013.

\bibitem[Sukkar et~al.(2023)Sukkar, Moreno, Vidal-Calleja, and Deuse]{sukkar2023guided}
F.~Sukkar, V.~H. Moreno, T.~Vidal-Calleja, and J.~Deuse.
\newblock Guided learning from demonstration for robust transferability.
\newblock In \emph{2023 IEEE International Conference on Robotics and Automation (ICRA)}, pages 5048--5054. IEEE, 2023.

\bibitem[Pettinger et~al.(2020)Pettinger, Elliott, Fan, and Pryor]{pettinger2020reducing}
A.~Pettinger, C.~Elliott, P.~Fan, and M.~Pryor.
\newblock Reducing the teleoperator’s cognitive burden for complex contact tasks using affordance primitives.
\newblock In \emph{2020 IEEE/RSJ International Conference on Intelligent Robots and Systems (IROS)}, pages 11513--11518. IEEE, 2020.

\bibitem[Lin et~al.(2022)Lin, Krishnan, and Li]{lin2022comparison}
T.-C. Lin, A.~U. Krishnan, and Z.~Li.
\newblock Comparison of haptic and augmented reality visual cues for assisting tele-manipulation.
\newblock In \emph{2022 International Conference on Robotics and Automation (ICRA)}, pages 9309--9316. IEEE, 2022.

\bibitem[ste()]{steampoweredSteamVR}
{S}team{V}{R} --- store.steampowered.com.
\newblock \url{https://store.steampowered.com/steamvr}.
\newblock [Accessed 31-01-2025].

\bibitem[thr()]{threejsThreejsDocs}
three.js docs --- threejs.org.
\newblock \url{https://threejs.org/docs/index.html#manual/en/introduction/Creating-a-scene}.
\newblock [Accessed 30-01-2025].

\end{thebibliography}
